\documentclass[sigconf]{acmart}

\usepackage{graphicx}
\usepackage{booktabs}
\usepackage{tabularx}
\usepackage{multirow}
\usepackage{mwe}  
\usepackage{tikz}
\usepackage{pifont}
\usepackage{caption} 
\usepackage{subcaption}  
\usepackage{pgfplots}
\usepackage[percent]{overpic} 
\pgfplotsset{compat=newest}
\usepackage{mathrsfs}
\usepackage{adjustbox}
\usepackage{cleveref}
\usepackage{titlesec}

\newcommand{\mypara}[1]{\noindent\textbf{#1.}\xspace}
\newcommand{\name}{\textit{FreeInsert}\xspace}

\newcommand{\cmark}{\checkmark}  
\newcommand{\xmark}{$\times$}  

\AtBeginDocument{%
  }

\setcopyright{acmlicensed}
\copyrightyear{2018}
\acmYear{2018}
\acmDOI{XXXXXXX.XXXXXXX}
\acmConference[Conference acronym 'XX]{Make sure to enter the correct
  conference title from your rights confirmation email}{June 03--05,
  2018}{Woodstock, NY}
\acmISBN{978-1-4503-XXXX-X/2018/06}

\begin{document}

\title{FreeInsert: Disentangled Text-Guided Object Insertion in 3D
Gaussian Scene without Spatial Priors}

\author{Chenxi Li}
\affiliation{
\institution{
Tianjin University
}
\city{Tianjin}
\country{China}
}
\email{chenxili2024@tju.edu.cn}

\author{Weijie Wang$^\dag$}
\affiliation{%
  \institution{University of Trento}
  \city{Trento}
  \country{Italy}}
\email{weijie.wang@unitn.it}

\author{Qiang Li}
\affiliation{%
\institution{Tianjin University}
\city{Tianjin}
\country{China}}
\email{liqiang@tju.edu.cn}

\author{Bruno Lepri}
\affiliation{%
 \institution{Fondazione Bruno Kessler}
  \city{Trento}
  \country{Italy}}
  \email{lepri@fbk.eu}

\author{Nicu Sebe}
\affiliation{%
  \institution{University of Trento}
  \city{Trento}
  \country{Italy}}
  \email{niculae.sebe@unitn.it }

\author{Weizhi Nie}
\affiliation{%
  \institution{Tianjin University}
  \city{Tianjin}
  \country{China}}
\email{weizhinie@tju.edu.cn}

\thanks{
Weijie Wang$^\dag$ is the Corresponding author.
}

\renewcommand{\shortauthors}{Trovato et al.}

\begin{abstract} 
Text-driven object insertion in the 3D scene is an emerging task that enables intuitive scene editing through natural language.  
Despite its potential, existing 2D editing-based methods often suffer from reliance on spatial priors such as 2D masks, 3D bounding boxes to. 
And they struggle to ensure inserted object consistency.
These constraints hinder flexibility and scalability in real-world applications.
In this paper, we propose \name, a novel framework that leverages foundation models (MLLMs, LGM, and diffusion model) to disentangle object generation and spatial placement, enabling unsupervised and flexible object insertion in 3D scenes without spatial priors. 
\name begins with an MLLM-based parser that extracts structured semantics—including object types, spatial relationships, and attachment regions—from user instructions.
These semantics guide both the reconstruction of the inserted object for 3D consistency and the learning of its degrees of freedom. 
We first leverage the spatial reasoning capabilities of MLLMs to initialize the object's pose and scale. 
To further enhance natural integration with the scene, a hierarchical spatially-aware stage is employed to refine the object’s placement, incorporating both the spatial semantics and priors inferred by the MLLM. 
Finally, the object’s appearance is enhanced using inserted-object image to improve visual fidelity.
Experimental results demonstrate that \name enables semantically coherent, spatially precise, and visually realistic 3D insertions, without requiring any spatial priors, offering a user-friendly and flexible editing experience.

\end{abstract}



\received{20 February 2007}
\received[revised]{12 March 2009}
\received[accepted]{5 June 2009}

\maketitle

\section{Introduction}
Text-driven 3D generation~\cite{poole2022dreamfusiontextto3dusing2d,zhou2024gala3d,bokhovkin2024scenefactor} and editing~\cite{xu2024gg,zhou2024edit3d,barda2024instant3dit} are gaining traction for enabling the intuitive customization of digital content with a few words. 
Despite recent advances~\cite{dong2023vica,haque2023instruct,mirzaei2025watch,khalid2024latenteditor,wang2024gaussianeditor,zhuang2023dreameditor,song2023blending} that have made significant progress in editing the geometry and appearance of scene components, flexibly inserting new objects into the scene remains challenging due to difficulties in precise placement and seamless integration.

\begin{figure}[!t]
\includegraphics[width=\columnwidth]{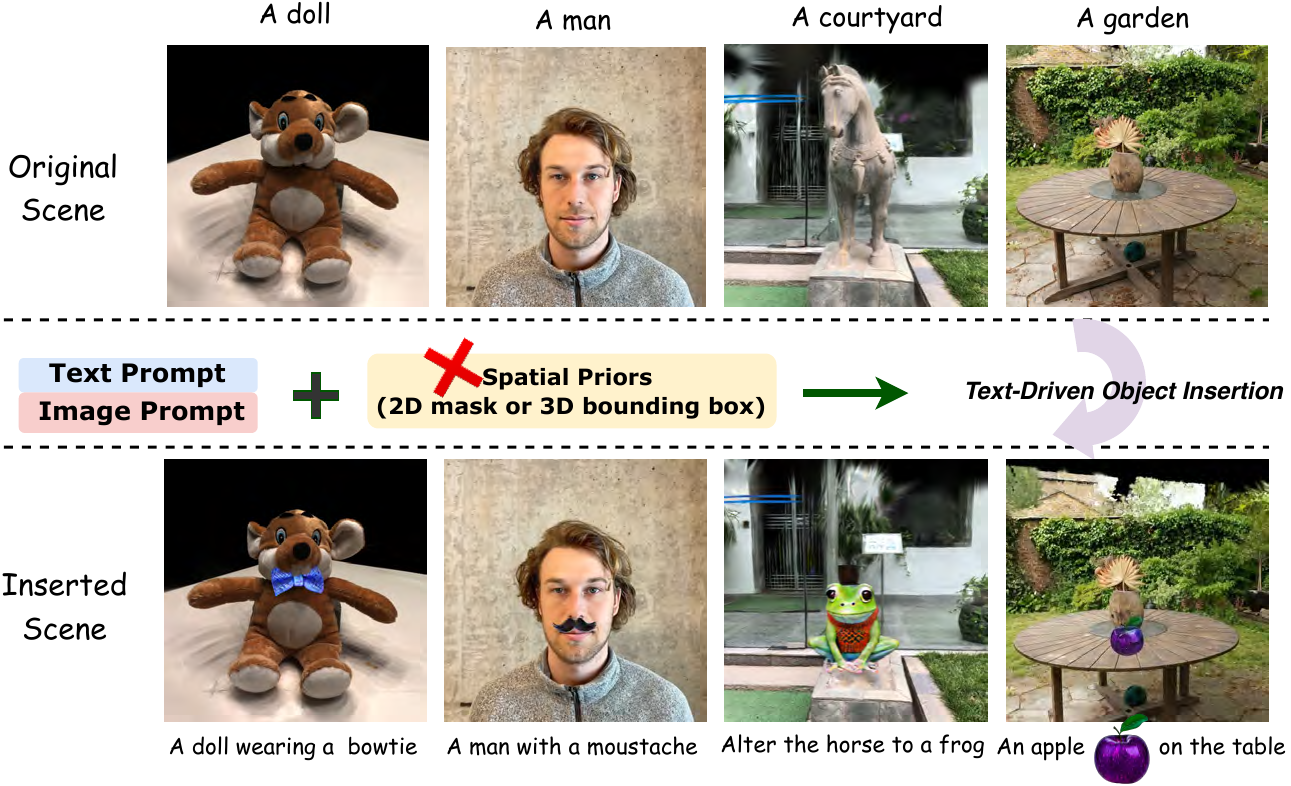}
  \caption{\textbf{``No Spatial Priors, Just Prompts.''} Compared to existing methods that require user-provided spatial priors, limiting their practicality, our method enables flexible text-driven object insertion without any need for such priors (e.g., 2D masks or 3D bounding boxes).
  Given only a text prompt (The image prompt is optional), \name naturally inserts objects across diverse scenes.}
  \vspace{-5mm}
  \label{fig:teaser}
\end{figure}

Recent 3D editing methods leverage diffusion models by first performing text-guided 2D edits on single~\cite{shahbazi2024inserf,chen2024gaussianeditor} or multi-view images~\cite{haque2023instruct,zhuang2023dreameditor}, then lifting them to 3D.
Relying solely on textual descriptions for object insertion often leads to insertion failure and suboptimal results due to misinterpretation of the text~\cite{zhuang2024tip, zhuang2023dreameditor}, as shown in \Cref{fig:comparision with advanced}, exemplified by methods like Instruct-NeRF2NeRF~\cite{haque2023instruct} and GaussCtrl~\cite{wu2024gaussctrl}.
Some methods introduce attention mechanisms to capture the spatial relationship between the inserted object and the scene.
However, these methods still struggle with accurately determining the inserted object’s pose and scale~\cite{zhuang2023dreameditor, sun2024gseditpro, zhou2024edit3d}.
To address this limitation, other methods leverage user-provided 2D masks \cite{chen2024gaussianeditor,shahbazi2024inserf} or 3D bounding boxes \cite{zhuang2024tip,shum2024language} as strong constraints to achieve more controllable and concise insertion.
Nevertheless, they often demand specialized expertise~\cite{zhuang2024tip} and considerable manual effort, limiting their practical usability.
In addition, they still face challenges with inaccurate depth estimation~\cite{chen2024gaussianeditor} and inconsistent 3D multi-view reconstruction due to the modality gap between 2D and 3D.
We summarize the above discussion as shown in ~\Cref{tab:intro:diff}.

Inspired by this, achieving flexible and high-quality object insertion into 3D scenes without manual supervision remains underexplored. 
The advent of large-scale models \cite{achiam2023gpt,deitke2024molmo,xiao2024florence,liu2024deepseek,grattafiori2024llama}, which have acquired human commonsense knowledge, has made unsupervised learning increasingly promising.
In this paper, we propose \name, a method that leverages foundation models (MLLMs~\cite{achiam2023gpt,deitke2024molmo}, LGM~\cite{tang2024lgm} and Diffusion model~\cite{rombach2022high} ) to assist object insertion in 3D scenes without relying on any spatial priors.
Our method removes the need for spatial priors by inferring object insertion directly from high-level textual cues (e.g., \textit{“Add [object] to/on [target]”}) as~\Cref{fig:teaser} shows.
We argue that \textbf{the insertion process can essentially be viewed as first generating a object, followed by estimating the transformation, which defines the inserted object's degrees of freedom (pose and scale) relative to the scene.} 


Specifically,  we disentangle the object insertion process into object generation and its parameterized degrees of freedom (DoF) estimation, both guided by textual descriptions with foundation models.
We first obtain a text instruction from the user and parse it into structured semantics (e.g., object type, spatial relation, attachment region) using an MLLM-based \cite{achiam2023gpt} object insertion parser. 
This enables precise, controllable object insertion that aligns with the user’s intent.
We then employ a 3D-consistent reconstruction model~\cite{tang2024lgm} to obtain an initial Gaussian-based object model, which is coarsely inserted into the scene guided by the visual and spatial reasoning capabilities of MLLMs\cite{achiam2023gpt,deitke2024molmo}.
This step inherently circumvents the inconsistency issues associated with 2D editing-based methods.
While the feed-forward procedure provides a lightweight 3D layout, it often suffers from suboptimal placement and imperfect geometry.
To address these issues, we propose a two-stage refinement.
First, the Hierarchical Spatial-Aware Refinement stage optimizes the object’s DoF via spatially-aware score distillation sampling (SSDS)~\cite{chen2024comboverse} from pretrained diffusion model~\cite{rombach2022high}.
This stage leverages MLLM-derived reasoning results to align the object's DoF with both local and global spatial semantics, enhancing more precise and controllable placement. 
These reasoning results also help the model handle rare spatial composition e.g., ``\textit{Add a pair of sunglasses on the forehead}'', thereby improving robustness.
In the final appearance refinement stage, we fine-tune a pretrained diffusion model on multi-view renderings of the optimized inserted object and its corresponding inserted-object image,
and use it to enhance the object's appearance.
By disentangling placement from object generation, our method enables flexible semantic control over insertion while preserving object quality and ensuring coherent, plausible 3D scenes.

\begin{table}[t]
\centering
\small
\caption{Comparison of existing methods of object insertion in 3D scenes. 
Ours can achieve high-quality object insertion without manual supervision while keeping 3D consistency. }
\vspace{-3mm}
\begin{adjustbox}{max width=0.9\columnwidth}
\begin{tabular}{lcccc}
\toprule
 & \begin{tabular}[c]{@{}c@{}}No Required \\Manual Supervision\end{tabular}
 & \begin{tabular}[c]{@{}c@{}}3D View\\Consistency\end{tabular} 
 & \begin{tabular}[c]{@{}c@{}}Support\\Image-Prompts\end{tabular} \\
\midrule
Instruct-N2N\cite{haque2023instruct} & \cmark & \xmark & \xmark\\
GaussCtrl~\cite{wu2024gaussctrl}  & \cmark & \cmark & \xmark \\
GaussianEditor~\cite{chen2024gaussianeditor}  & \xmark & \xmark & \xmark \\
TIP-Editor~\cite{zhuang2024tip}  & \xmark & \xmark & \cmark \\
\name  &\cmark & \cmark & \cmark \\
\bottomrule
\end{tabular}
\end{adjustbox}
\label{tab:intro:diff}
\end{table}

To evaluate the proposed method, we applied it to various scenarios, including object-centric, human-centric, and complex outdoor scenes. Our experimental results demonstrate that the proposed approach can insert diverse objects into 3D scenes without requiring manual supervision while achieving multi-view consistent object quality. 
In summary, our contributions are as follows:
\begin{itemize}
    \item We address consistent object insertion in diverse 3D scenes using only textual input, removing the need for spatial priors and outperforming existing methods through a framework that disentangle object generation and spatial placement.
    \item We propose a DoF optimization method for object insertion, using the reasoning capabilities of MLLMs and diffusion models in place of manual supervision.
    The MLLM's semantic and spatial priors further support SSDS in enhancing precision and robustness.
    \item  
    We ensure high-quality object generation by maintaining 3D shape consistency via a reconstruction model and refining visual appearance.
    \item We present the first baseline for evaluating unsupervised 3D scene insertion, with experiments showing competitive performance against state-of-the-art methods.

\end{itemize}

\begin{figure*}
\begin{center}
    \includegraphics[width=0.95\linewidth]{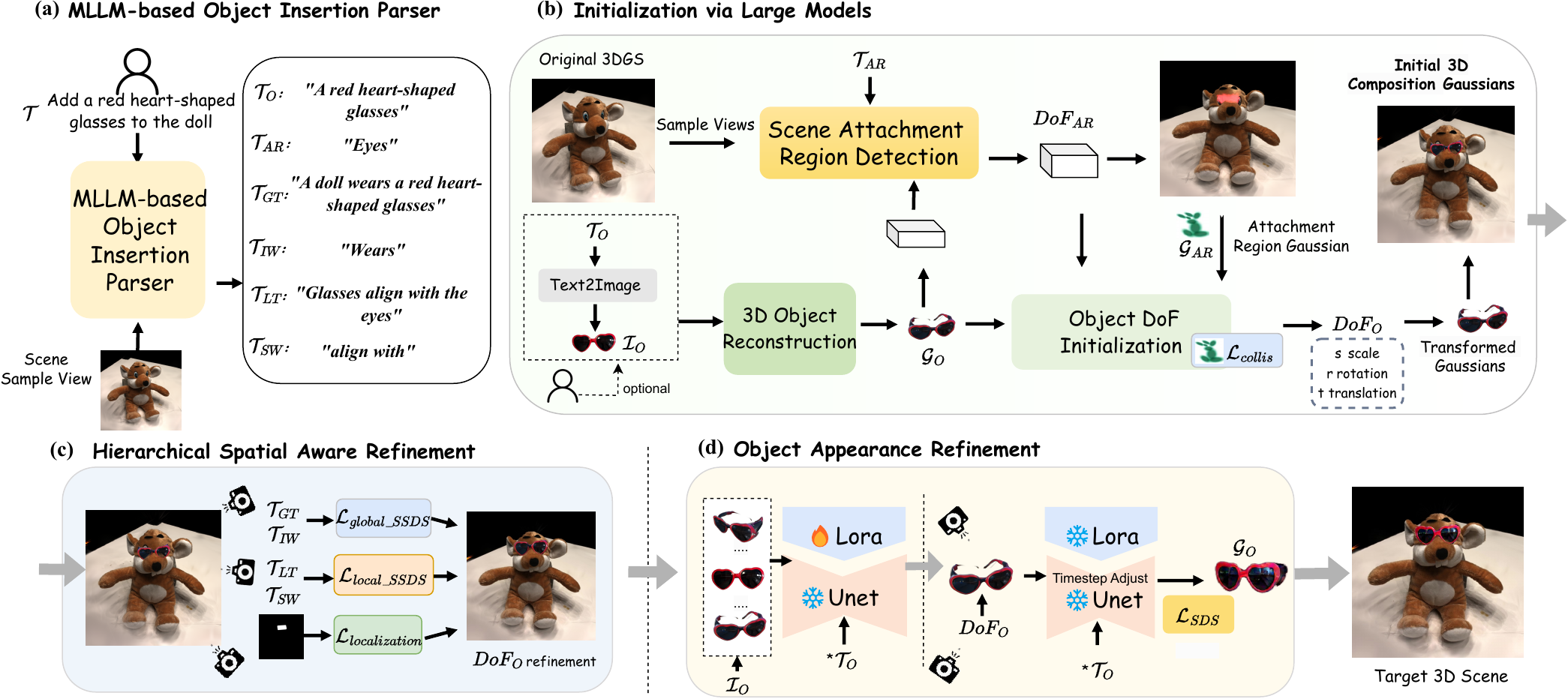}
\end{center}
    \vspace{-10pt}
   \caption{Overview of \name. 
    Given an text prompt \( \mathcal{T} \) and optionally an image prompt \( \mathcal{I}_{\textit{O}} \), the object insertion process includes four stages:
    (a) The MLLM-based Object Insertion Parser (see \Cref{sec:llm_parser}) first extracts structured semantics to support the subsequent stages. 
   (b) The Initialization via Large Models (see \Cref{sec:dof_init}) stage generates object and initializes its \( \textit{DoF}_{\textit{O}} \) in the scene . 
   (c) The Hierarchical Spatial Aware Refinement (see \Cref{sec:spatial_aware}) stage refines the \( \textit{DoF}_{\textit{O}} \). 
   (d) The final stage, Object Appearance Refinement (see \Cref{sec:appear_opt}), enhances the object’s visual quality using object image\( \mathcal{I}_{\textit{O}} \).
   }  
\label{fig:overview}
\end{figure*}

\section{Related Works}
\noindent\textbf{Text-Guided 3D Scene Editing.} 
Text-driven 3D scene editing has seen rapid progress, thanks to the rise of diffusion models~\cite{rombach2022high,ho2020denoising}. 
Most methods \cite{kim2023collaborative,koo2024posterior,haque2023instruct,mirzaei2025watch,park2024ednerf,dong2023vica,wang2024innerf360} focus on modifying existing content, either globally or locally. 
Local editing requires precise localization to avoid affecting unrelated regions, which remains challenging.
While some works use implicit cues from models like InstructPix2Pix~\cite{brooks2023instructpix2pix} or ControlNet~\cite{zhang2023adding}, others incorporate explicit constraints such as segmentation masks~\cite{wang2024gaussianeditor} or cross-attention maps~\cite{zhuang2023dreameditor}.
However, these methods struggle with 3D object insertion, which demands reasoning about semantically appropriate yet physically unoccupied regions for placement.
In this work, we primarily focus on the task of object insertion in 3D scenes.

\noindent\textbf{Object Insertion in 3D Scene.} 
In contrast to modifying existing scene content, object insertion remains underexplored. MVInpainter~\cite{caomvinpainter} leverages segmentation to identify support regions like table surfaces, but struggles with fine-grained insertions on objects or humans. GaussianEditor~\cite{chen2024gaussianeditor} and InseRF~\cite{shahbazi2024inserf}  insert objects using 3D reconstruct models, yet still require user-provided 2D masks and suffer from depth-related localization issues. FocalDreamer~\cite{li2024focaldreamer} attaches parts to base shapes but depends on user-specified 3D parameters (e.g., rotation, translation, scale), and lacks generalization to complex scenes. 
Other methods~\cite{zhuang2024tip,shum2024language} guide object generation via diffusion using 3D bounding boxes, which imposes a burden on users.
In this work, we aim for unsupervised and broadly applicable 3D object insertion, removing the need for manual annotations or spatial priors.

\noindent\textbf{Large Language Models in 3D Generation and Editing.}
LLMs, like GPT \cite{achiam2023gpt}
and Llama \cite{grattafiori2024llama} series, have exhibited outstanding efficacy in many text-related tasks. 
\citet{zhou2024gala3d} and \citet{zhou2025layoutyourd} utilize LLMs to provide coarse compositional spatial information from textual descriptions to construct the 3D scene. 
The multi-modal variants of LLMs \cite{deitke2024molmo,achiam2023gpt} incorporate images and are additionally trained on image-text pairs, showing impressive results for visual captioning and vision question-answering (VQA). 
Notably, Molmo \cite{deitke2024molmo} can perform pixel-level localization mainly because it was trained with richly annotated image data.
This capability is crucial for robust spatial grounding in vision-language tasks.
GG-Editor \cite{xu2024gg} first exploits GPT-4V~\cite{achiam2023gpt} to better understand both the textual and 3D visual inputs and then infer reasonable local regions for 3D editing.
However, it primarily targets object editing and its preliminary use of MLLM struggles to ensure spatial precision.
In this work, we leverage the text reasoning capabilities and spatial relationship understanding of GPT-4 \cite{achiam2023gpt} and Molmo~ \cite{deitke2024molmo}, supplemented by a basic detection model \cite{xiao2024florence}, to eliminate the reliance on manually provided priors in object insertion.

\section{Method}
\subsection{Problem Statement}
\Cref{fig:overview} illustrates the overall framework of \name.
Given a group of 3D Gaussians \( \mathcal{G}_{\textit{S}} \) for an input scene and a text prompt \( \mathcal{T} \) guiding the insertion of an object into the scene, our algorithm performs high-quality, semantically consistent object insertion \emph{without any manual supervision (e.g., 3D bounding boxes or masks).}
We decouple the object insertion task into object generation and the optimization of the object's 3D degrees of freedom \( \textit{DoF}_{\textit{O}} \) (rotation, translation, scale), both guided by semantic alignment between the resulting scene and the user prompts.
The resulting scene is represented by a new set of Gaussians, \( \mathcal{G}_{\textit{inserted}} \).
Moreover, our method allows image prompt \( \mathcal{I}_{\textit{O}} \) as input to specify the object's appearance. 
Formally, the insertion process is defined as:
\begin{equation}
\small
\mathcal{G}_{\textit{inserted}} = \mathcal{E} \left( \mathcal{G}_{\textit{S}},\ \mathcal{T},\ \mathcal{I}_{\textit{O}}  \right),
\end{equation}
where $\mathcal{E}$ denotes the process applied to the inserted object, including its generation, DoF learning in the context of the scene \( \mathcal{G}_{\textit{S}} \), and appearance refinement.
\subsection{MLLM-based Object Insertion Parser}
\label{sec:llm_parser}
A key challenge in unsupervised object insertion is converting high-level user intent into structured, fine-grained guidance. 
To address this, we introduce an MLLM-based Object Insertion Parser (MLLM-OIP) that utilizes the MLLM’s spatial understanding capability to parse the instruction $\mathcal{T}$ into semantically prompts, providing essential guidance for the subsequent object insertion.
Specifically, we provide a prompt template \( \mathcal{T}_{\textit{parser}} \) and a sampled scene image \( \mathcal{I}_{\textit{S}} \) as input to the multimodal LLM \( \mathcal{M}_{\textit{MLLM}} \)~\cite{achiam2023gpt} to obtain structured outputs.
The prompts generation process is formalized as : 
\begin{equation}
\small
\left( 
\mathcal{T}_{\textit{O}},\ 
\mathcal{T}_{\textit{AR}},\ 
\mathcal{T}_{\textit{GT}},\ 
\mathcal{T}_{\textit{IW}},\ 
\mathcal{T}_{\textit{LT}},\ 
\mathcal{T}_{\textit{SW}} 
\right) = 
\mathcal{M}_{\textit{MLLM}} \left( \mathcal{T},\ \mathcal{T}_{\textit{parser}}, \
\mathcal{I}_{\textit{S}} \right)
\end{equation}
Here, the Object Prompt (\( \mathcal{T}_{\textit{O}} \)) is used for 3D object generation and appearance refinement stage. 
The Attachment Region Prompt(\( \mathcal{T}_{\textit{AR}} \)) plays a crucial role in the initialization of the object’s degrees of freedom \( \textit{DoF}_{\textit{O}} \). 
The remaining four prompts including the Global Target Prompt (\( \mathcal{T}_{\textit{GT}} \)) and its Object Interaction Word (\( \mathcal{T}_{\textit{IW}} \)), the Local Target Prompt (\( \mathcal{T}_{\textit{LT}} \)) and its Spatial Relationship Word (\( \mathcal{T}_{\textit{SW}} \)) are employed during the hierarchical spatial-aware refinement stage to refine the \( \textit{DoF}_{\textit{O}} \), supporting global-local semantic alignment.

\subsection{Initialization via Large Models}
\label{sec:dof_init}
\noindent\textbf{Object from Prompts.} 
To avoid 3D inconsistency, we first use a text-to-image (T2I)~\cite{rombach2022high} model to synthesize a Text-generated image \( \mathcal{I}_{\textit{O}} \) of the object from the object description prompt \( \mathcal{T}_{\textit{O}} \).  
The synthesized image is then used to recover the 3D geometry $\mathcal{G}_{O}$ via LGM~\cite{tang2024lgm}, a single-view reconstruction model that achieve a trade-off between reconstruction quality and efficiency.
Other lightweight 3D reconstruction methods \cite{long2024wonder3d,hong2023lrm} can also be adopted.
In addition, \( \mathcal{I}_{\textit{O}} \) can be directly specified by the user, allowing for more precise control over the object's appearance. 

\noindent\textbf{Scene's Attachment Region Detection.} 
Intuitively, an object's placement is influenced by the attachment region of the scene and the degrees of freedom within that region.
Driven by this, we extract an attachment region \( \mathcal{G}_{\textit{AR}} \) and its associated degrees of freedom \( \textit{DoF}_{\textit{AR}} \)(\( \textit{s}_{\textit{AR}} \), \( \textit{r}_{\textit{AR}} \), \( \textit{t}_{\textit{AR}} \)) from the 3D scene based on the Attachment Region Prompt \( \mathcal{T}_{\textit{AR}} \). 
This region serves as a crucial spatial reference, guiding initializing the inserted object \( \textit{DoF}_{\textit{O}} \).
Specifically, we employ an open-vocabulary detection model Florence2 \cite{xiao2024florence} to localize 2D bounding boxes across smapled views from the scene with camera poses \( \mathcal{C}_{\textit{cam}} \), guided by \( \mathcal{T}_{\textit{AR}} \).
For each view, the detected box is converted into a binary mask \( \mathcal{I}_{\textit{BAR}} \), representing the candidate attachment region.
The 3D attachment area is parameterized by the degrees of freedom of a initial 3D bounding box \( \mathcal{B}_{\textit{init}} \). 
We optimize the attachment by computing cross-entropy between the projected transformed bounding box and the detected attachment region mask \( \mathcal{B}_{\text{AR}} \) across all camera views.
Thus, \( \textit{DoF}_{\textit{AR}} \) is calculated as follows:
\begin{equation}
\small
\textit{DoF}_{\textit{AR}} = \arg\min_{\theta} \sum_{\mathcal{T}_{\textit{cam}} \in \mathcal{C}_{\textit{cam}}} \mathcal{L}_{\textit{BCE}}\left( \textit{Proj}\left( \mathcal{B}, \mathcal{T}_{\textit{cam}} \right), \mathcal{I}_{\textit{BAR}}^{(\mathcal{T}_{\textit{cam}})} \right),
\end{equation}
where \( \theta=(\textit{s},\textit{r},\textit{t}) \) denotes the transformation parameters for the canonical box \( \mathcal{B}_{\textit{init}} \), \( \mathcal{F}_{\textit{affine}} \) is the affine transformation function, and the transformed 3D bounding box is computed as
\(
\mathcal{B} = \mathcal{F}_{\textit{affine}}(\mathcal{B}_{\textit{init}}, \theta)
\). 
The function \( \text{Proj}(\mathcal{B}, \mathcal{T}_{\textit{cam}}) \) denotes the 2D projection of the 3D bounding box \( \mathcal{B} \) onto the image plane under the camera pose \( \mathcal{T}_{\textit{cam}} \).
After obtaining \( \textit{DoF}_{\textit{AR}} \), the attachment region \( \mathcal{G}_{\textit{AR}} \) is extracted by selecting 3D Gaussians from the scene representation \( \mathcal{G}_{\textit{S}} \) within the transformed bounding box \( \mathcal{B}_{\textit{AR}} = \mathcal{F}_{\textit{affine}}(\mathcal{B}_{\textit{init}}, \textit{DoF}_{\textit{AR}}) \). 
Formally, the attachment region is defined as:
\[
\mathcal{G}_{\textit{AR}} = \{ g \in \mathcal{G}_s \mid g \in \mathcal{B}_{\textit{AR}} \}
\]
\noindent\textbf{Object's DOF Initialization.} 
Once obtained the attachment region \( \mathcal{G}_{\textit{AR}} \) and its associated transformation \( \textit{DoF}_{\textit{AR}} \), 
we initialize the inserted object's degrees of freedom \( \textit{DoF}_{\textit{O}} \) 
(\( \textit{s}_{\textit{O}} \), \( \textit{r}_{\textit{O}} \), \( \textit{t}_{\textit{O}} \)) 
accordingly.
For the \( \textit{s}_{\textit{O}} \) initialization, we assume an intuitive real-world prior: there exists a reasonable relative scale ratio \( \lambda_{\textit{rel}} \) between the inserted object and the attachment region, which helps ensure a plausible insertion. 
This ratio is implicitly understood by large-scale language models. 
Therefore, we leverage \(\mathcal{M}_{\textit{MLLM}} \) to predict \( \lambda_{\textit{rel}} \), and compute the object scale as \( s_{\textit{O}} = \textit{s}_{\textit{AR}} \cdot \lambda_{\textit{rel}} \). 
Considering the uncertainty in MLLM predictions and the influence of scale initialization quality on subsequent refinement, we adopt an iterative strategy. 
After initializing \( \textit{r}_{\textit{O}} \) and \( \textit{t}_{\textit{O}} \), we render the scene with the inserted object and iteratively interact with the MLLM, using visual feedback to adjust \( \textit{s}_{\textit{O}} \) and improve realism and integration.

For the \( \textit{r}_{\textit{O}} \) initialization, 
we leverage \( \mathcal{M}_{\textit{MLLM}} \) to initialize a semantically appropriate object rotation.
Given a or MLLM-suggested primary scene viewpoint, we render a scene image \( \mathcal{I}_{\textit{S}} \), and sample a set of object-centric renderings \( \{ \mathcal{I}_{\textit{O}}^{(r)} \}_{r \in \mathcal{R}} \) for the inserted object, where each \( r \in \mathcal{R} \) corresponds to a unique azimuth-elevation rotation \( (\phi, \theta) \in [0, 2\pi) \times [0, \pi) \).
Based on the \( \mathcal{I}_{\textit{S}} \), the rendering set \( \{ \mathcal{I}_{\textit{O}}^{(r)} \} \), and the Global Target Prompt \( \mathcal{T}_{\textit{GT}} \), the model can select the optimal rotation \( \textit{r}_{\textit{O}} \) that maximizes a semantic alignment score:
\begin{equation}
\small
\textit{r}_{\textit{O}} = \arg\max_{r \in \mathcal{R}} \; \mathcal{M}_{\textit{MLLM}} \big( \mathcal{I}_{\textit{S}}, \mathcal{I}_{\textit{O}}^{(r)}, \mathcal{T}_{\textit{GT}} \big)
\end{equation}
where \( \mathcal{M}_{\textit{MLLM}} \) evaluates sementic plausibility the placement aligns with the scene.
 
To initialize the \( \textit{t}_{\textit{O}} \), 
we use strong pixel-level semantic spatial localization capability of Molmo~\cite{deitke2024molmo} to predict a set of 2D object centers \( \{ c_O^{(v)} \}_{v \in \mathcal{V}} \) across multiple scene views with the Local Target \( \mathcal{T}_{\textit{LT}} \), using prompt like ``\textit{Point the position to add <\( \mathcal{T}_{\textit{LT}} \)>}''. 
Let \( \hat{\mathcal{G}}_O^{(t)} \) denote the object geometry after applying the transformation \( ( \textit{s}_{\textit{O}} \), \( \textit{r}_{\textit{O}}\), $t$ ), where $t$ is a optimized parameter.
For each view \( v \), we project the transformed object and compute the 2D centroid of its projection. 
The \( \textit{t}_{\textit{O}}\) is obtained by optimizing 
$t$ to minimize the discrepancy between the projected and the predicted centroids:
\begin{equation}
\small
\textit{t}_{O} = \arg\min_{t} \sum_{v \in \mathcal{V}} \left\| 
\emph{Centroid} \left( \pi_v \left( \hat{\mathcal{G}}_O^{(t)} \right) \right)
- c_O^{(v)} \right\|_2^2 + \mathcal{L}_{\text{coll}}(\mathcal{G}_{\text{AR}}, O_c)
\end{equation}
Here, \( \pi_v(\cdot) \) is the camera projection function for view \( v \), and \emph{Centroid ($\cdot$)} computes the 2D projected center of the object.
To ensure physical plausibility during object insertion,
we introduce the collision loss~\cite{zhou2025layoutyourd} \( \mathcal{L}_{\text{coll}} \), which penalizes interpenetration between the object centriod \( O_c \) and the scene attachment region \( \mathcal{G}_{\text{AR}} \).


\subsection{Hierarchical Spatial Aware Refinement}
\label{sec:spatial_aware}
The initial \( \textit{DoF}_{\textit{O}} \) from $\mathcal{M}_{\textit{MLLM}}$ often lack spatial accuracy, hindering seamless scene integration.
Base on that, we then optimize the \( \textit{DoF}_{\textit{O}} \) using SSDS Loss~\cite{chen2024comboverse}, refining the object's placement in the scene.
The loss is defined as:
\begin{equation}
\small
\nabla_{\theta} \mathcal{L}_{\text{SSDS}} (\phi^{\star}, x) = \mathbb{E}_{t, \epsilon} \left[ w(t) (\hat{\epsilon}_{\phi^{\star}} (x_i; y, t) - \epsilon) \frac{\partial x}{\partial \theta} \right]
\end{equation}
Here, \( \theta \), \( x \), \( \phi^* \), and \( \hat{\epsilon}_{\phi^*} (x_t;  \mathcal{T} , t) \) denote the 3D representation, rendered image, spatial attention map, and the score function predicting noise \( \epsilon \) from the noised image \( x_t \) with text prompt \( \mathcal{T} \).
Unlike the original design for multi-object composition with high timesteps,
we find that lower timesteps are more effective for fine-grained DoF refinement in our setting, as it emphasizes local spatial details critical for precise alignment.

\noindent\textbf{Global-Local Collaborative Spatial Awareness.} 
Diffusion models often exhibit spatial biases due to training data imbalance, e.g., generating moustaches at a relatively large scale across the lower face. 
Large models trained on data-driven priors often fail to meet human expectations in spatial reasoning. (see \Cref{fig:Hierarchical Spatial Aware})
To address spatial ambiguity, we leverage spatial relation terms (e.g., ``on'', ``in front of'') to impose explicit constraints on object localization. 
Compared to general verbs like “wearing” or “with”, these relations encode more precise spatial priors, leading to more effective supervision for optimizing object placement.
We leverage spatial prompts inferred from MLLM-OIP,
which offers both global semantic grounding \(\mathcal{T}_{\textit{GT}} \) with interaction word\(\mathcal{T}_{\textit{IW}} \) and fine-grained positional cues \(\mathcal{T}_{\textit{LT}}\) with spatial relationship word\(\mathcal{T}_{\textit{SW}} \). 
We define a hierarchical spatial loss that jointly supervises local and global alignment:
\begin{equation}
\small
\mathcal{L}_\text{spatial} = \beta \cdot \mathcal{L}_\textit{ssds-global}(\mathcal{T}_{\textit{GT}}, \mathcal{T}_{\textit{IW}}) + (1 - \beta) \cdot \mathcal{L}_\textit{ssds-local}(\mathcal{T}_{\textit{LT}}, \mathcal{T}_{\textit{SW}}),
\end{equation}
\noindent\textbf{Attention-based Localization.} We found that due to the bias in the T2I model's training data, SSDS Loss exhibits limitations when handling rare spatial relationships. 
These inherent limitations restrict the effectiveness of prompt instructions.
To enable stronger spatial conditioning, we adopt attention-based localization loss \cite{zhuang2024tip}, enforcing tighter regional constraints as follows:
\begin{equation}
\small
\mathcal{L}_{loc} = \left(1 - \max_{s \in \mathcal{S}} (A_t^s) \right) + \lambda \sum_{s \in \tilde{\mathcal{S}}} \left\| A_t^s \right\|_2^2
\end{equation}
where $\lambda$ balances the two terms, $\mathcal{S}$ denotes the multi-view mask region projected from the 3D bounding box $\mathcal{B}$, obtained by tightly enclosing the object after DoF initialization, and $\tilde{\mathcal{S}}$ denotes the complementary region.
As shown in our ablation (\Cref{fig:attention-based}), it is essential for precise object placement within the designated area.

\subsection{Object Appearance Refinement}
\label{sec:appear_opt}

Once the object’s degrees of freedom are determined, a refinement module is introduced to enhance the visual quality of the inserted object $\mathcal{G}_{O}$.
Specifically, we refine $\mathcal{G}_{O}$ using the high-quality appearance from the inserted-object image $\mathcal{I}_{O}$ via LoRA \cite{hu2022lora}.

\noindent\textbf{Viewpoint Frequency Balancing.} 
To avoid the overfitting caused by using a single-view optimization, which often leads to 3D inconsistencies and missing object parts (e.g., a side view causing missing legs) as shown in \Cref{fig:view frequency} (b). 
We perform multi-view sampling of the inserted object.
Specifically,
given a set of views \( \{I_i\}_{i=1}^{N} \), rendered from the inserted object \( \mathcal{G}_{O} \). 
We estimate the pose \( P^* \) of the object image \( \mathcal{I}_{\textit{O}} \) by selecting the most similar view based on DINO feature similarity \cite{oquab2024dinov}. To ensure both appearance fidelity and geometric consistency, 
We construct the training set $\mathcal{D}_{\text{ref}}$ by combining the rendered multi-view images with repeated samples of the inserted-object image $\mathcal{I}_{\textit{O}}$ and its estimated pose \( P^* \), as follows:
\begin{equation}
\mathcal{D}_{\text{ref}} =
\{(I_i, P_i)\}_{i=1}^{N}
\cup \{(\mathcal{I}_O, P^*)\}_{j=1}^{M}
\quad (\text{repeated},\ M > N)
\end{equation}
The following objective is used to fine-tune the LoRA layers:
\begin{equation}
\small
\mathcal{L}_{\text{ref}} = \mathbb{E}_{z_i, I_i, P_i, y^{*}, \epsilon, t} \left\| \epsilon_{\phi_2}(z_i, t, P_i, I_i, y^{*}) - \epsilon \right\|_2^2, \quad (I_i, P_i) \sim \mathcal{D}_{\text{ref}}.
\end{equation}
\( z_i \) is the noisy latent of image \( I_i \), \( t \) is the diffusion timestep, and \( \epsilon \) is the target noise. 
The denoising network $\epsilon_{\phi_2}$, augmented with LoRA~\cite{hu2022lora}, is conditioned on $I_i$, its pose $P_i$, and the object-specific prompt $y^{*}$, e.g., ``A \texttt{<token>} dog'', which is formatting from \( \mathcal{T}_{\textit{O}} \).

\noindent\textbf{Appearance-Focused Refinement.} We employ fine-tuning diffusion to update the object Gaussian $\mathcal{G}_{O}$, guided by \(\mathcal{L}_{sds}\) \cite{poole2022dreamfusiontextto3dusing2d}. 
Under our setting, we sample from a lower range of timesteps during optimization to reduce the impact on the object's geometry (e.g., shape and scale), thereby encouraging the model to focus more on refining appearance details. The corresponding objective is defined as follows:
\begin{equation}
\small
\nabla_{\theta} \hat{\mathcal{L}}_{sds}(\phi, x) = \mathbb{E}_{\hat{t}, \epsilon} \left[ w(\hat{t}) \left( \hat{\epsilon}_{\phi}(x_i; y_i, \hat{t}) - \epsilon \right) \frac{\partial x}{\partial \theta} \right],
\end{equation}
where \( \hat{t} \) denotes the adjusted (lower) diffusion timestep.
\subsection{Object Replacement}
\label{sec:replace}
In addition, our method can be naturally extended to \textbf{object replacement} in the scene. 
Specifically, given a user prompt such as \textit{``Add a [new object] to replace [existing object]''}, the corresponding object to be replaced is identified through the Attachment Region \( \mathcal{G}_{\textit{AR}} \). 
We remove \( \mathcal{G}_{\textit{AR}} \) and then execute the standard insertion pipeline, enabling replacement without being constrained by the original object's geometry or structure.

\section{Experiments}
\begin{figure*}[t]
    \begin{minipage}{0.12\textwidth}
        \raggedright
        \textbf{Original Scene}
    \end{minipage}
    \begin{subfigure}{0.28\textwidth}
        \centering
        \begin{minipage}{\textwidth}
            \includegraphics[width=0.45\textwidth]{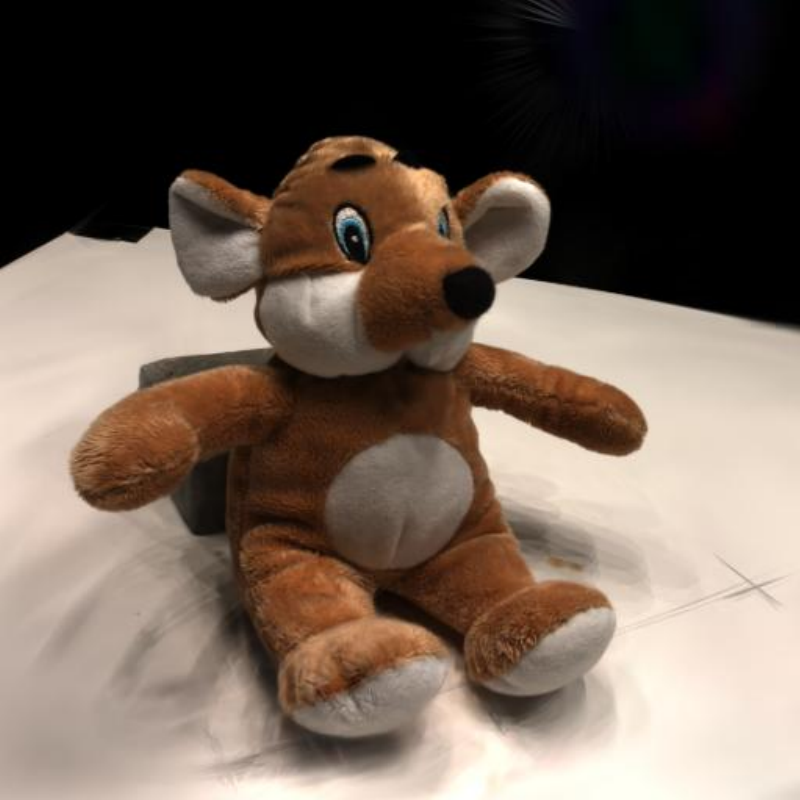}%
            \hspace{-0.5pt}%
            \includegraphics[width=0.45\textwidth]{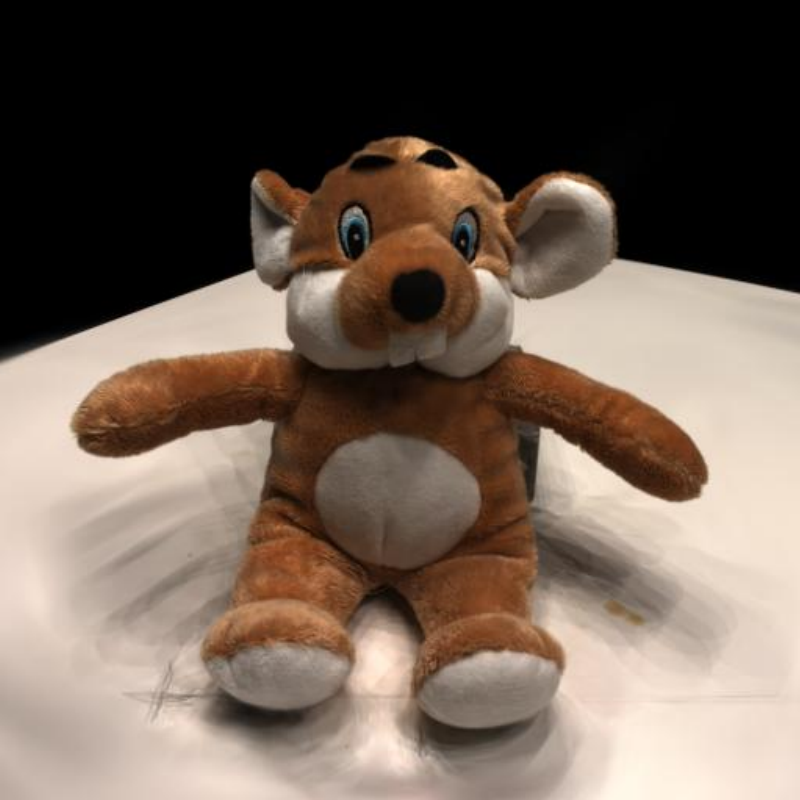}
        \end{minipage}
    \end{subfigure}
    \begin{subfigure}{0.28\textwidth}
        \centering
        \begin{minipage}{\textwidth}
            \includegraphics[width=0.45\textwidth]{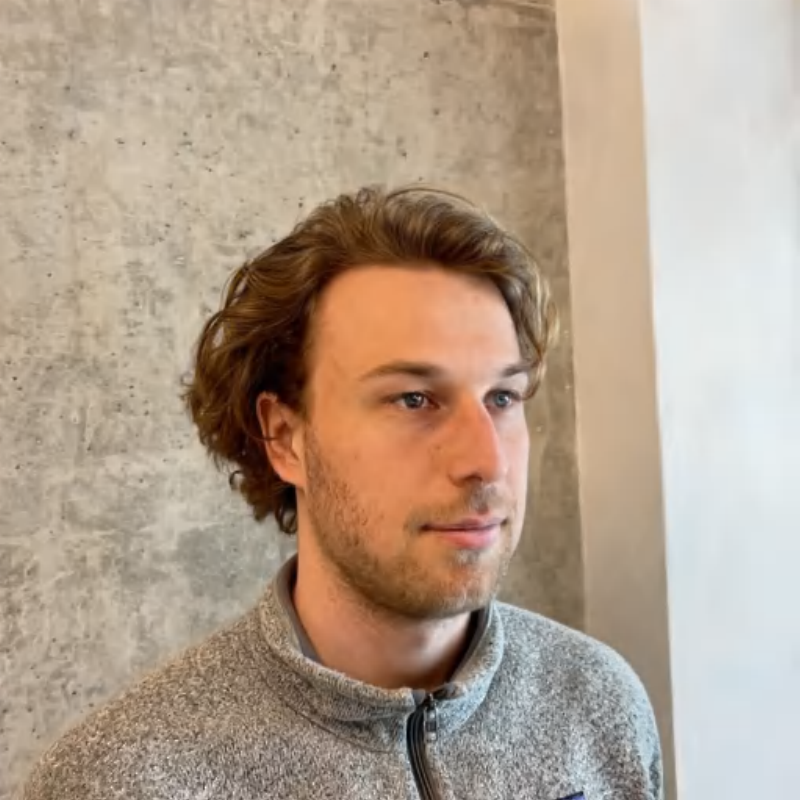}%
            \hspace{-0.5pt}%
            \includegraphics[width=0.45\textwidth]{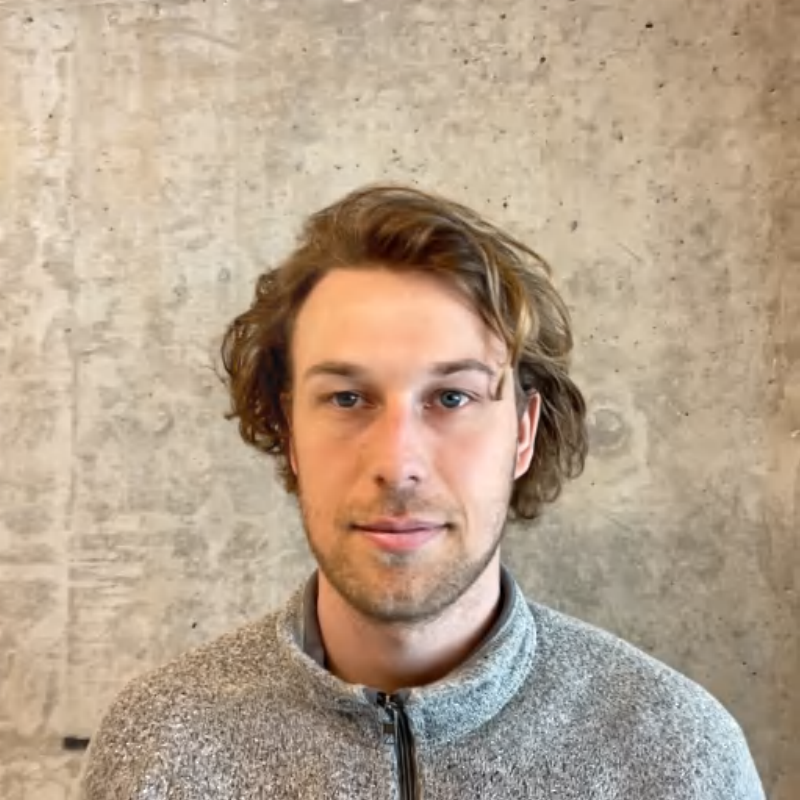}
        \end{minipage}
    \end{subfigure}
    \begin{subfigure}{0.28\textwidth}
        \centering
        \begin{minipage}{\textwidth}
            \includegraphics[width=0.45\textwidth]{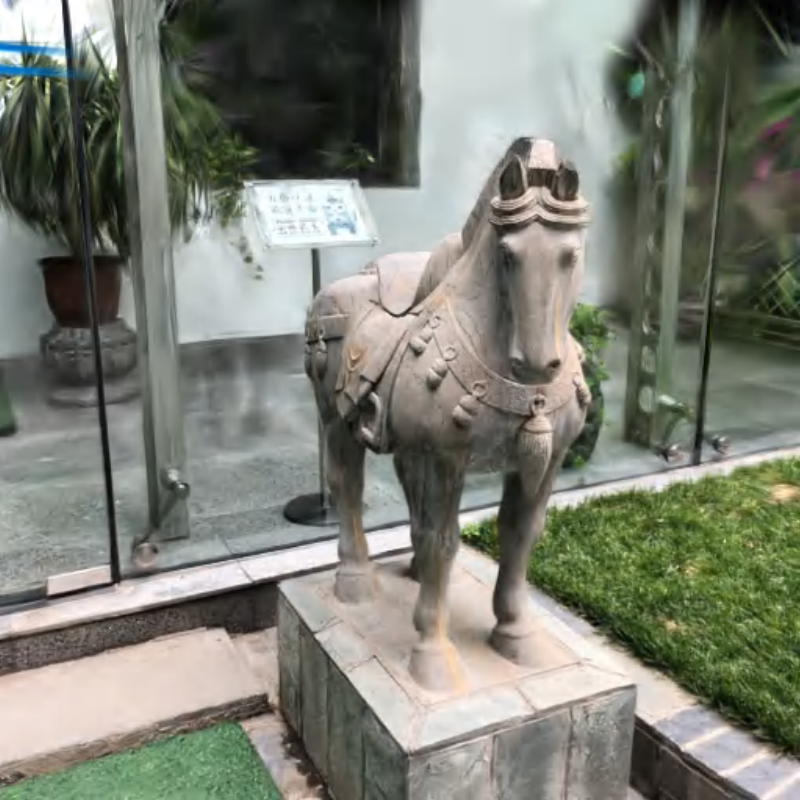}%
            \hspace{-0.5pt}%
            \includegraphics[width=0.45\textwidth]{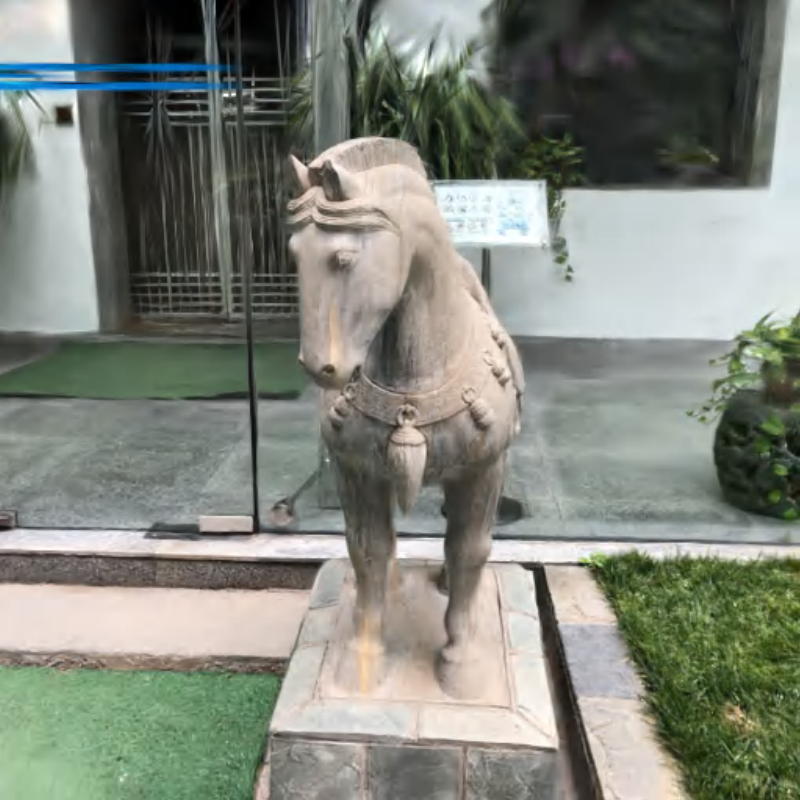}
        \end{minipage}
    \end{subfigure}
    \noindent\makebox[\textwidth]{\dotfill}
    \begin{minipage}{0.12\textwidth}
        \raggedright
        \textbf{IN2N(GS)}
    \end{minipage}  
    \begin{subfigure}{0.28\textwidth}
        \centering
        \begin{minipage}{\textwidth}
            \includegraphics[width=0.45\textwidth]{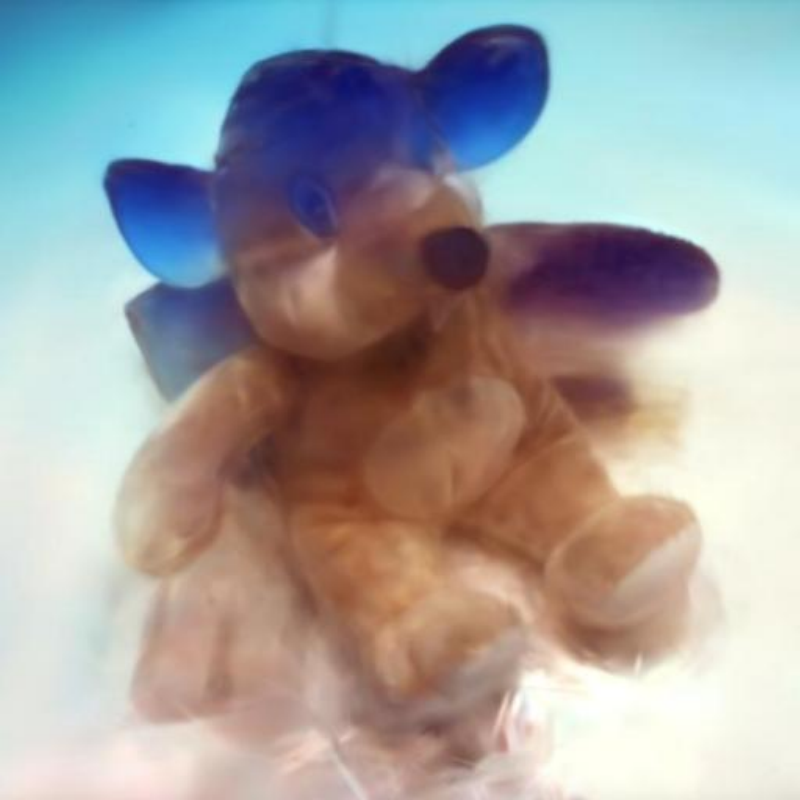}%
            \hspace{-0.5pt}%
            \includegraphics[width=0.45\textwidth]{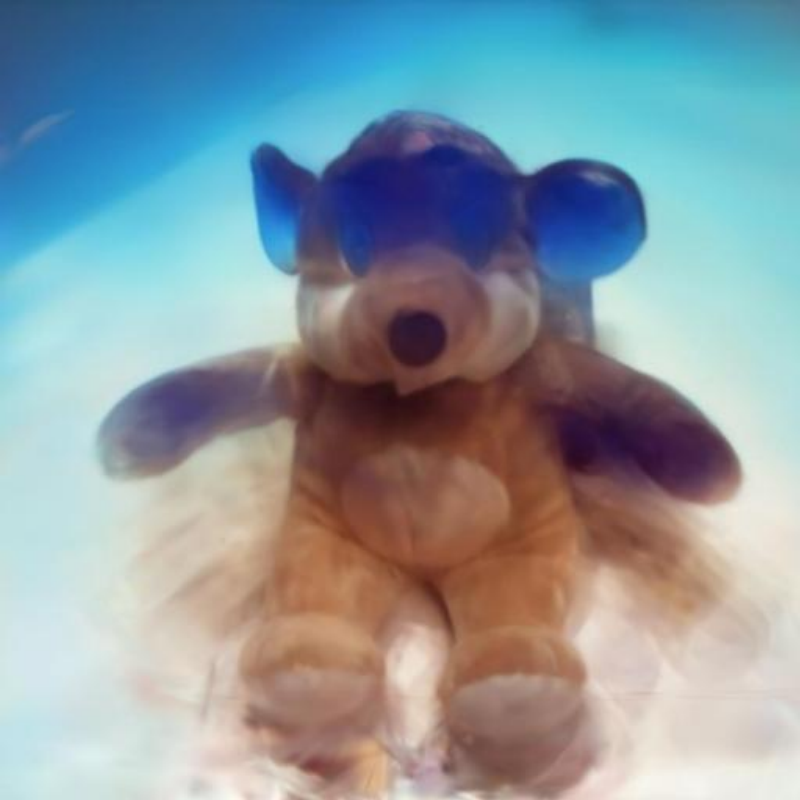}
        \end{minipage}
    \end{subfigure}
    \begin{subfigure}{0.28\textwidth}
        \centering
        \begin{minipage}{\textwidth}
            \includegraphics[width=0.45\textwidth]{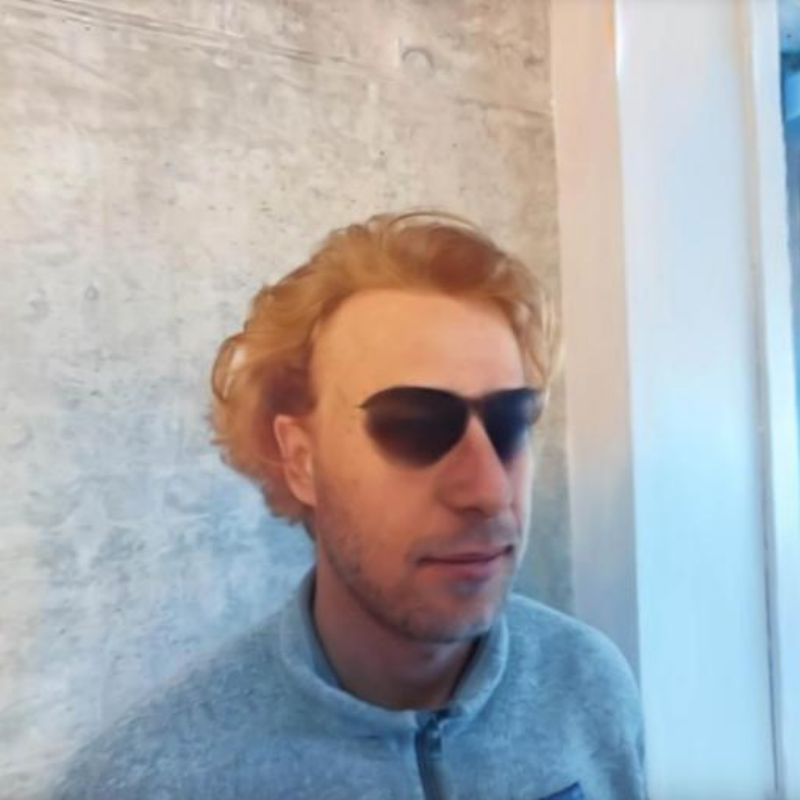}%
            \hspace{-0.5pt}%
            \includegraphics[width=0.45\textwidth]{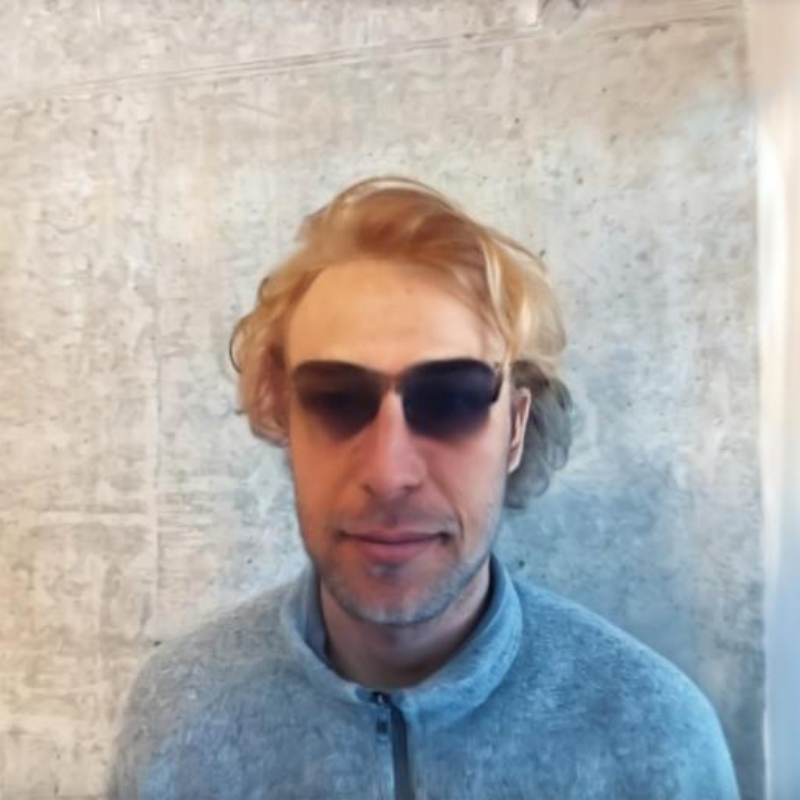}
        \end{minipage}
    \end{subfigure}
    \begin{subfigure}{0.28\textwidth}
        \centering
        \begin{minipage}{\textwidth}
            \includegraphics[width=0.45\textwidth]{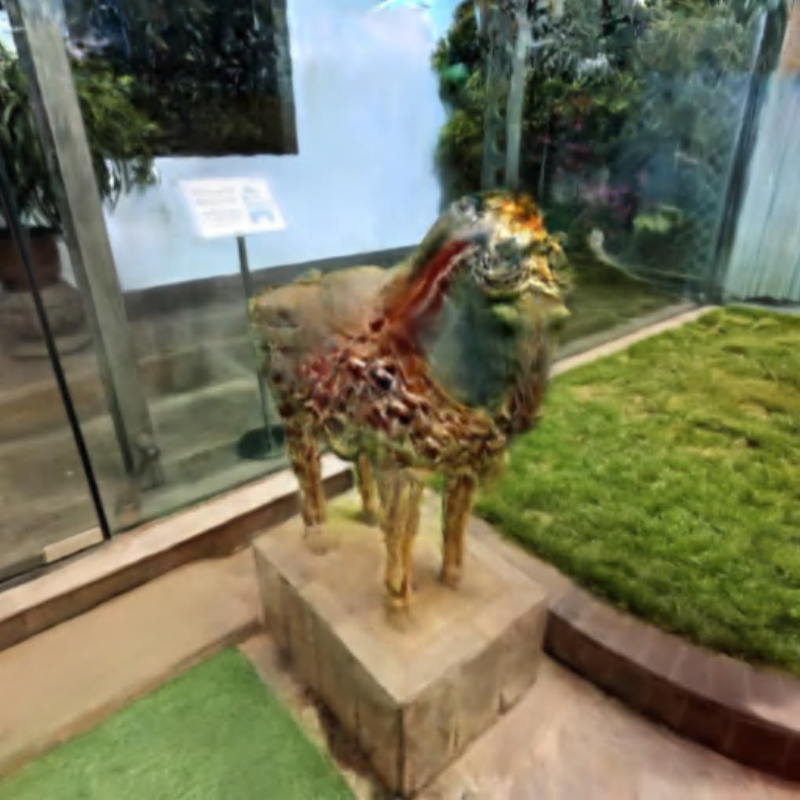}%
            \hspace{-0.5pt}%
            \includegraphics[width=0.45\textwidth]{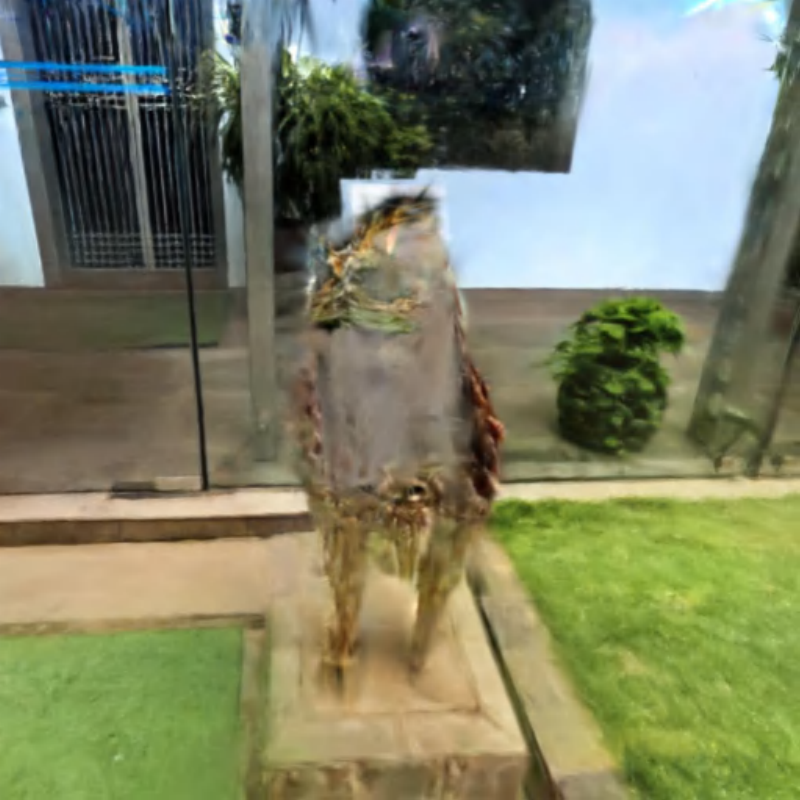}
        \end{minipage}
    \end{subfigure}

    \hspace*{0.12\textwidth} 
    \begin{minipage}{0.28\textwidth}
      \centering \small \textit{``Give the doll a pair of glasses''}
    \end{minipage}
    \begin{minipage}{0.28\textwidth}
      \centering \small \textit{``Give the man a pair of glasses''}
    \end{minipage}
    \begin{minipage}{0.28\textwidth}
      \centering \small \textit{``Turn the stone horse into a giraffe''}
    \end{minipage}
    
    \begin{minipage}{0.12\textwidth}
        \raggedright
        \textbf{GaussCtrl}
    \end{minipage} 
    \begin{subfigure}{0.28\textwidth}
        \centering
        \begin{minipage}{\textwidth}
            \includegraphics[width=0.45\textwidth]{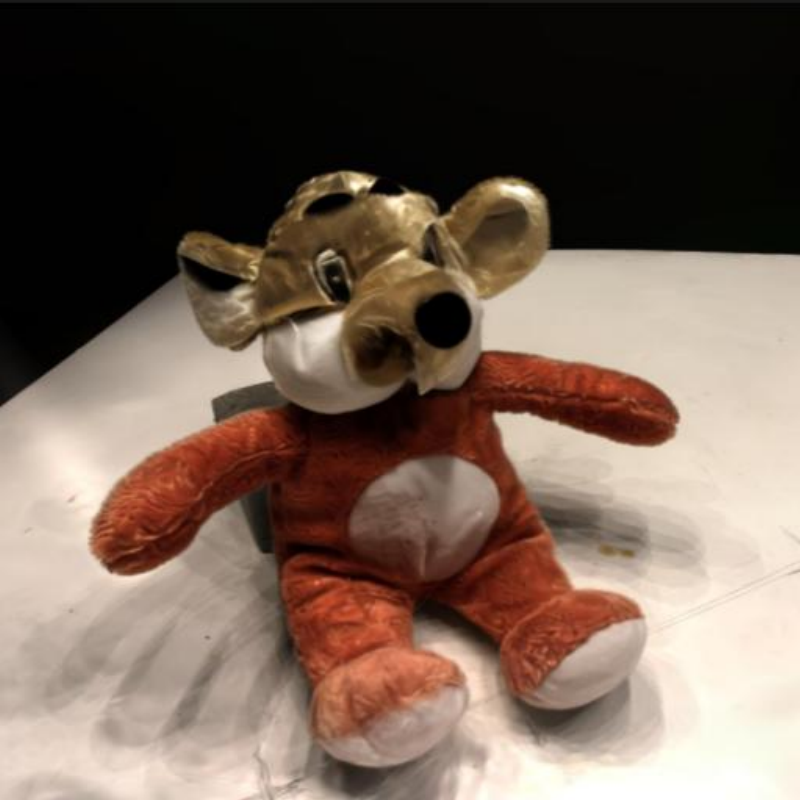}%
            \hspace{-0.5pt}%
            \includegraphics[width=0.45\textwidth]{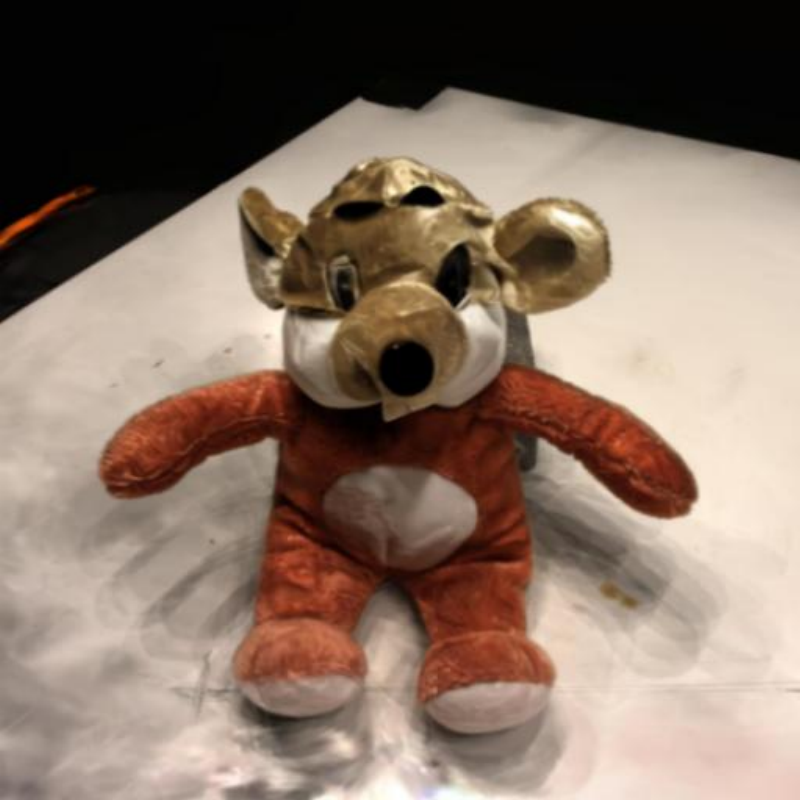}
        \end{minipage}
    \end{subfigure}
    \begin{subfigure}{0.28\textwidth}
        \centering
        \begin{minipage}{\textwidth}
            \includegraphics[width=0.45\textwidth]{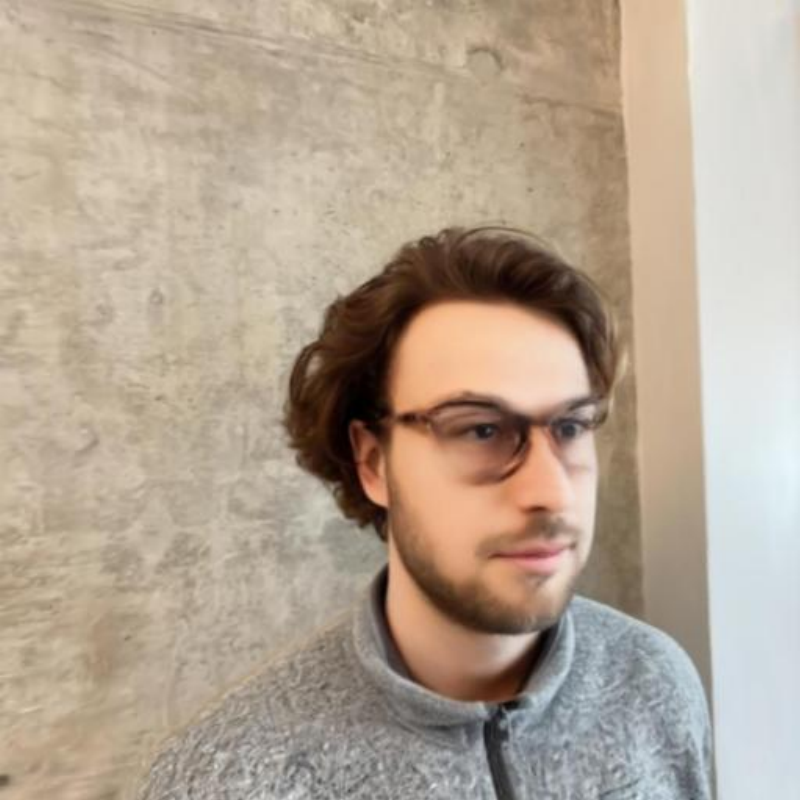}%
            \hspace{-0.5pt}%
            \includegraphics[width=0.45\textwidth]{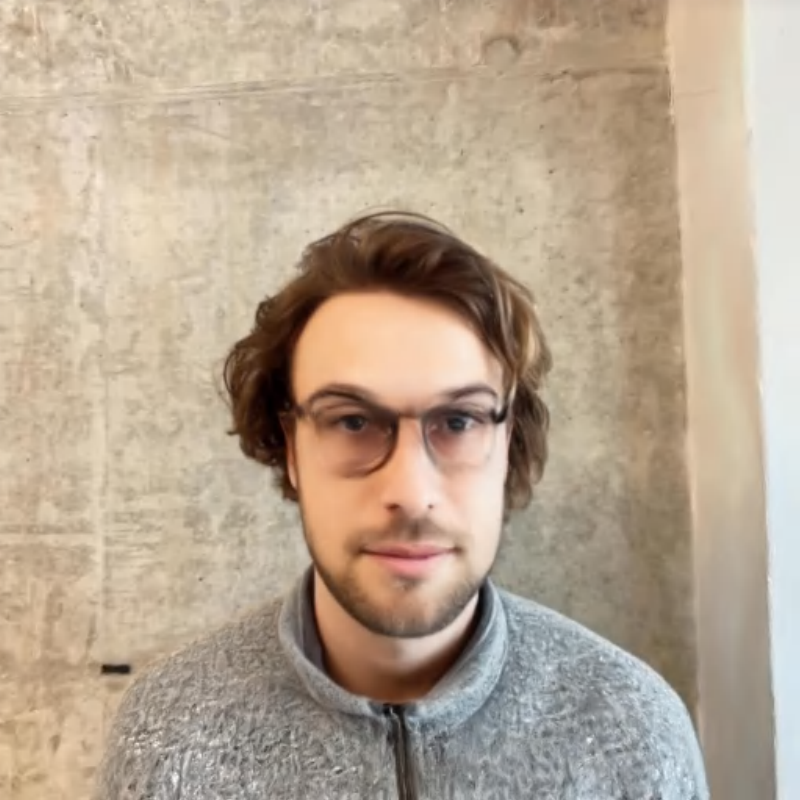}
        \end{minipage}
    \end{subfigure}
    \begin{subfigure}{0.28\textwidth}
        \centering
        \begin{minipage}{\textwidth}
            \includegraphics[width=0.45\textwidth]{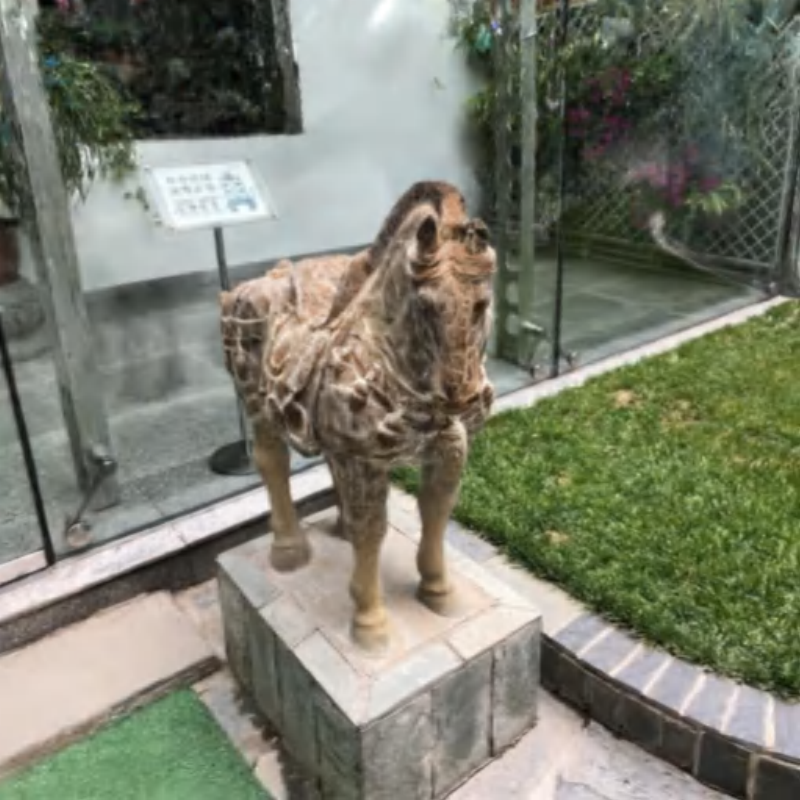}%
            \hspace{-0.5pt}%
            \includegraphics[width=0.45\textwidth]{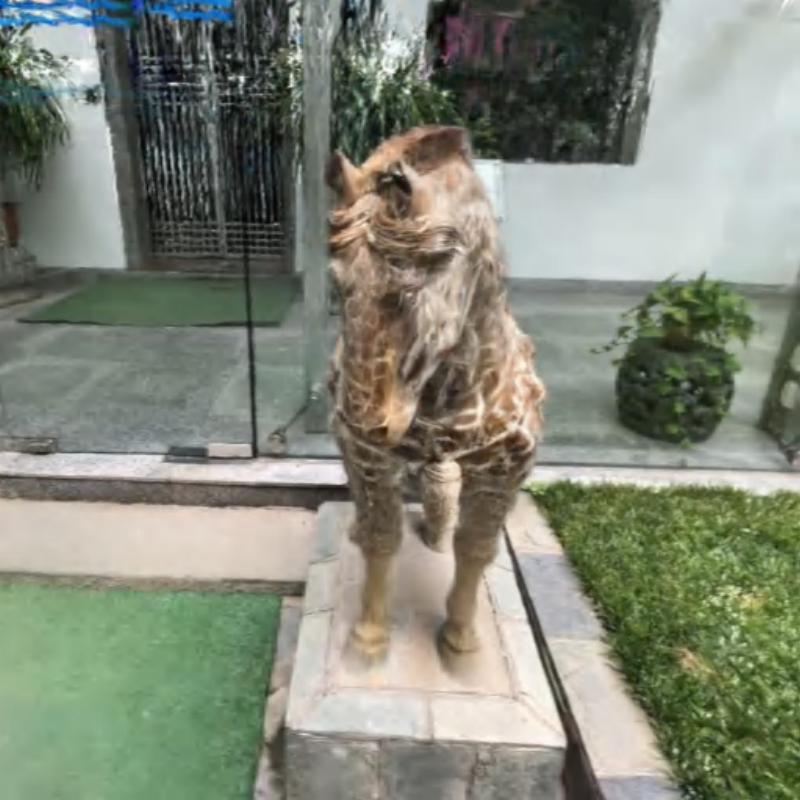}
        \end{minipage}
    \end{subfigure}

    \hspace*{0.12\textwidth} 
    \begin{minipage}{0.28\textwidth}
      \centering \small \textit{``A photo of a doll wearing a pair of glasses''}
    \end{minipage}
    \begin{minipage}{0.28\textwidth}
      \centering \small \textit{``A photo of a man wearing a pair of glasses''}
    \end{minipage}
    \begin{minipage}{0.28\textwidth}
      \centering \small \textit{``A photo of a giraffe in front of the museum''}
    \end{minipage}
    
    \begin{minipage}{0.12\textwidth}
        \raggedright
        \textbf{GaussianEditor}
    \end{minipage}  
    \begin{subfigure}{0.28\textwidth}
        \centering
        \begin{minipage}{\textwidth}
            \includegraphics[width=0.45\textwidth]{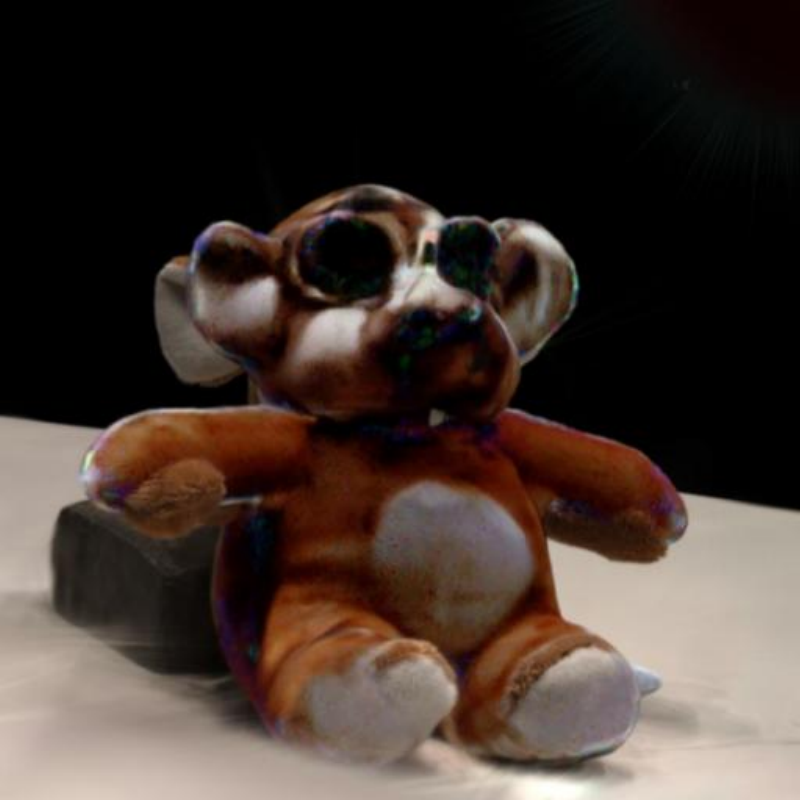}%
            \hspace{-0.5pt}%
            \includegraphics[width=0.45\textwidth]{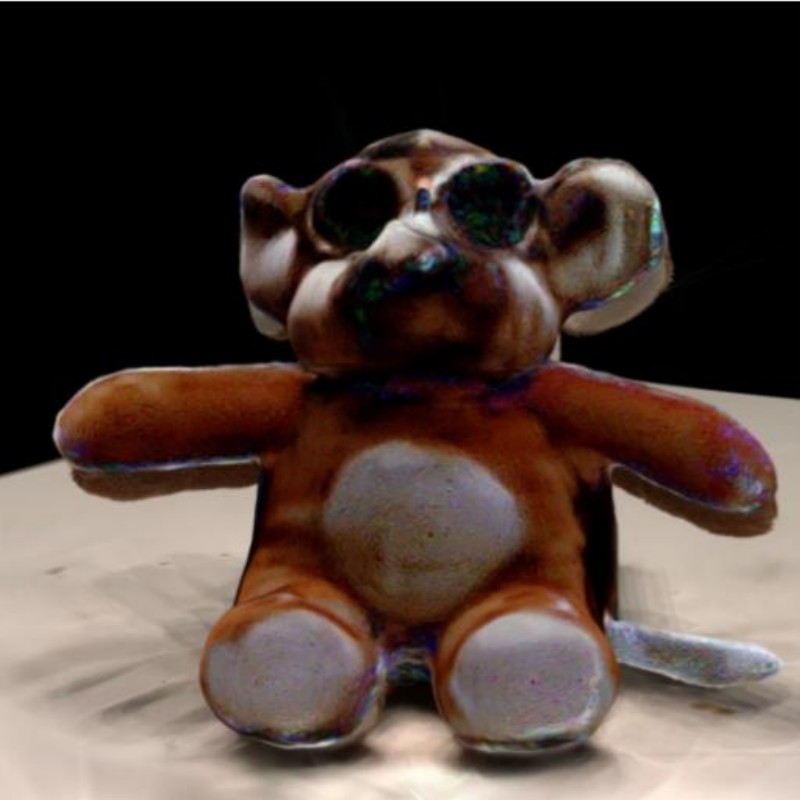}
        \end{minipage}
    \end{subfigure}
    \begin{subfigure}{0.28\textwidth}
        \centering
        \begin{minipage}{\textwidth}
            \includegraphics[width=0.45\textwidth]{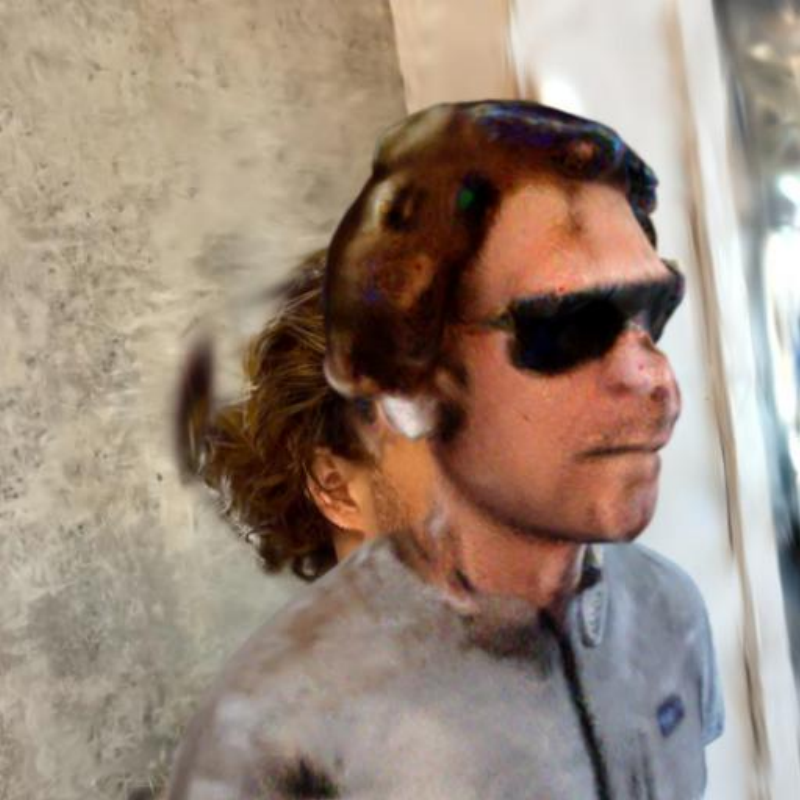}%
            \hspace{-0.5pt}%
            \includegraphics[width=0.45\textwidth]{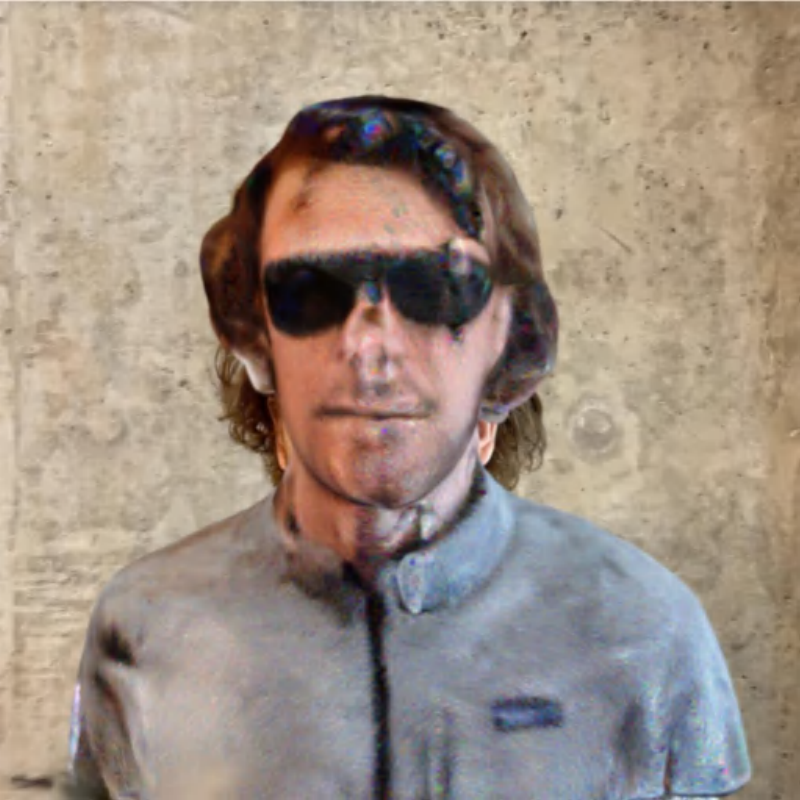}
        \end{minipage}
    \end{subfigure}
    \begin{subfigure}{0.28\textwidth}
        \centering
        \begin{minipage}{\textwidth}
            \includegraphics[width=0.45\textwidth]{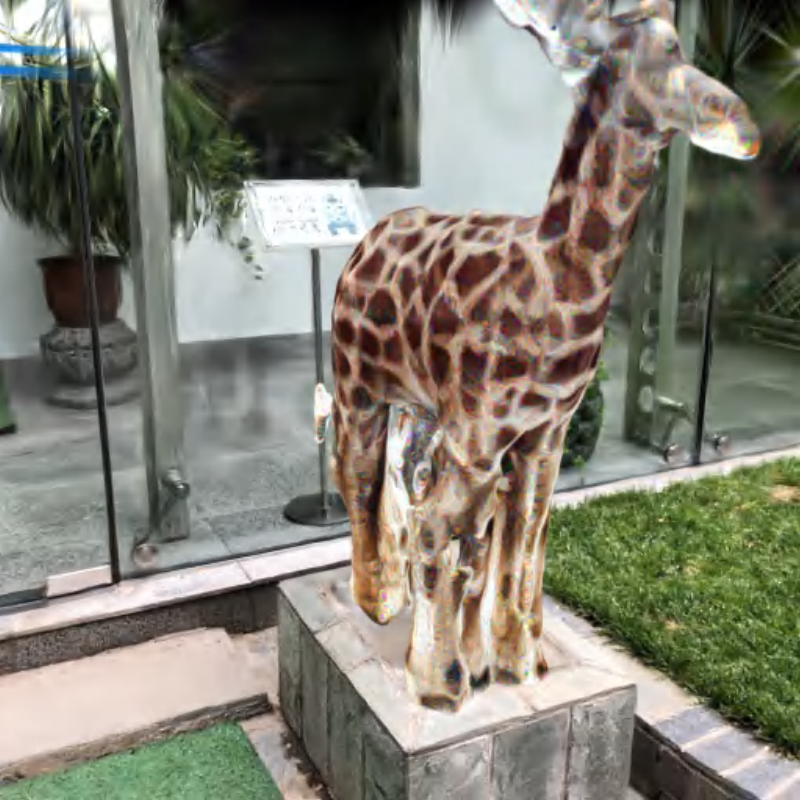}%
            \hspace{-0.5pt}%
            \includegraphics[width=0.45\textwidth]{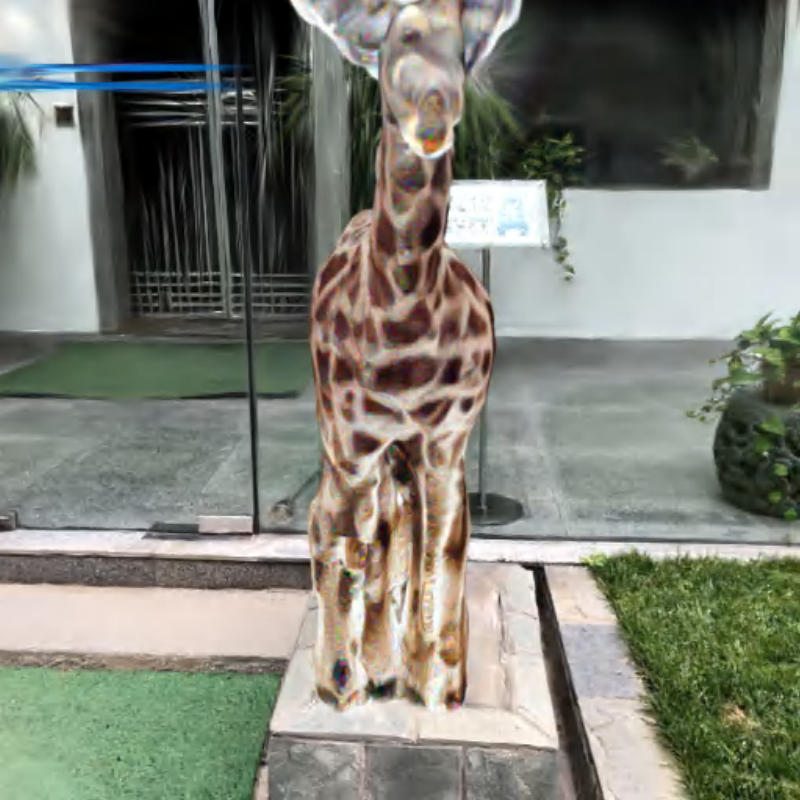}
        \end{minipage}
    \end{subfigure}

    \hspace*{0.12\textwidth} 
    \begin{minipage}{0.28\textwidth}
      \centering \small \textit{``A doll wearing a pair of glasses''}
    \end{minipage}
    \begin{minipage}{0.28\textwidth}
      \centering \small \textit{``A man wearing a pair of glasses''}
    \end{minipage}
    \begin{minipage}{0.28\textwidth}
      \centering \small \textit{``A giraffe standing on the stone platform''}
    \end{minipage}
    
    \begin{minipage}{0.12\textwidth}
        \raggedright
        \textbf{\name}
    \end{minipage}  
    \begin{subfigure}{0.28\textwidth}
        \centering
        \begin{minipage}{\textwidth}
            \includegraphics[width=0.45\textwidth]{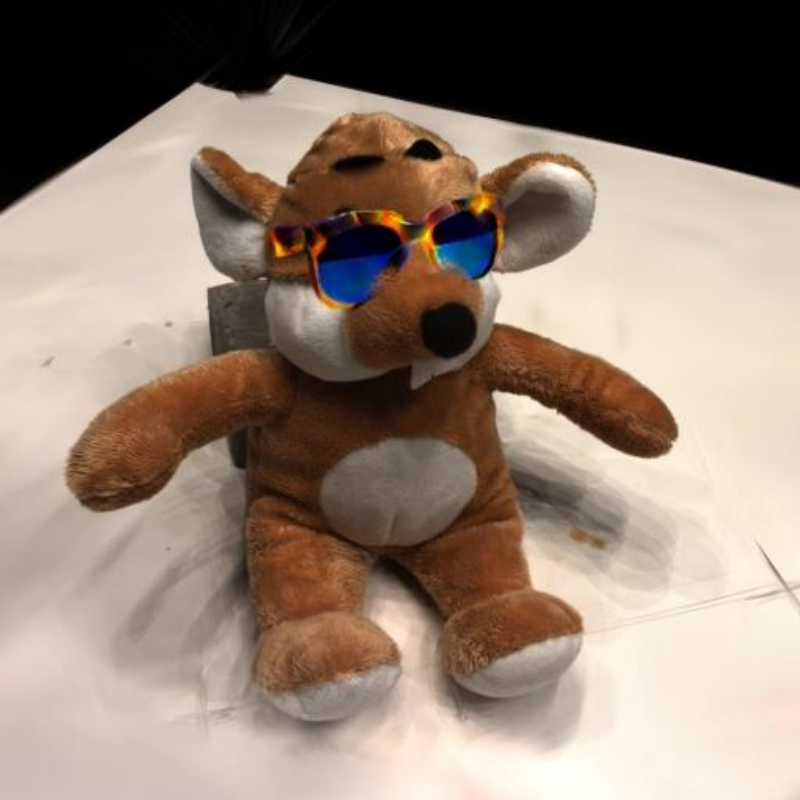}%
            \includegraphics[width=0.45\textwidth]{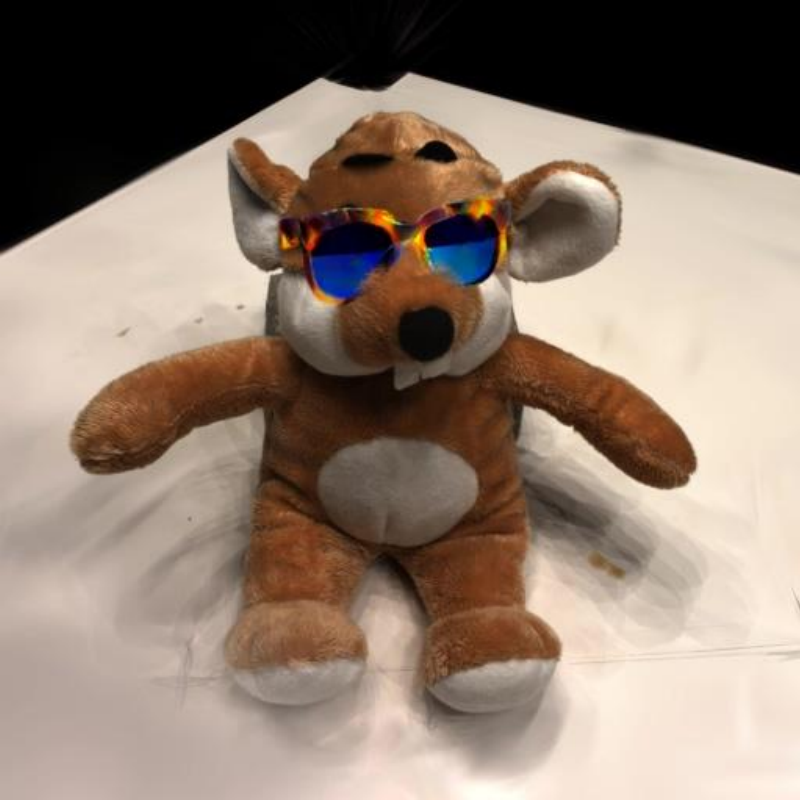}
        \end{minipage}
    \end{subfigure}
    \begin{subfigure}{0.28\textwidth}
        \centering
        \begin{minipage}{\textwidth}
            \includegraphics[width=0.45\textwidth]{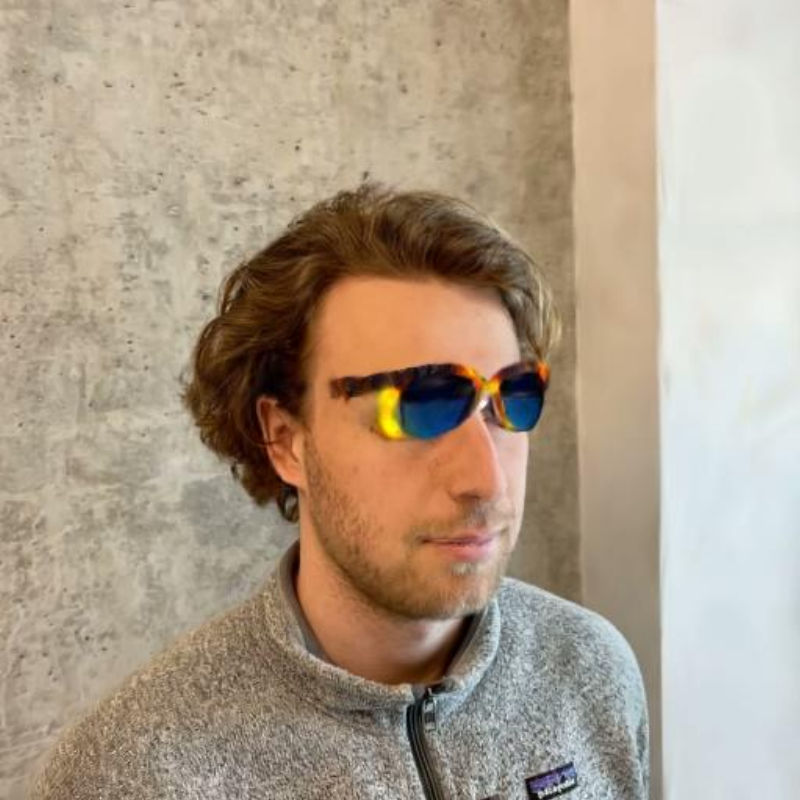}%
            \includegraphics[width=0.45\textwidth]{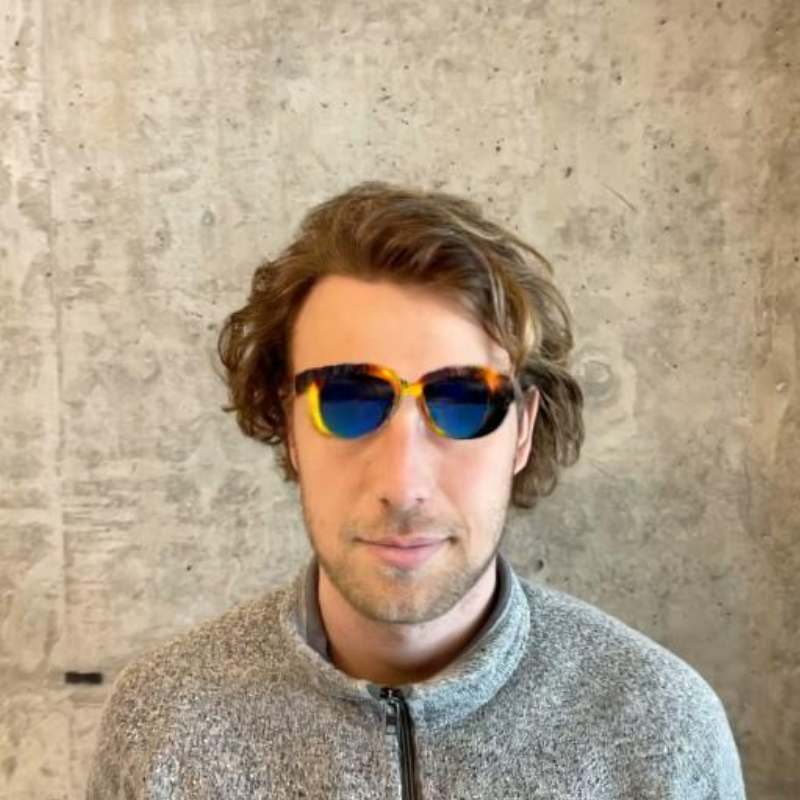}
        \end{minipage}
    \end{subfigure}
    \begin{subfigure}{0.28\textwidth}
        \centering
        \begin{minipage}{\textwidth}
            \includegraphics[width=0.45\textwidth]{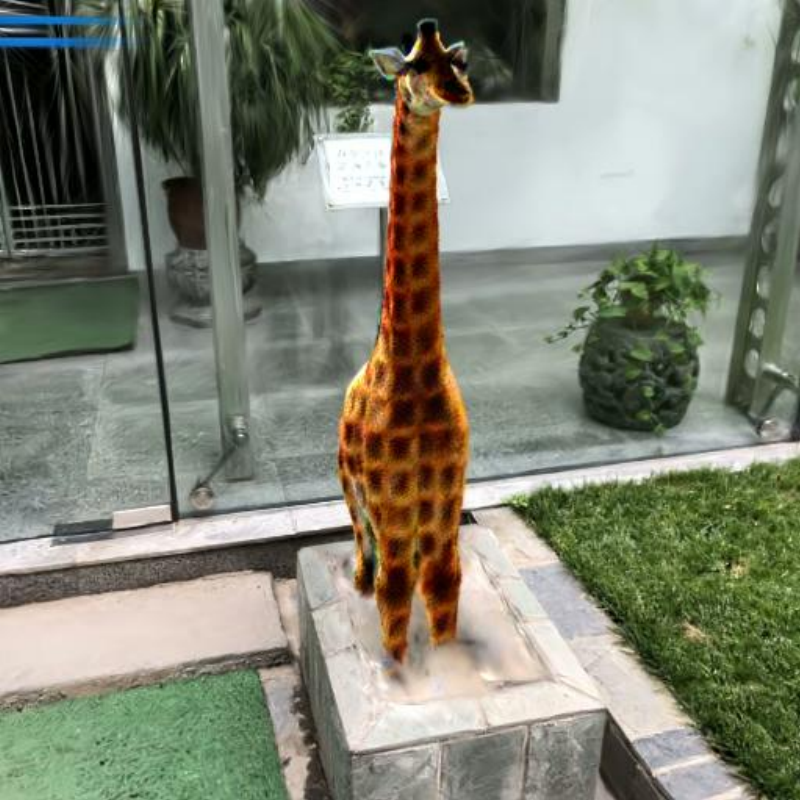}%
            \includegraphics[width=0.45\textwidth]{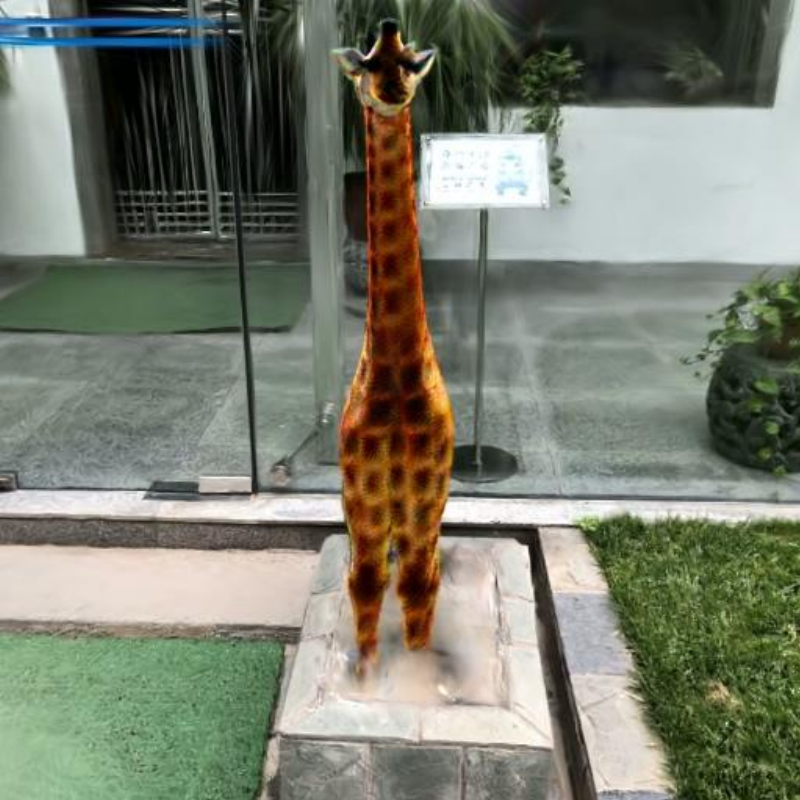}
        \end{minipage}
    \end{subfigure}
    \vspace{2pt} 

    \hspace*{0.12\textwidth} 
    \begin{minipage}{0.28\textwidth}
      \centering \small \textit{``Add a pair of glasses to the doll''}
    \end{minipage}
    \begin{minipage}{0.28\textwidth}
      \centering \small \textit{``Add a pair of glasses to the man''}
    \end{minipage}
    \begin{minipage}{0.28\textwidth}
      \centering \small \textit{``Add a giraffe to replace the stone horse''}
    \end{minipage}
    
    \captionsetup{skip=2pt}
    \caption{
    Visual comparison with state-of-the-art methods for text-guided object insertion (Cols 1–2) and replacement (Col 3).
    Our method generates higher-quality results while preserving scene integrity. IN2N (GS) and GaussCtrl sometimes misunderstand the prompt and fail to complete insertion (e.g., “Give the doll a pair of glasses to the doll”), and struggle to produce clear shape changes in replacement (Col 3, Rows 2–3). GaussianEditor requires manual masks and depth adjustment, and suffers from artifacts and low-quality objects due to post-inpainting and 3D reconstruction limitations.
    }
    \label{fig:comparision with advanced}
\end{figure*}

\subsection{Experiments Setup.}
\noindent\textbf{Implementation Details} 
During initialization, we use a learning rate of \(5 \times 10^{-3}\) for optimizing both \( \mathcal{G}_{\textit{AR}} \) and \( \textit{t}_{\textit{O}} \) of inserted object.
When estimating the coarse rotation $\textit{r}_{\textit{O}}$, we render the object at 10-degree intervals .
During the Hierarchical Spatial Aware Refinement stage, we apply a learning rate of \(5 \times 10^{-4}\) with diffusion timesteps in the range of [0.02, 0.2], $\lambda = 0.1$ is set in \(\mathcal{L}_{loc}\) and $\beta$ is linearly increased from 0 to 1 during training. 
For appearance refinement, we optimize the object appearance using timesteps in [0.02, 0.5$\sim$0.25]. 
The object image \(\mathcal{I}_{\textit{O}}\) is upsampled with a sampling ratio of \(M / N = 3\) relative to multi-view inputs. 
All experiments are conducted on a single NVIDIA A40 GPU. 
More details are provided in the Appendix.

\noindent\textbf{Dataset} To comprehensively evaluate our method, we follow prior works~\cite{chen2024gaussianeditor,zhuang2024tip,haque2023instruct} and select representative scenes of varying complexity, including simple backgrounds, human faces, and complex outdoor environments. 
In these scenes, we insert commonly associated objects (e.g., glasses, giraffes) and evaluate diverse categories such as bowties and moustaches to assess generalization. For GaussianEditor~\cite{chen2024gaussianeditor}, we manually annotate masks, while for TIP-Editor~\cite{zhuang2024tip}, we use the author-provided bounding boxes and object images for comparison.

\noindent\textbf{Baselines} We compare our method with state-of-the-art 3D scene editing approaches that support object insertion and replacement, under two types of guidance: text prompt and text-image prompt.The text-guided baselines include three methods: Instruct-GS2GS \cite{vachhainstruct}, which extends Instruct-NeRF2NeRF (IN2N) \cite{haque2023instruct} by replacing the NeRF in IN2N with a 3DGS model; GaussCtrl \cite{wu2024gaussctrl}, and GaussianEditor \cite{chen2024gaussianeditor}.
For text-image prompt methods, we compare with TIP-Editor~\cite{zhuang2024tip}, which uses an example image to specify object appearance. 
As TIP-Editor provides only limited insertion scripts (e.g., ``A doll wearing sunglasses'', ``A man with beard'').
For fairness, we use official code and pre-trained weights.


\begin{figure*}[t]

    \hspace{0.02\textwidth}
    \begin{subfigure}{0.145\textwidth}
    \centering
        \textbf{Original/Ref} 

      \begin{minipage}{\textwidth}
        \begin{overpic}[width=0.9\linewidth]{figures/scene/doll/1.3_75_0.pdf}
          \put(65,-5){\includegraphics[width=0.35\linewidth]{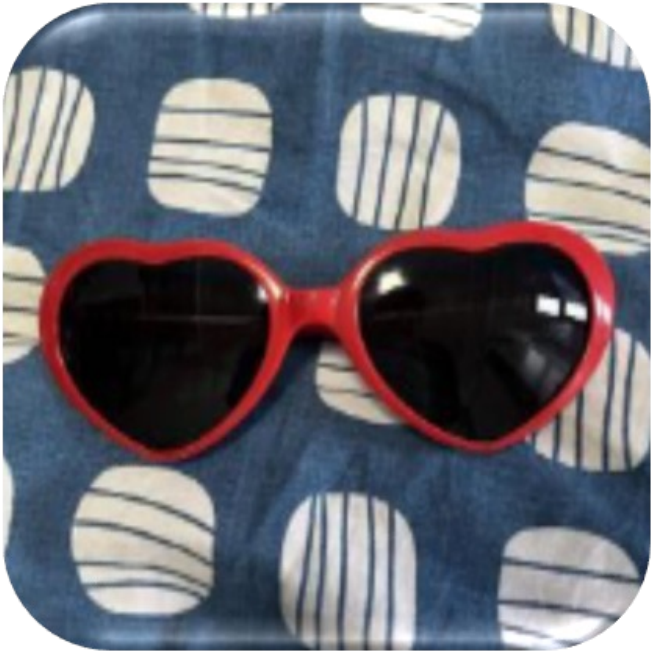}}
        \end{overpic}
      \end{minipage}
    \end{subfigure}
    \begin{subfigure}{0.40\textwidth}
        \centering
        \textbf{TIP-Editor} 
        \begin{minipage}{\textwidth}
            \includegraphics[width=0.3\textwidth]{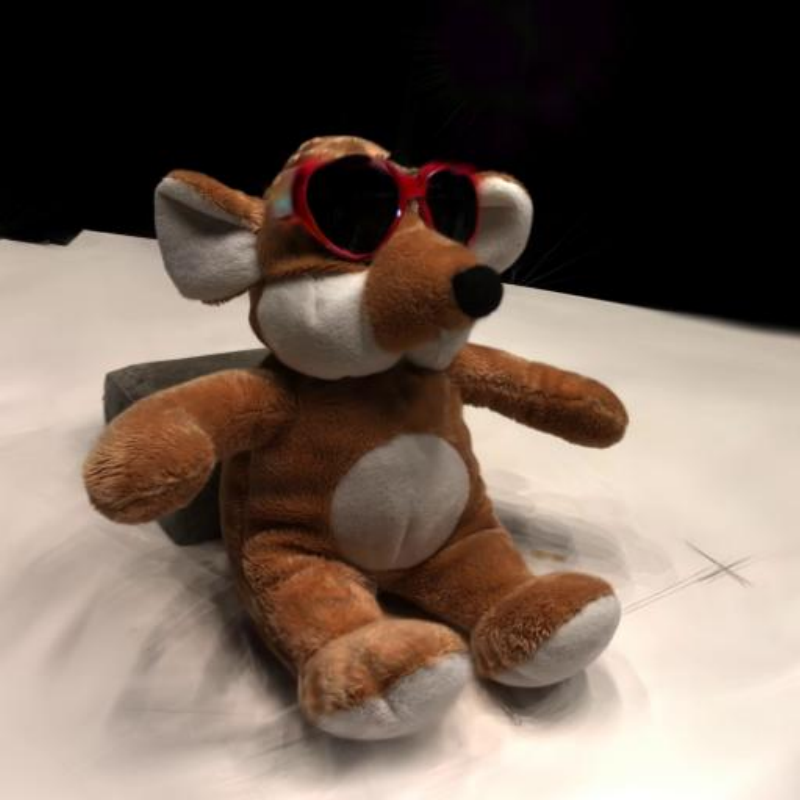}%
            \includegraphics[width=0.3\textwidth]{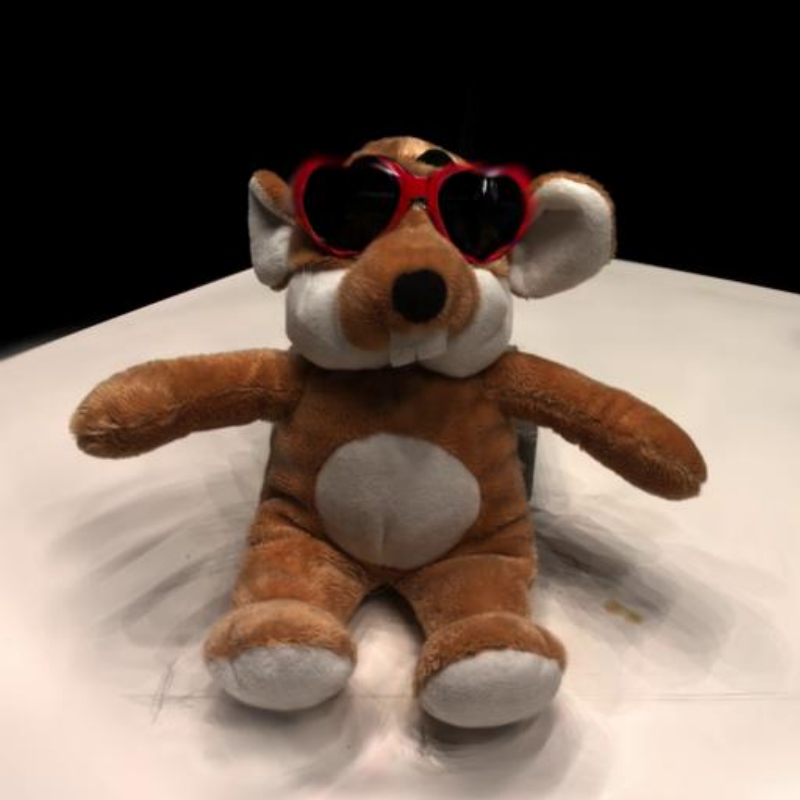}%
            \includegraphics[width=0.3\textwidth]{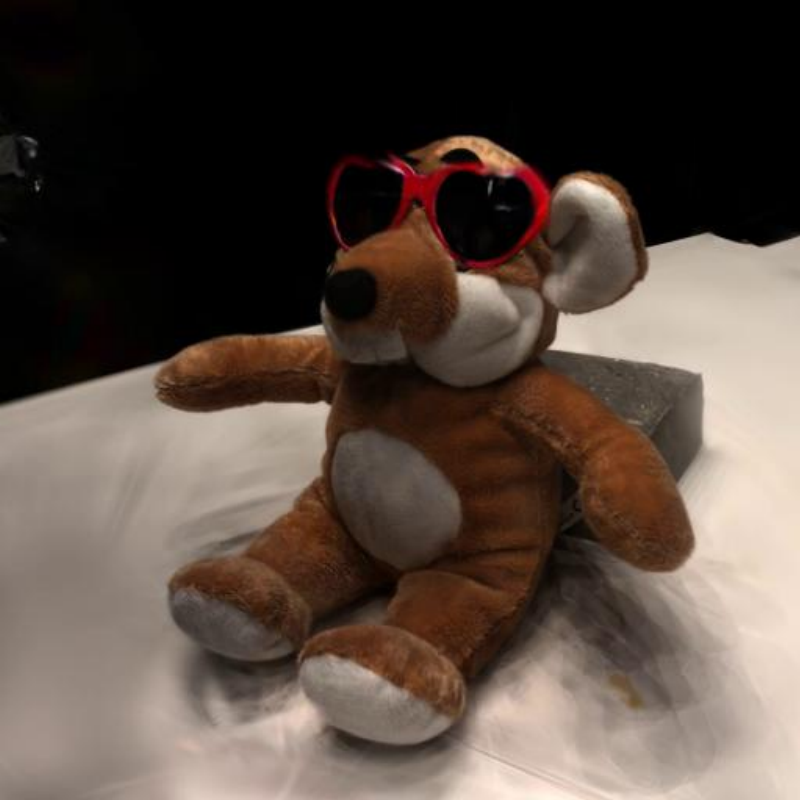}
        \end{minipage}
    \end{subfigure}
    \hspace{-0.03\textwidth}
    \begin{subfigure}{0.40\textwidth}
        \centering
        \textbf{\name} 
        \begin{minipage}{\textwidth}
            \includegraphics[width=0.3\textwidth]{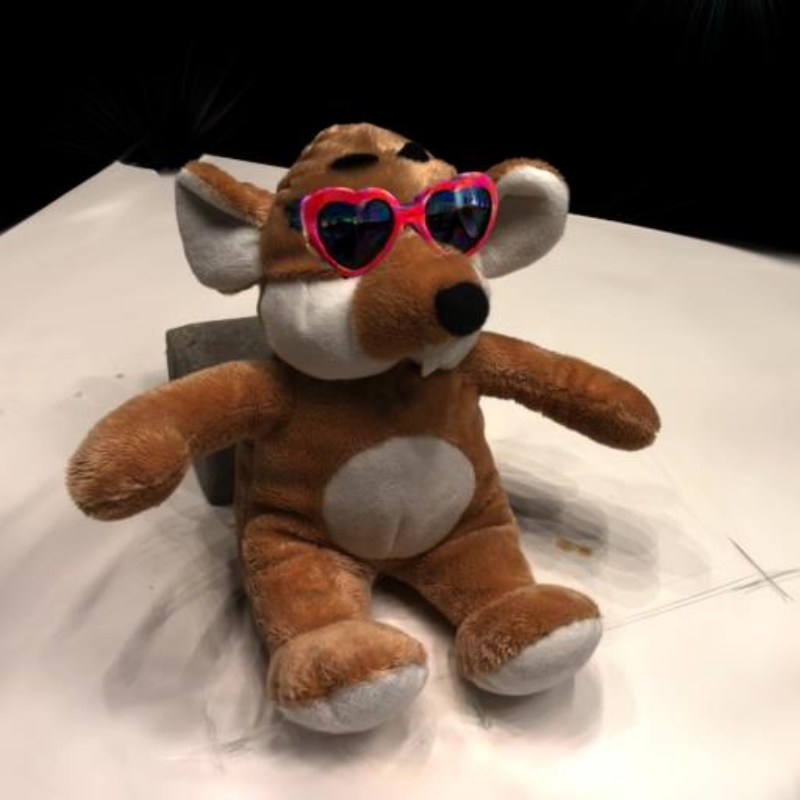}%
            \includegraphics[width=0.3\textwidth]{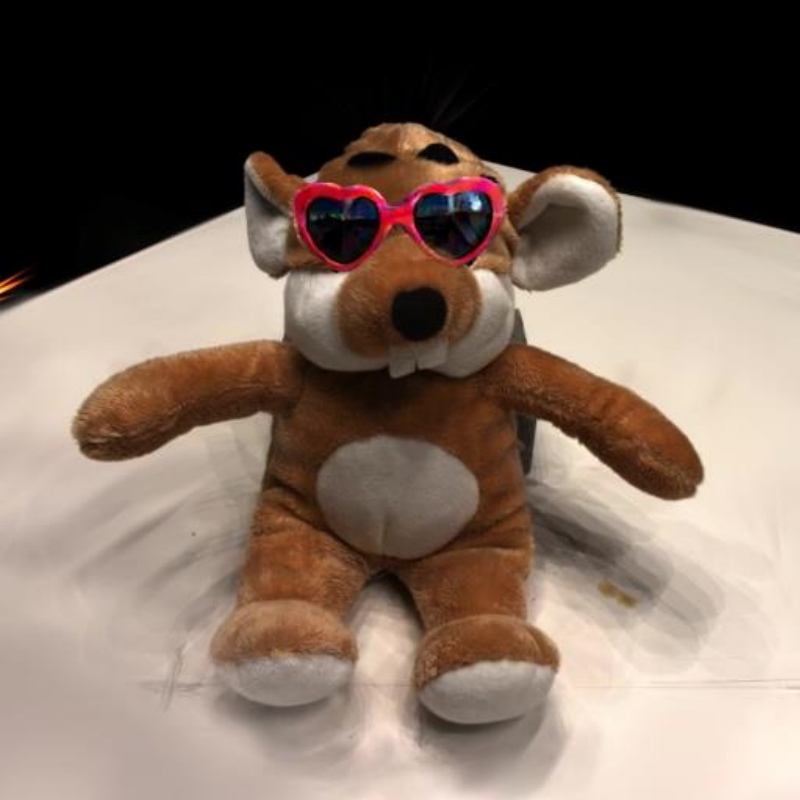}%
            \includegraphics[width=0.3\textwidth]{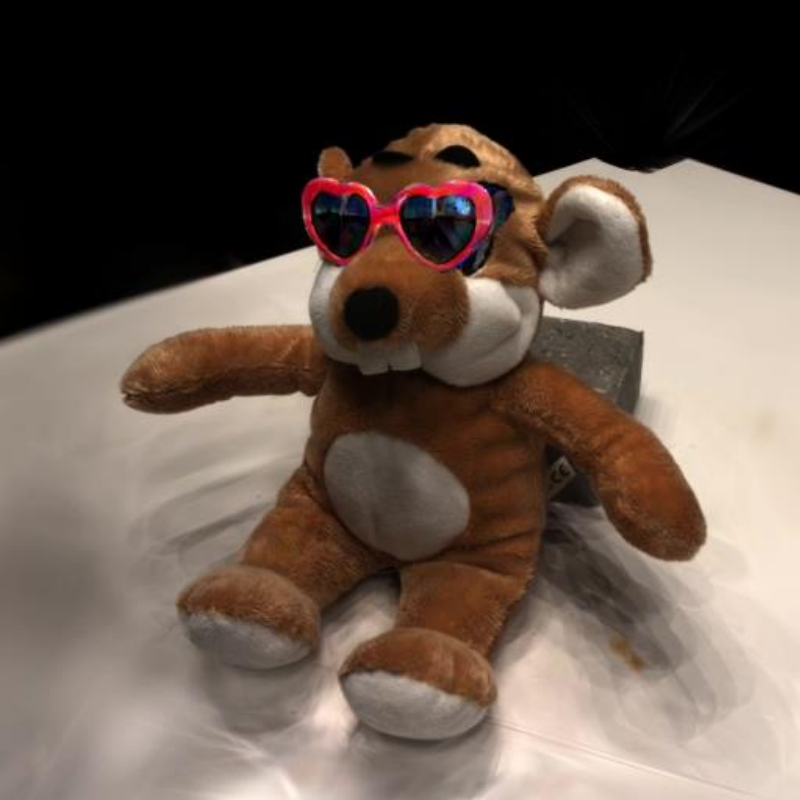}
        \end{minipage}
    \end{subfigure}
    \hspace*{0.145\textwidth} 
    \begin{minipage}{0.4\textwidth}
      \centering \small \textit{``A doll wearing a pair of $V_1$ sunglasses''}
    \end{minipage}
    \hspace{-0.03\textwidth}
    \begin{minipage}{0.4\textwidth}
      \centering \small \textit{``Add a pair of glasses to the doll''}
    \end{minipage}

    \hspace{0.02\textwidth}
    \begin{subfigure}{0.145\textwidth}
    \centering

      \begin{minipage}{\textwidth}
        \begin{overpic}[width=0.90\linewidth]{figures/scene/doll/1.3_75_0.pdf}
          \setlength{\fboxsep}{0pt}
            \put(65,-6){\includegraphics[width=0.35\linewidth]{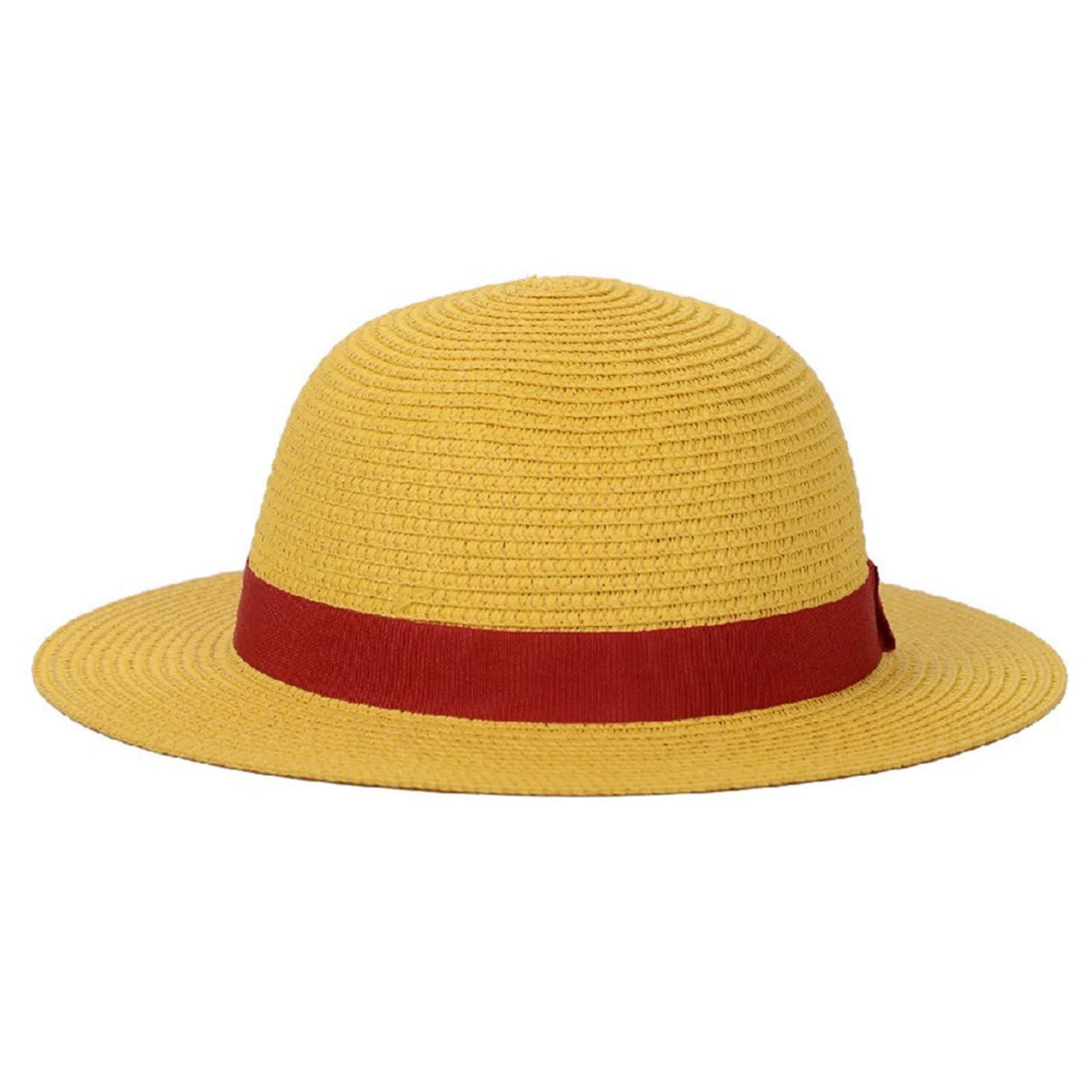}}

        \end{overpic}
      \end{minipage}
    \end{subfigure}
    \centering
    \begin{subfigure}{0.4\textwidth}
        \centering
        \begin{minipage}{\textwidth}   
            \includegraphics[width=0.3\textwidth]{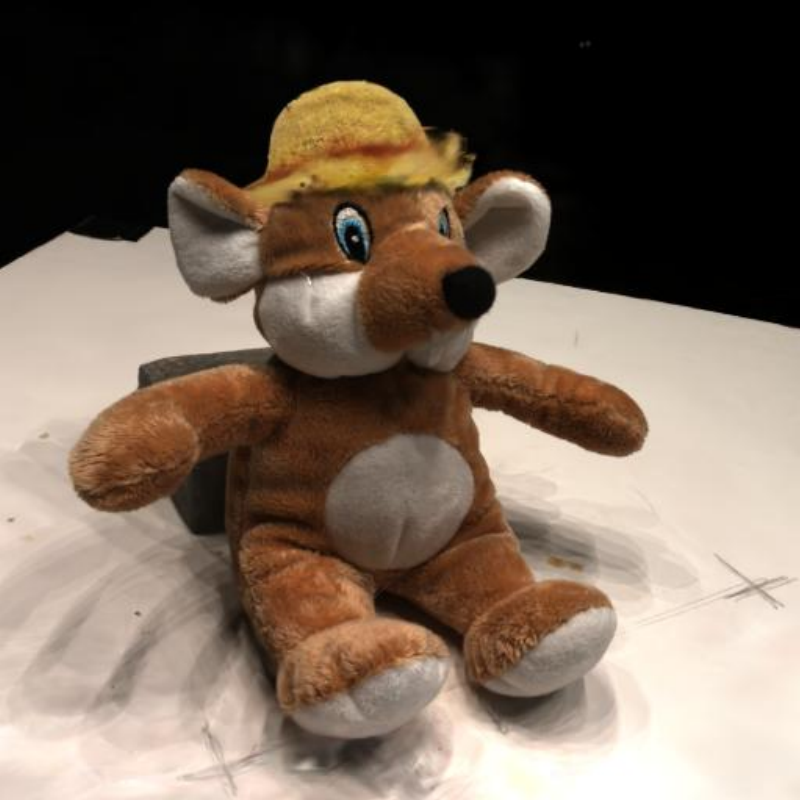}%
            \includegraphics[width=0.3\textwidth]{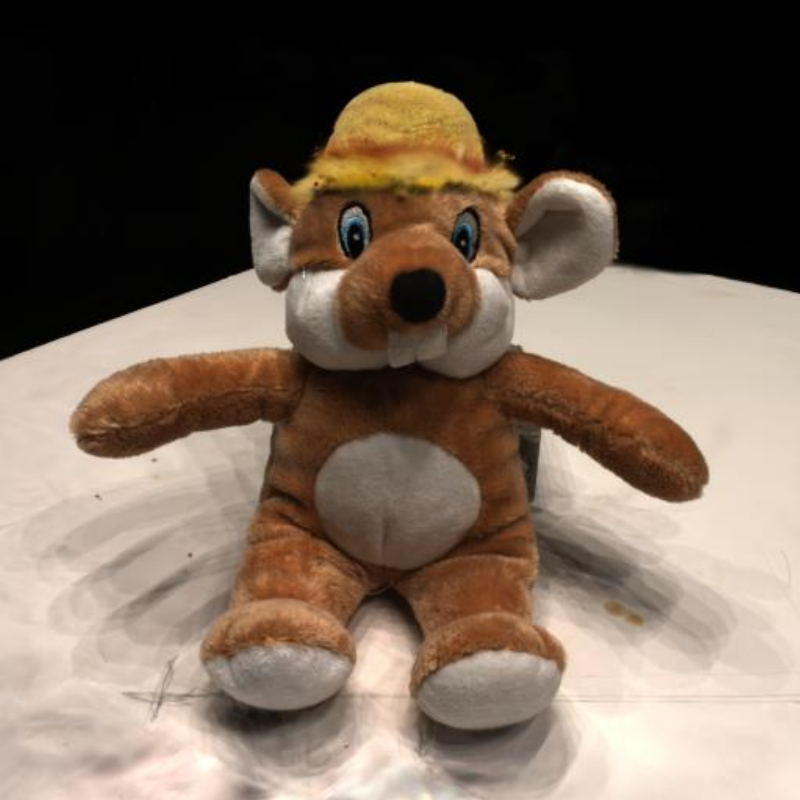}%
            \includegraphics[width=0.3\textwidth]{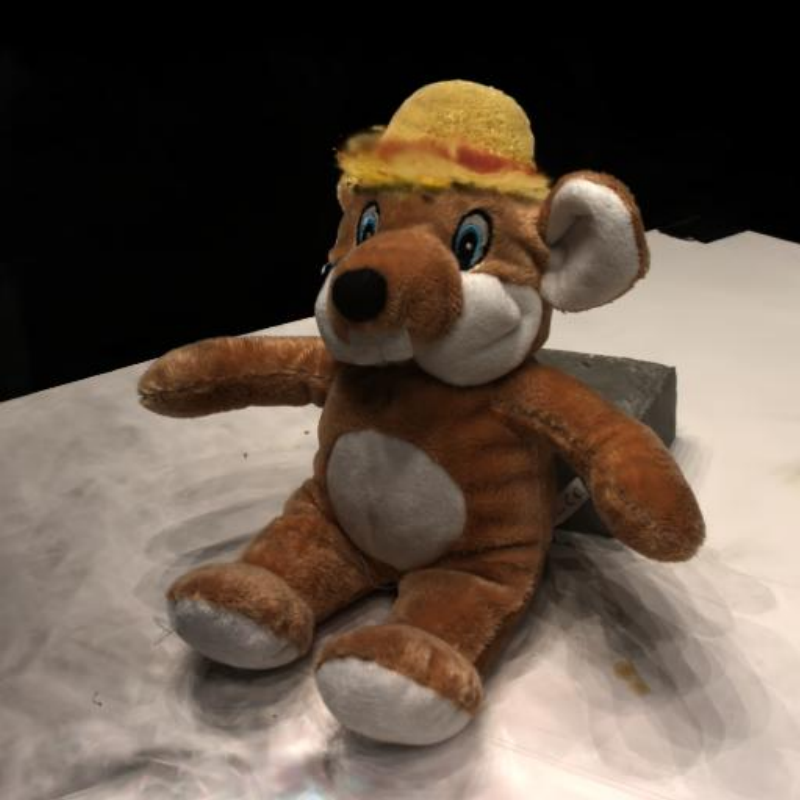}
        \end{minipage}
    \end{subfigure}
    \hspace{-0.03\textwidth}
    \begin{subfigure}{0.4\textwidth}
        \centering
        \begin{minipage}{\textwidth}
            \includegraphics[width=0.3\textwidth]{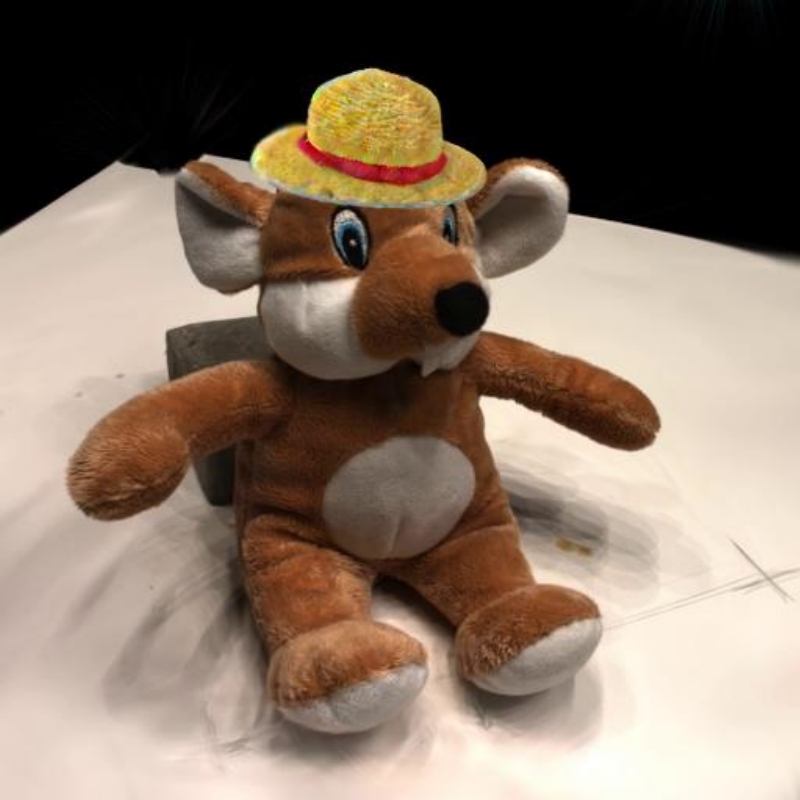}%
            \includegraphics[width=0.3\textwidth]{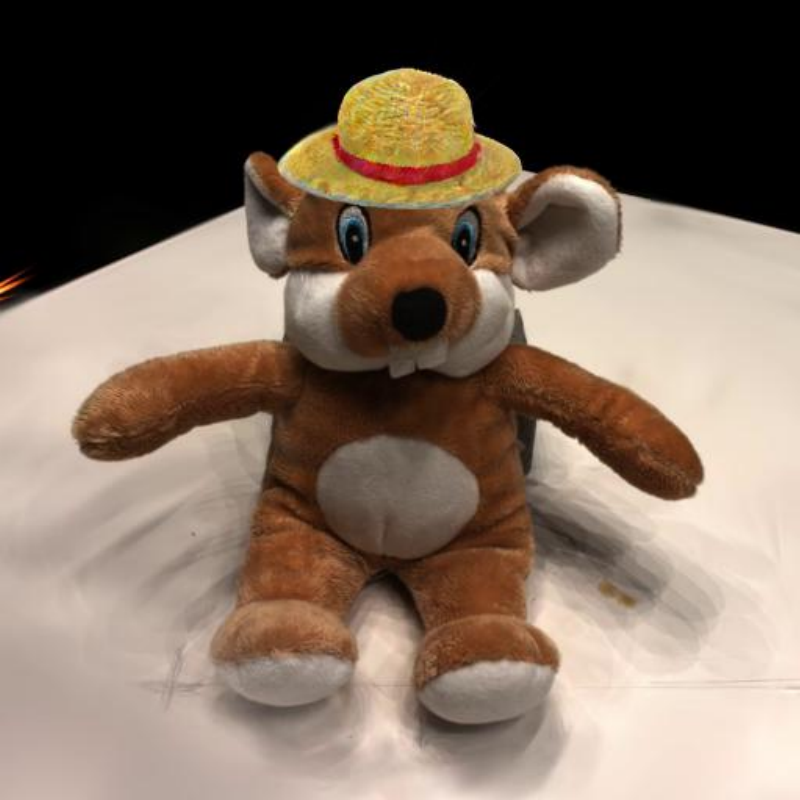}%
            \includegraphics[width=0.3\textwidth]{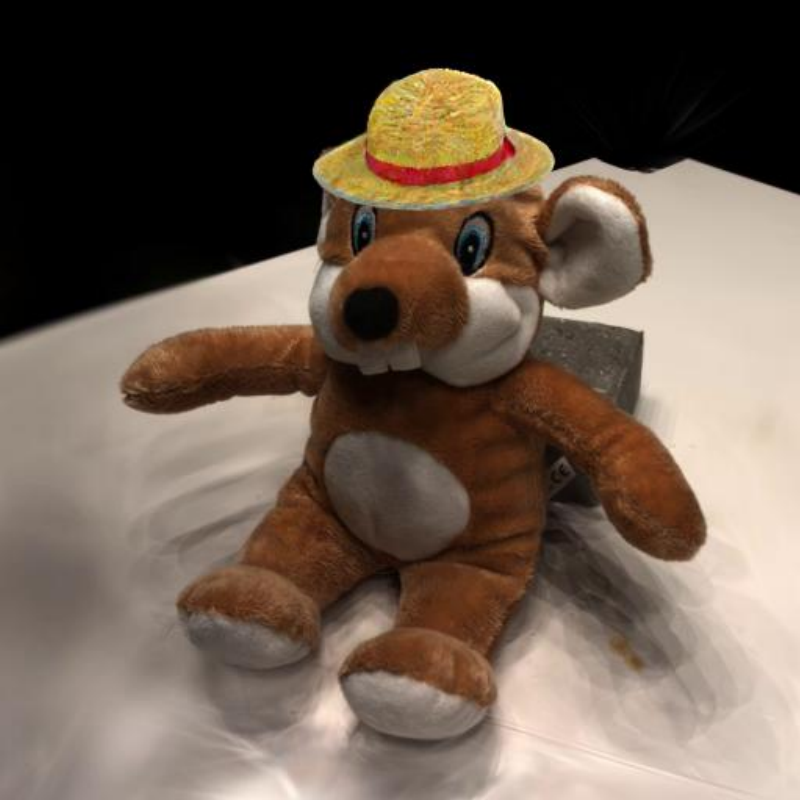}
        \end{minipage}
    \end{subfigure}
    \hspace*{0.145\textwidth} 
    \begin{minipage}{0.4\textwidth}
      \centering \small \textit{``A doll wearing a $V_1$ hat''}
    \end{minipage}
    \hspace{-0.03\textwidth}
    \begin{minipage}{0.4\textwidth}
      \centering \small \textit{``Add a hat to the doll''}
    \end{minipage}

    \hspace{0.02\textwidth}
    \begin{minipage}{0.145\textwidth}
    \begin{overpic}[width=0.9\linewidth]{figures/scene/face/0.75_90_0.pdf}
        \setlength{\fboxsep}{0pt}
        \put(65,-6){\includegraphics[width=0.35\linewidth]{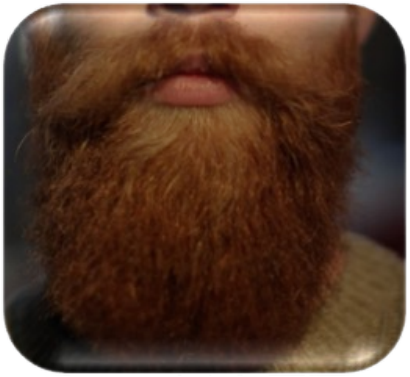}}
    \end{overpic}
  \end{minipage}
    \centering
    \begin{subfigure}{0.4\textwidth}
        \centering
        \begin{minipage}{\textwidth}
            \includegraphics[width=0.3\textwidth]{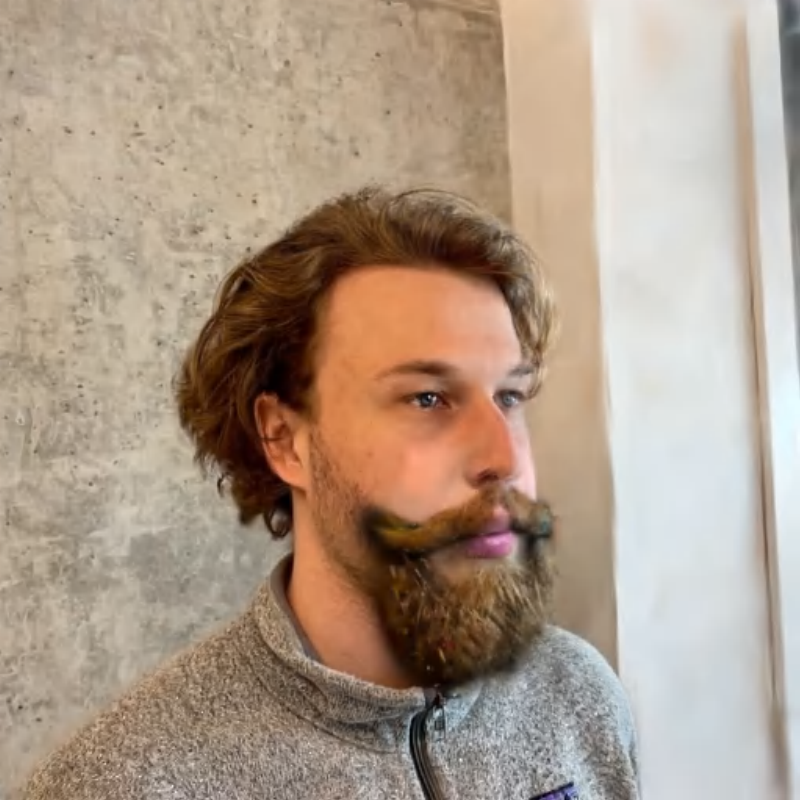}%
            \includegraphics[width=0.3\textwidth]{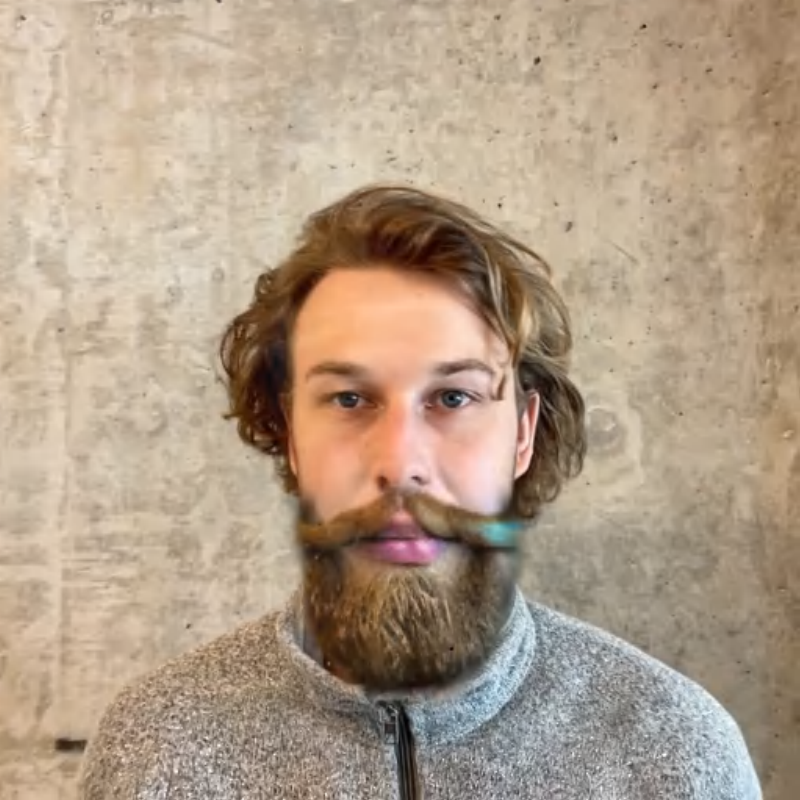}%
            \includegraphics[width=0.3\textwidth]{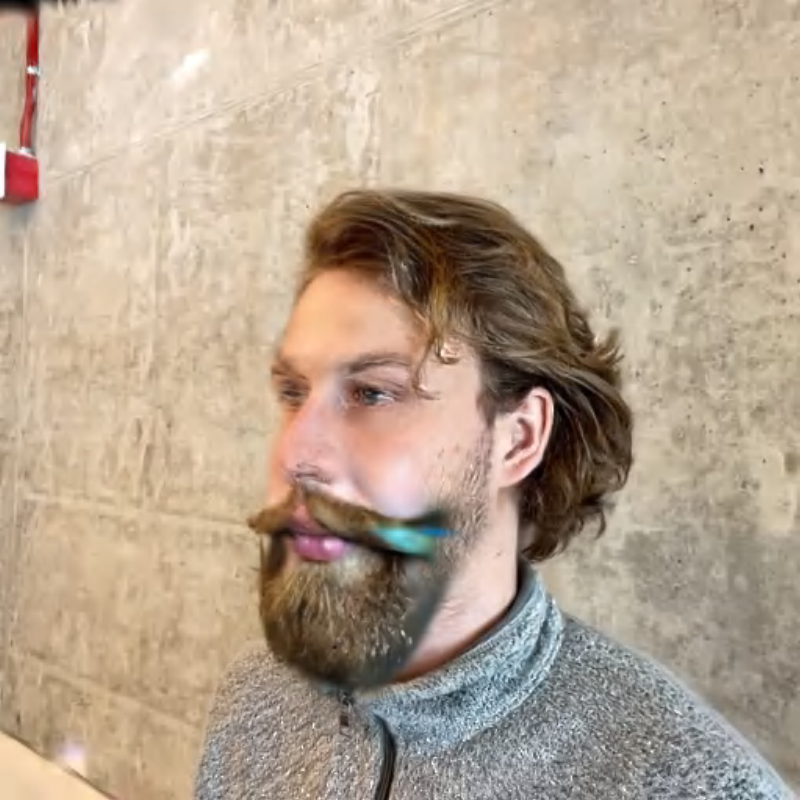}
        \end{minipage}
    \end{subfigure}
    \hspace{-0.03\textwidth}
    \begin{subfigure}{0.4\textwidth}
        \centering
        \begin{minipage}{\textwidth}
            \includegraphics[width=0.3\textwidth]{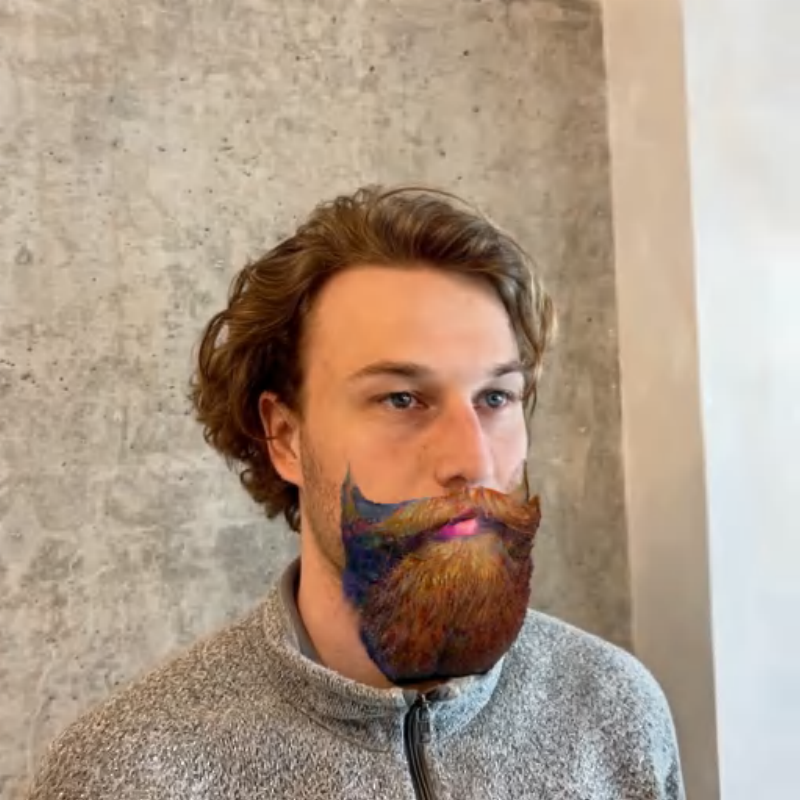}%
            \includegraphics[width=0.3\textwidth]{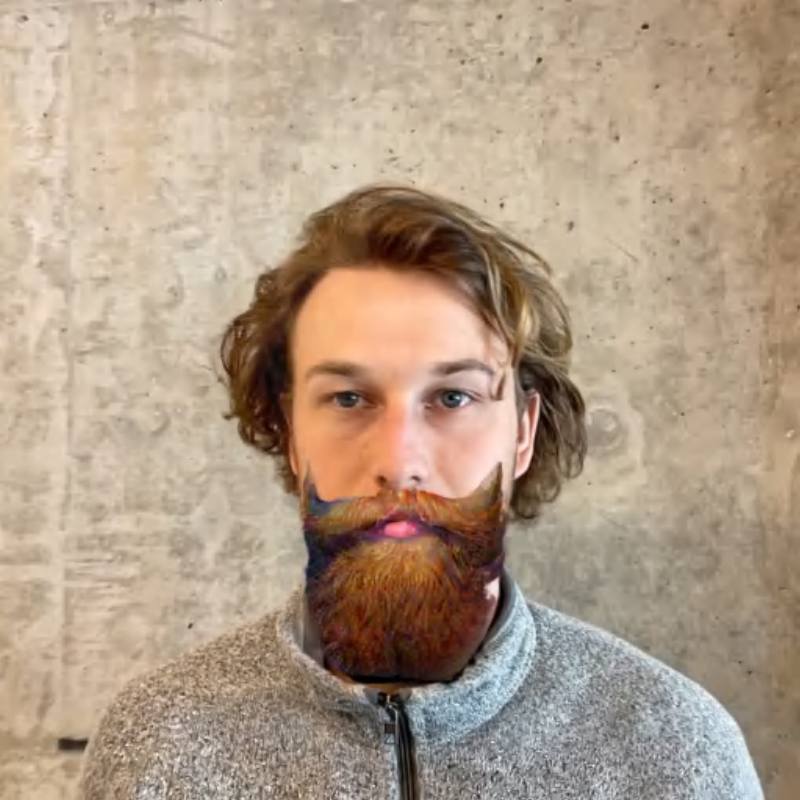}%
            \includegraphics[width=0.3\textwidth]{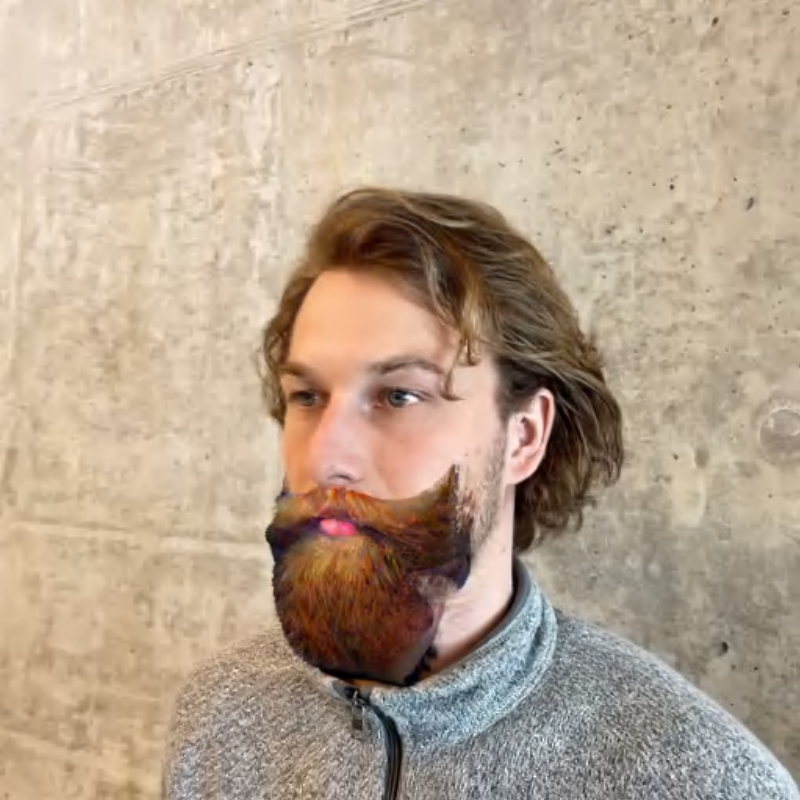}
        \end{minipage}
    \end{subfigure}
    \hspace*{0.145\textwidth} 
    \begin{minipage}{0.4\textwidth}
      \centering \small \textit{``A man with a $V_1$ beard''}
    \end{minipage}
    \hspace{-0.03\textwidth}
    \begin{minipage}{0.4\textwidth}
      \centering \small \textit{``Add a beard to the man''}
    \end{minipage}

    \hspace{0.02\textwidth}
    \begin{minipage}{0.145\textwidth}
    \begin{overpic}[width=0.9\linewidth]{figures/scene/horse/1.3_75_0.pdf}
      \put(65,-5){\includegraphics[width=0.35\linewidth]{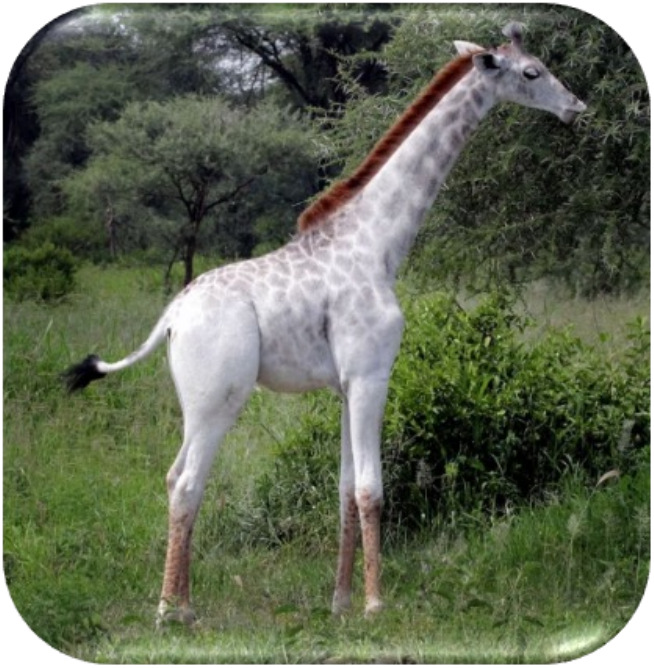}}
    \end{overpic}
  \end{minipage}
    \centering
    \begin{subfigure}{0.4\textwidth}
        \centering
        \begin{minipage}{\textwidth}
            \includegraphics[width=0.3\textwidth]{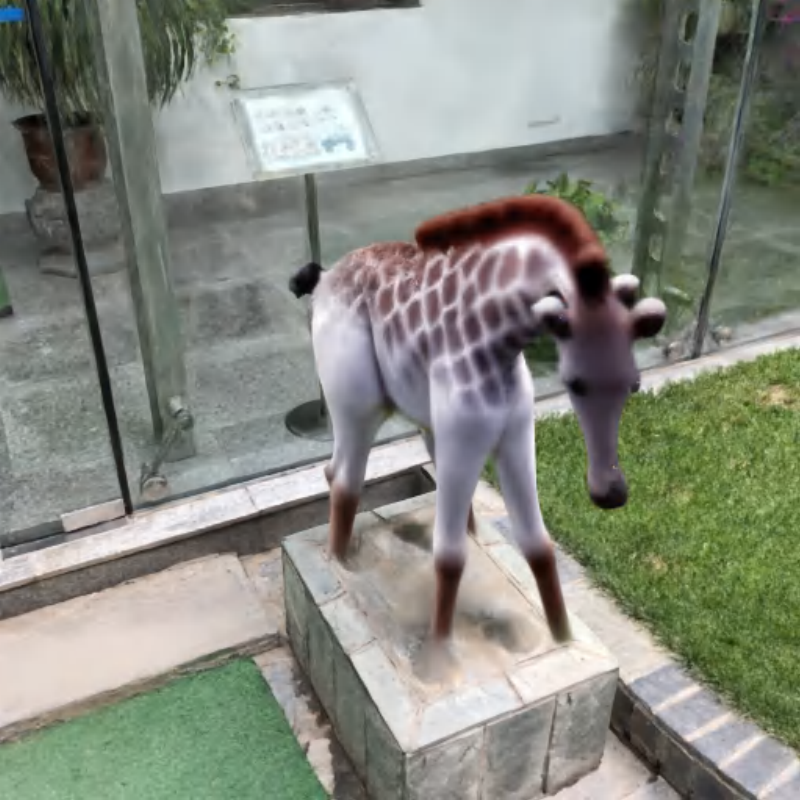}%
            \includegraphics[width=0.3\textwidth]{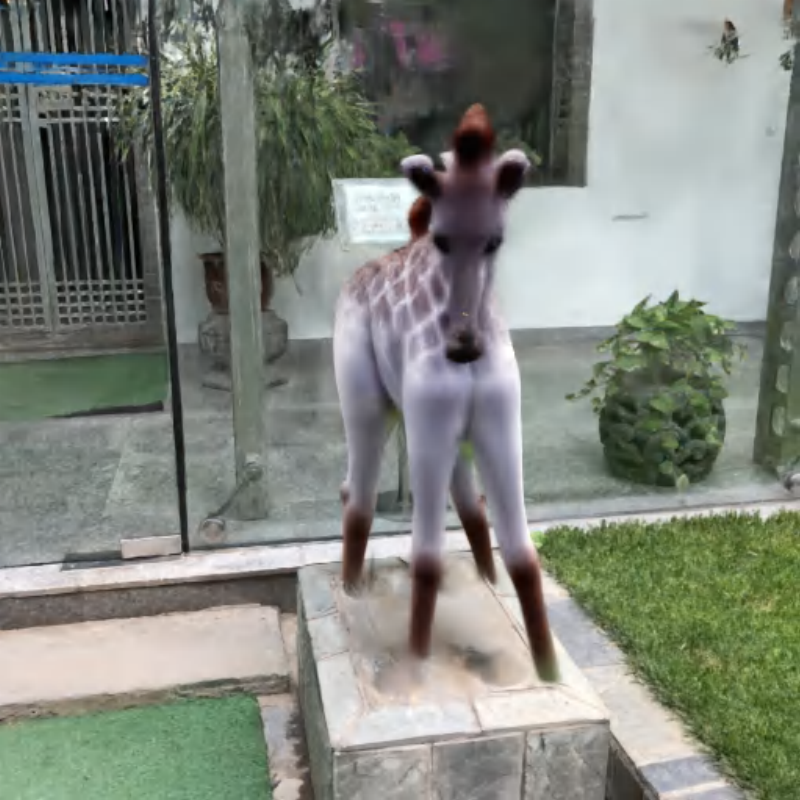}%
            \includegraphics[width=0.3\textwidth]{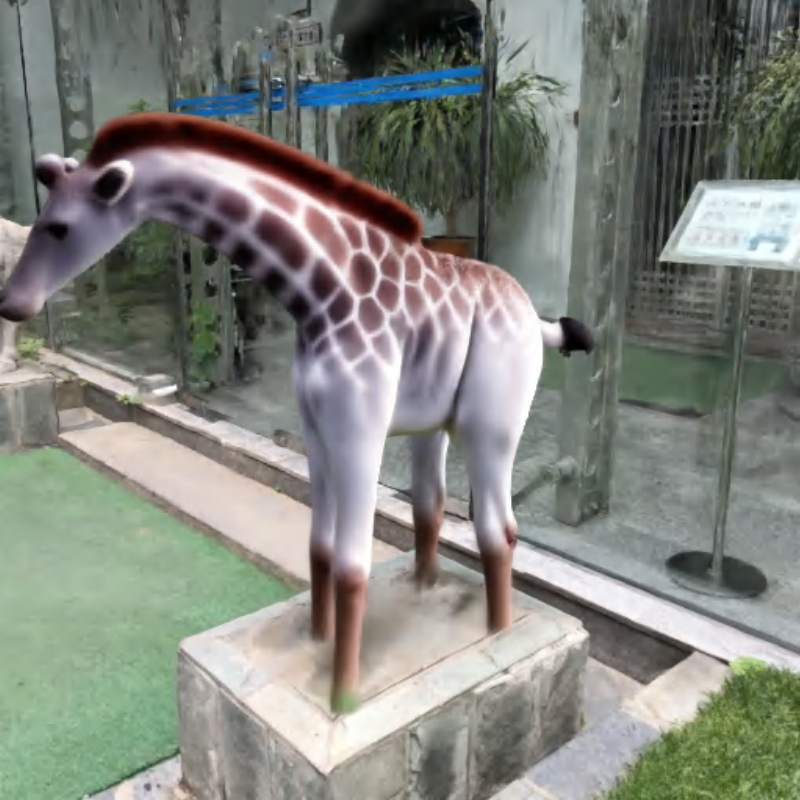}
        \end{minipage}
    \end{subfigure}
    \hspace{-0.03\textwidth}
    \begin{subfigure}{0.4\textwidth}
        \centering
        \begin{minipage}{\textwidth}
            \includegraphics[width=0.3\textwidth]{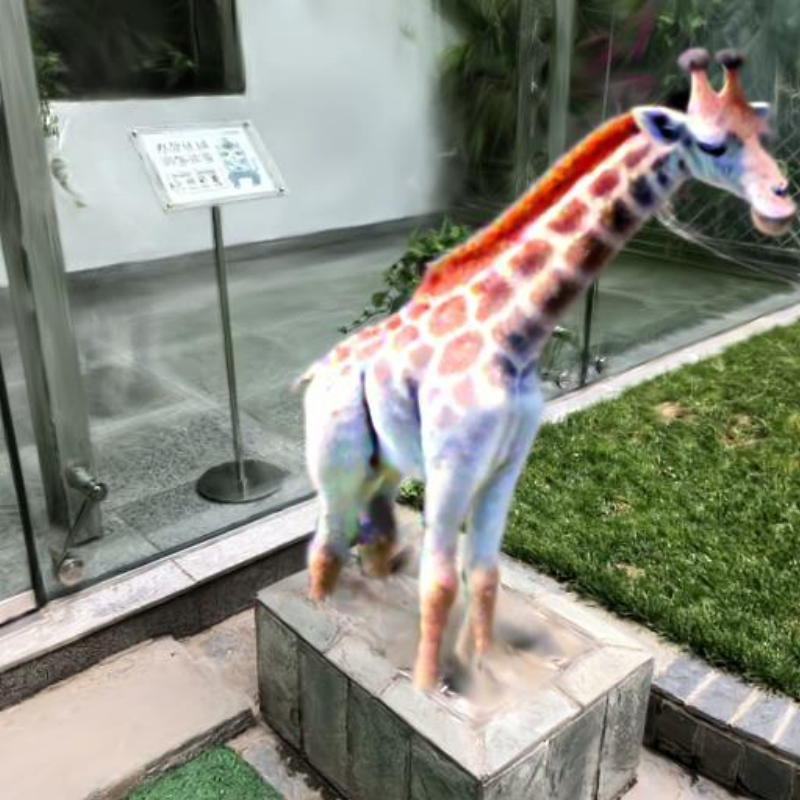}%
            \includegraphics[width=0.3\textwidth]{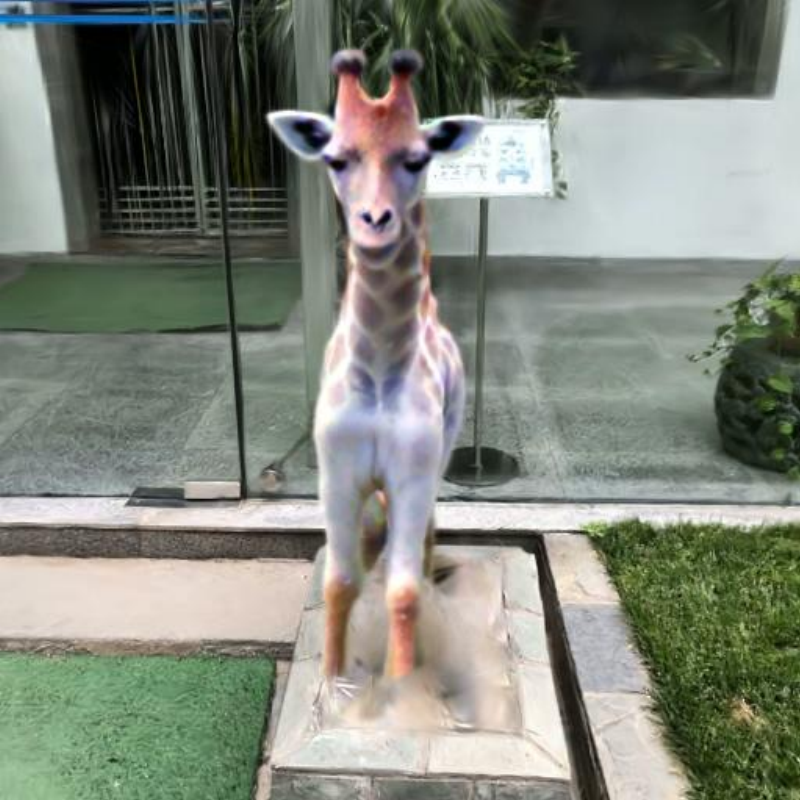}%
            \includegraphics[width=0.3\textwidth]{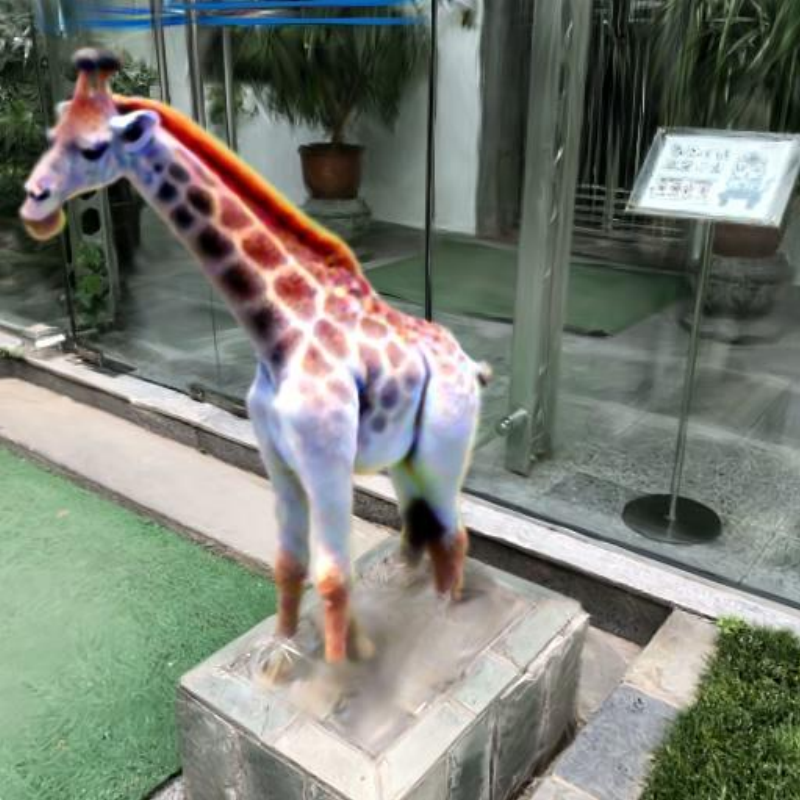}
        \end{minipage}

    \end{subfigure}
    \hspace*{0.145\textwidth} 
    \begin{minipage}{0.4\textwidth}
      \centering \small \textit{``A $V_1$ giraffe in a garden''} 
    \end{minipage}
    \hspace{-0.03\textwidth}
    \begin{minipage}{0.4\textwidth}
      \centering \small \textit{``Add a giraffe to replace the horse''}
    \end{minipage}

    \hspace{0.02\textwidth}
    \begin{minipage}{0.145\textwidth}
        \begin{overpic}[width=0.9\linewidth]{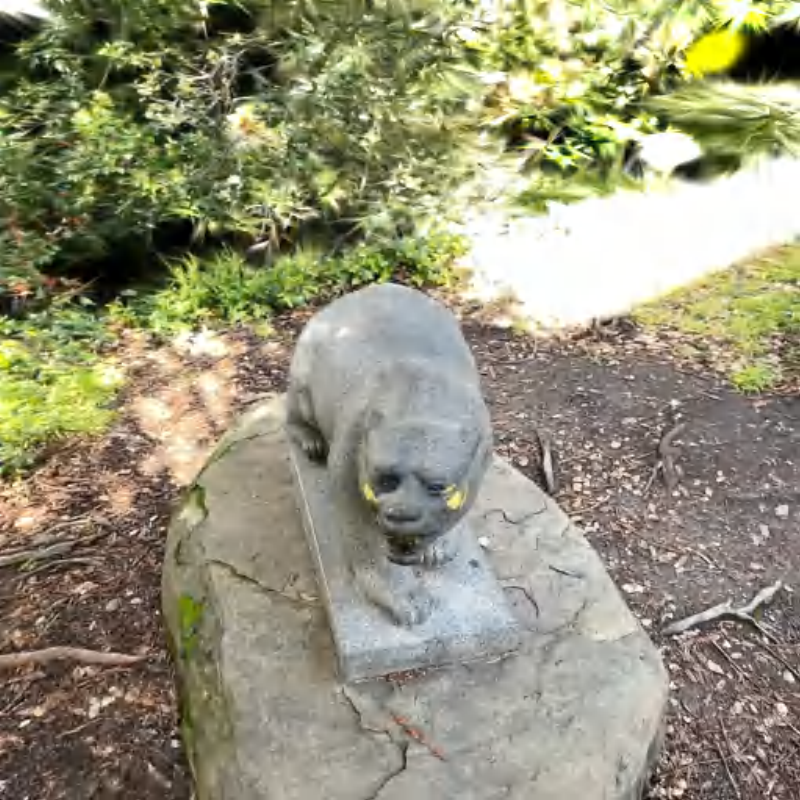}
          \setlength{\fboxsep}{0pt}
        \put(65,-6){\includegraphics[width=0.35\linewidth]{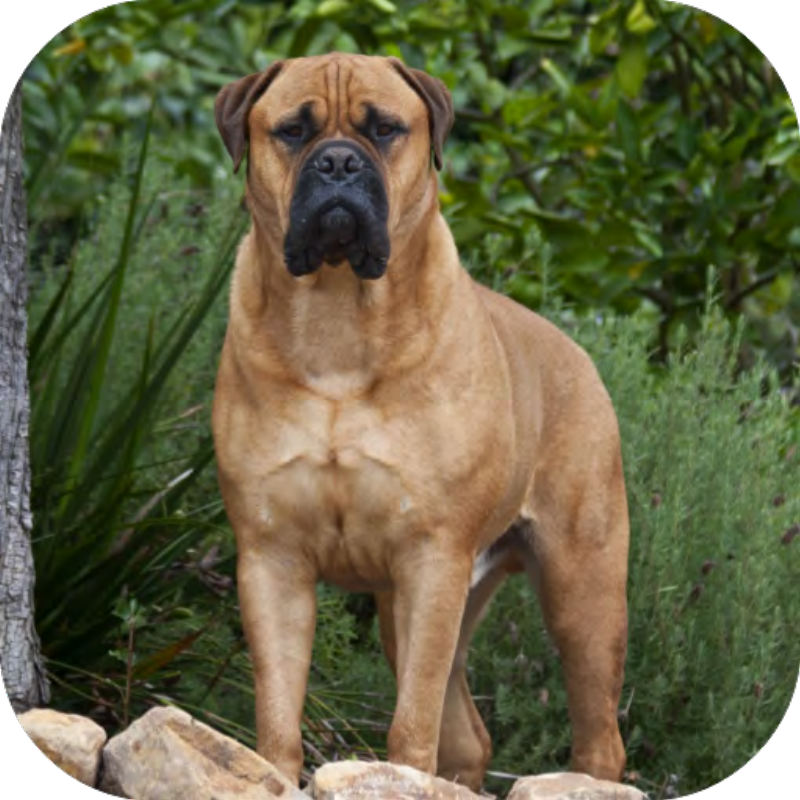}}
        \end{overpic}
      \end{minipage}
    \centering
    \begin{subfigure}{0.4\textwidth}
        \centering
        \begin{minipage}{\textwidth}
            \includegraphics[width=0.3\textwidth]{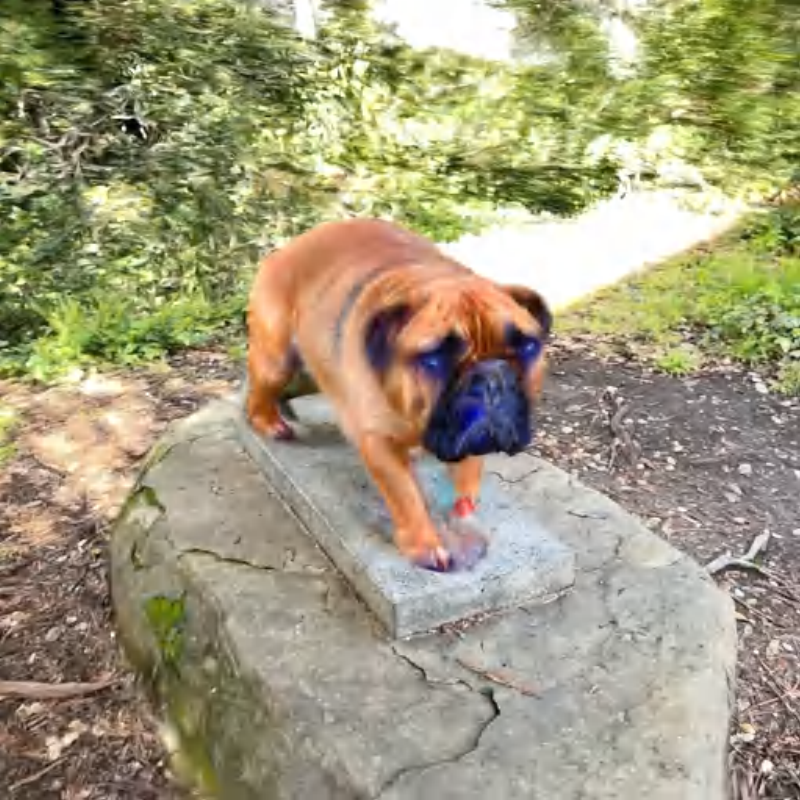}%
            \includegraphics[width=0.3\textwidth]{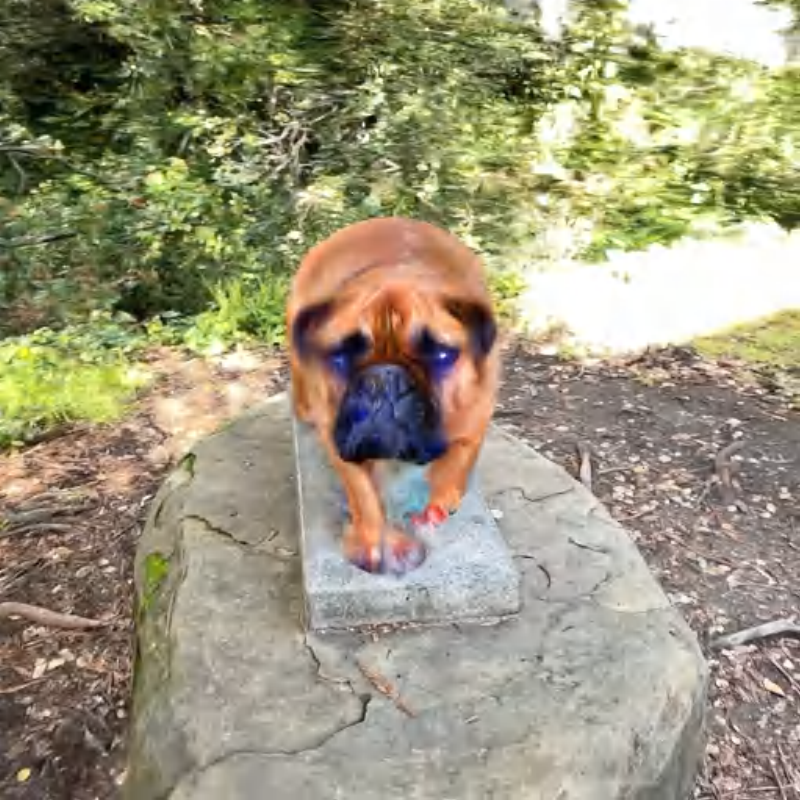}%
            \includegraphics[width=0.3\textwidth]{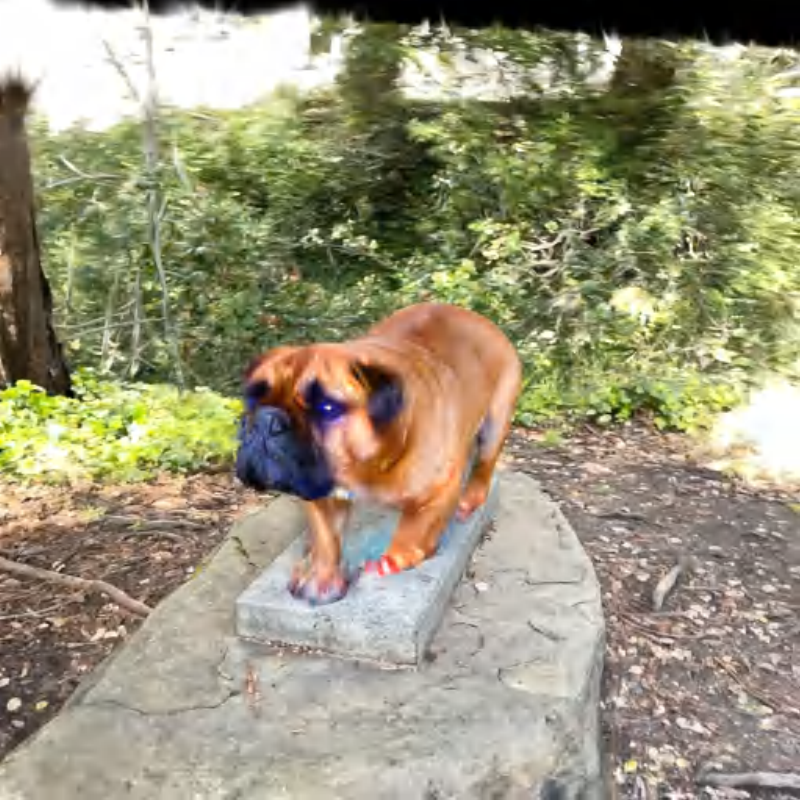}
        \end{minipage}
    \end{subfigure}
    \hspace{-0.03\textwidth}
    \begin{subfigure}{0.4\textwidth}
        \centering
        \begin{minipage}{\textwidth}
            \includegraphics[width=0.3\textwidth]{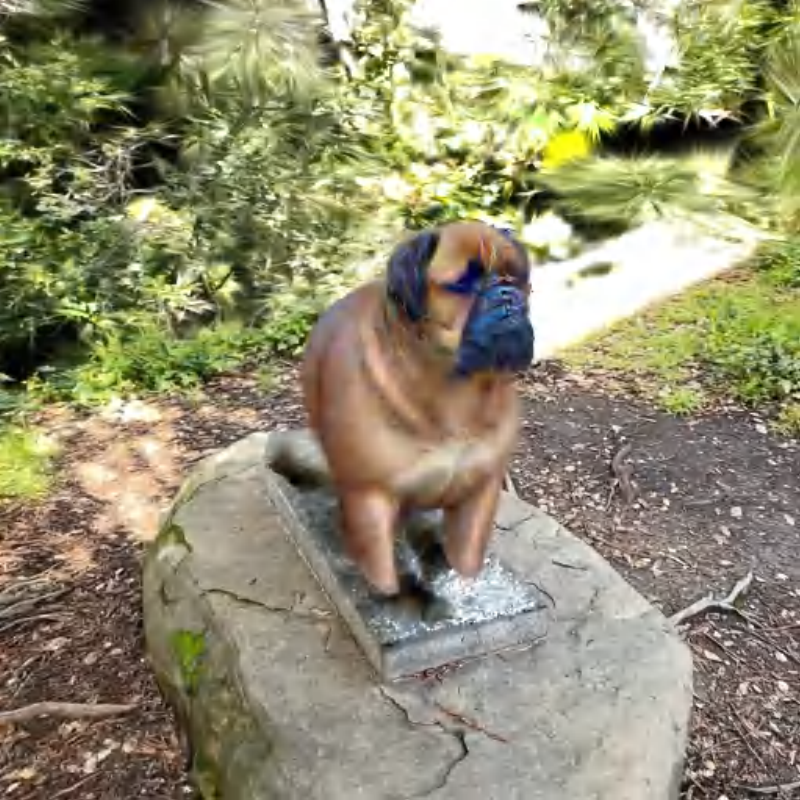}%
            \includegraphics[width=0.3\textwidth]{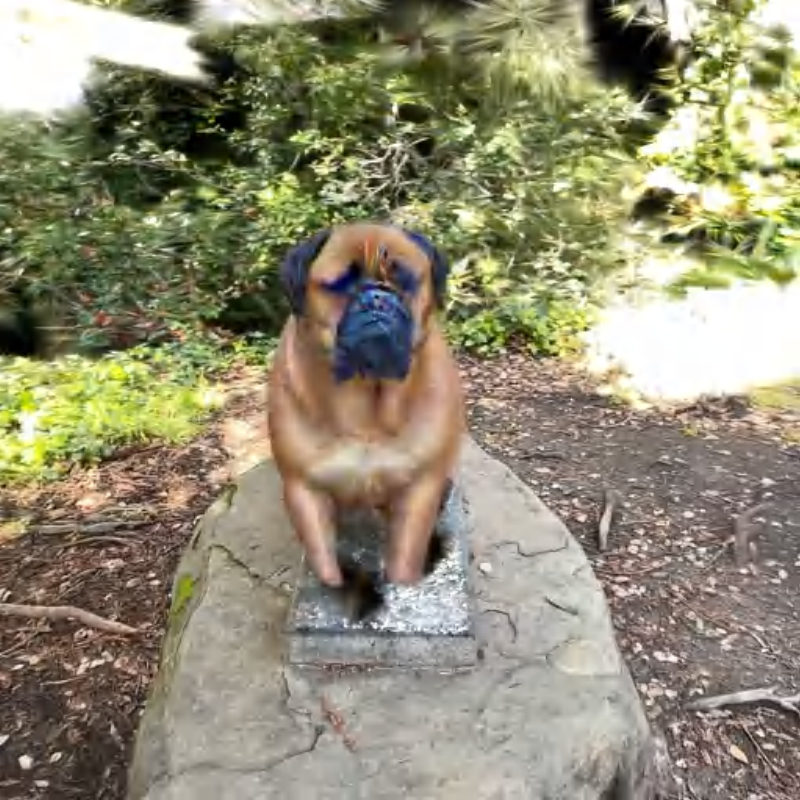}%
            \includegraphics[width=0.3\textwidth]{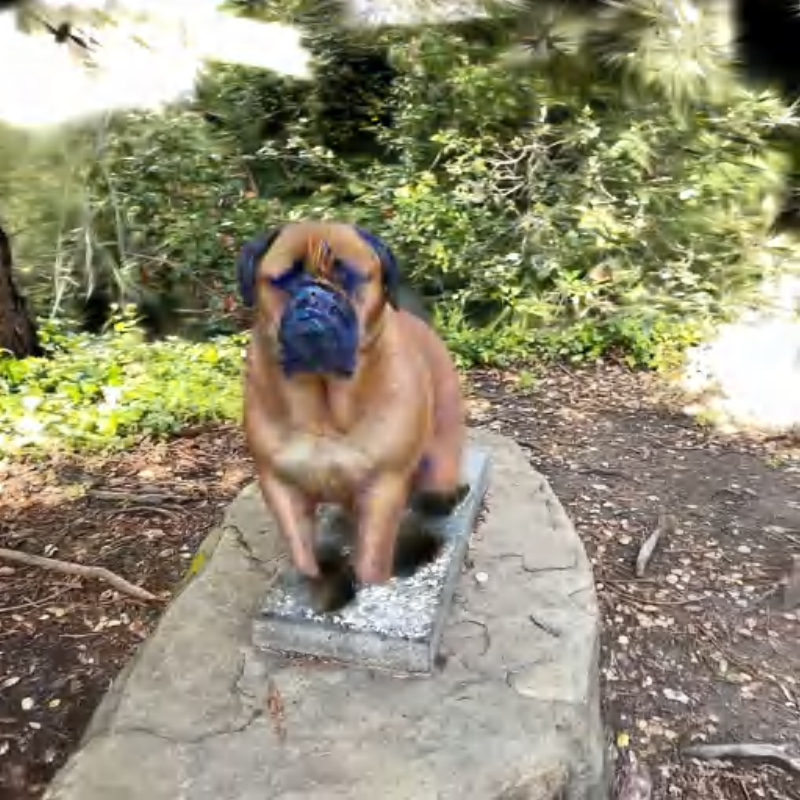}
        \end{minipage}
    \end{subfigure}
    \hspace*{0.145\textwidth} 
    \begin{minipage}{0.4\textwidth}
      \centering \small \textit{``A $V_1$ dog in woods''} 
    \end{minipage}
    \hspace{-0.03\textwidth}
    \begin{minipage}{0.4\textwidth}
      \centering \small \textit{``Add a dog to replace the stone bear''}
    \end{minipage}
    \captionsetup{skip=2pt}
    \caption{Visual comparisons with TIP-Editor using text-image prompts.
    Our method achieves competitive results with TIP-Editor, \emph{without relying on 3D bounding boxes}.
    TIP-Editor struggles to maintain the 3D consistency of the inserted object (e.g., the misaligned hat across views in row 2, column 2, and the right front paw intersecting with the left in row 5, column 2), as its 2D editing process lacks cross-view constraints.
    Our method produces clearly more 3D-consistent results and more closely resembles the reference image.
    }
    \vspace{-3mm}
    \label{fig:comparision in specific appearance}
\end{figure*}

\noindent\textbf{Evaluation Criteria.} We use CLIP Text-Image directional similarity following \cite{chen2024gaussianeditor,zhuang2024tip,wu2024gaussctrl,zhuang2023dreameditor} to assess the alignment between the text and the editing results. 
For appearance-specified cases, we further employ DINO similarity~\cite{oquab2024dinov} following \cite{zhuang2024tip} to assess appearance preservation.
We also conducted a user study with 50 participants, who rated the 3D editing results (presented with prompts in shuffled order) on four criteria: Semantic Alignment, Object Integrity, Geometric Consistency, and Detail Preservation, using a 1–10 scale.

\vspace{-3mm}

\subsection{Comparisons with State-of-the-Art Methods}
\subsubsection{Qualitative comparisons}
In this part, we conduct a qualitative comparison with different baselines under two types of input settings (text prompt and text-image prompt) to evaluate their performance under identical conditions.
Video demonstrations are included in the supplementary.

\mypara{Text Prompt Comparisons} ~\Cref{fig:comparision with advanced} shows visual comparisons between our method with three baselines. 
Both IN2N(GS) and GaussCtrl, which rely solely on semantic guidance, struggle to successfully complete insertions in some scene-object combinations, such as ``Add a pair of glasses to the doll''. 
Although GaussCtrl improves consistency, the replacements remain too similar to the original (e.g., a horse), failing to convincingly resemble the target (e.g., a giraffe).
GaussianEditor relies on user-provided 2D masks for object insertion but struggles in object- or human-centric scenes due to inaccurate post-inpainting segmentation, leading to artifacts like foreground overlaps. 
While it performs better in outdoor scenes (e.g., the giraffe example), its depth estimation is often imprecise and requires manual adjustment.
By contrast, our method achieves high-quality object insertion and replacement results in both scene preservation and object completeness without requiring any manual annotations.

\mypara{Text-Image Prompt Comparisons} Besides the text-prompt methods, we further evaluate object insertion and replacement with a given image prompt of the specified object in \Cref{fig:comparision in specific appearance}, comparing against TIP-Editor~\cite{zhuang2024tip}.
Although TIP-Editor supports flexible insertion via 3D bounding boxes, it suffers from inconsistent multi-view appearances due to its reliance on 2D editing techniques.
Most critically, achieving such results remains dependent on finely user-provided 3D bounding boxes, which significantly hinders scalability and practicality.
In contrast, our method delivers the most complete geometry and better appearance fidelity to the image prompt without relying on any annotation.
\vspace{-3mm}

\subsubsection{Quantitative Comparisons.} 
\Cref{table:tab2} presents the quantitative comparison of our method against other baseline methods. 
Our method achieves CLIP text–image semantic alignment scores comparable to the state-of-the-art methods without requiring any manual annotations for the insertion region.
Moreover, compared to the approach that specifies object appearances (TIP-Editor), our method exhibits higher DINO similarity to the image prompt.
The $\text{User}_{\textit{vote}}$ ratings clearly demonstrate that users prefer our method over the baselines.

\begin{table}[h]
    \small
    \centering
    \caption{Quantitative comparisons to SOTA.
    CLIP$_{dir}$ denotes the CLIP Text-Image directional similarity.
    DINO$_{sim}$ is the DINO similarity.
    }
    \vspace{-3mm}
    \begin{tabularx}{0.8\linewidth}{l@{\hskip 5mm}X@{\hskip 5mm}X@{\hskip 6mm}X}
        \toprule
        Method & CLIP$_{dir}$$\uparrow$ & DINO$_{sim}$$\uparrow$ &User$_{vote}$$\uparrow$  \\
        \midrule
        InstructN2N(GS) \cite{vachhainstruct} & 26.76\% & -  & 18.3\\
        GaussianEditor \cite{chen2024gaussianeditor} & 27.36\% & -  & 24.7\\
        GaussCtrl \cite{wu2024gaussctrl} & 25.39\% & -  & 26.3\\
        TIP-Editor  \cite{zhuang2024tip} & \textbf{30.01}\% & 83.30\% & 32.3  \\
        \name & 29.48\% & \textbf{83.45\%} & \textbf{36.9} \\
        \bottomrule
    \end{tabularx}
    \label{table:tab2}
    \vspace{-3mm}
\end{table}

\subsection{Ablation Studies}
\begin{figure}
\includegraphics[width=1\linewidth]{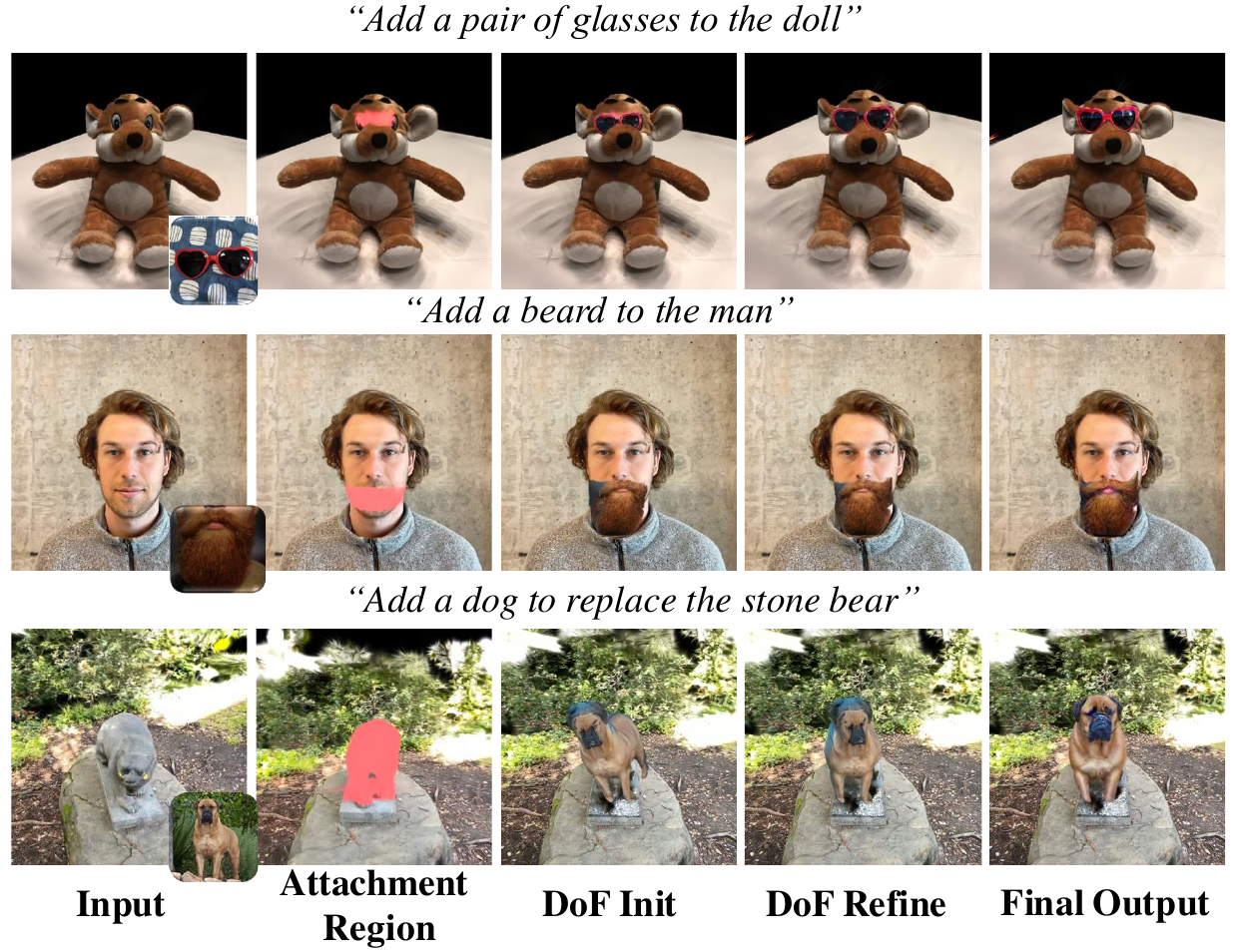}
\caption{Visualization of different stages in \name}
\label{fig:ablation_stage_0}
\end{figure}

\noindent\textbf{Ablation Visualization Across Stages.}
To better demonstrate how each stage in \name contributes to the final outcome, we visualize the intermediate results at each step, as shown in~\Cref{fig:ablation_stage_0}.
Col 2 highlights the attachment region Gaussians within the scene, marked in red. 
Col 3 shows the initial degrees of freedom (DoF), which are typically sub-optimal. 
After SSDS refinement, the object achieves a more accurate DoF (Col 4). 
Finally, Col 5 demonstrates the enhanced object appearance.

\begin{figure}[h]  
    \centering
    \small
    \makebox[\linewidth][c]{%
        \begin{subfigure}{0.25\linewidth}
            \centering
            \includegraphics[width=\linewidth]{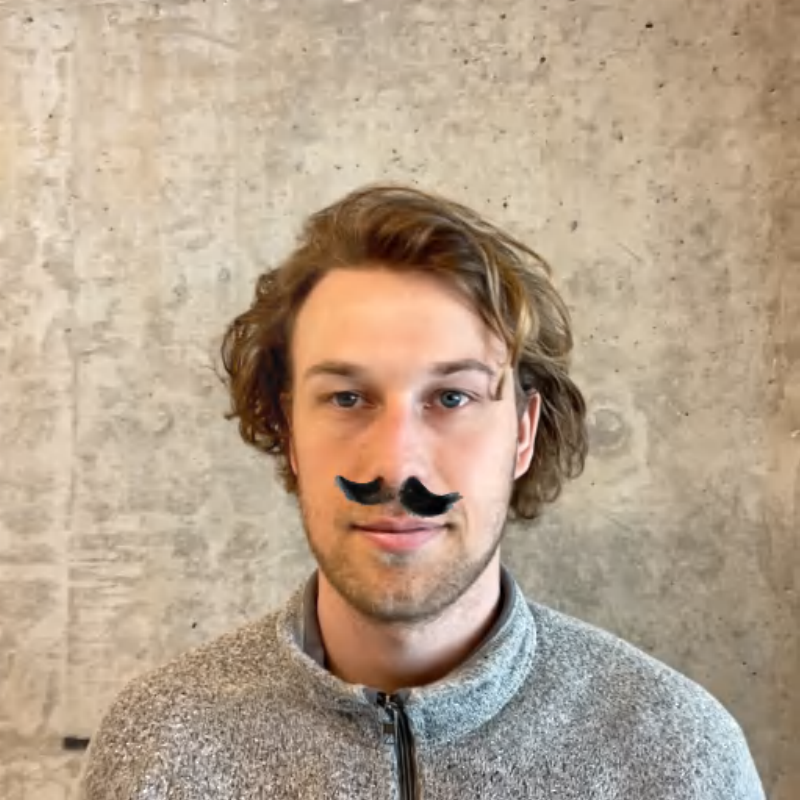}
            \caption{w/o \( \textit{ssds} \)}
        \end{subfigure}
        \hspace*{\fill}
        \begin{subfigure}{0.25\linewidth}
            \centering
            \includegraphics[width=\linewidth]{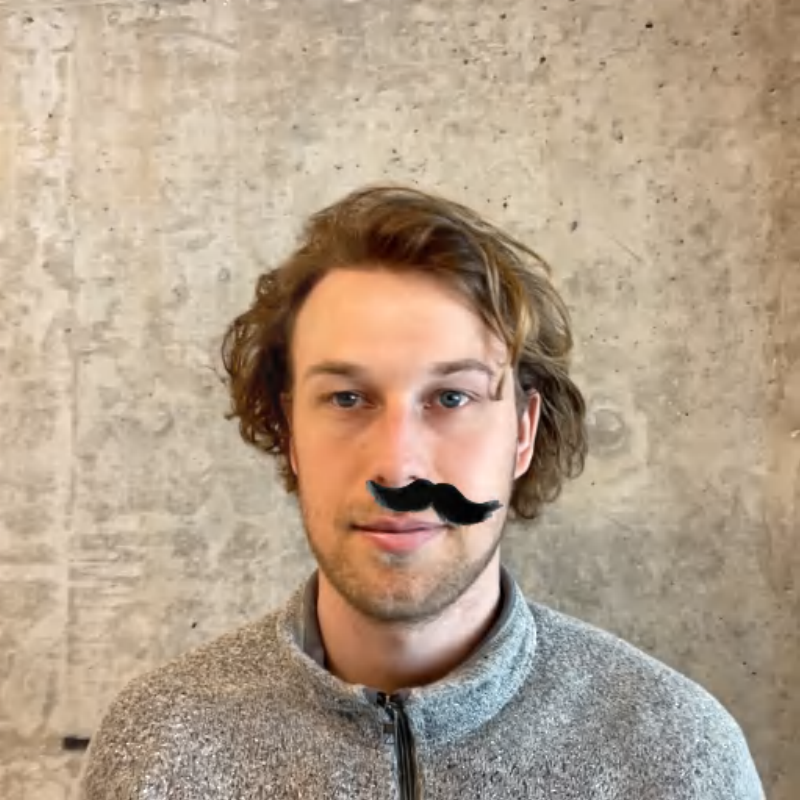}
            \caption{w/o\( \mathcal{L}_{\textit{ssds-local}} \)}
        \end{subfigure}
        \hspace*{\fill}
        \begin{subfigure}{0.25\linewidth}
            \centering
            \includegraphics[width=\linewidth]{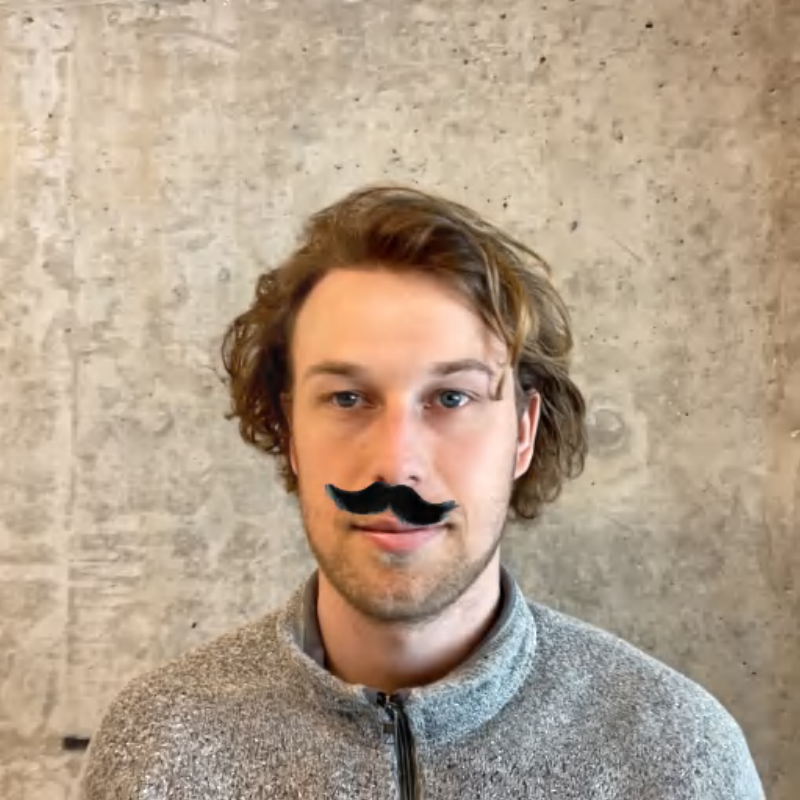}
         \caption{w/o\( \mathcal{L}_{\tiny\textit{ssds-global}} \)}
        \end{subfigure}
        \hspace*{\fill}
        \begin{subfigure}{0.25\linewidth}
            \centering
            \includegraphics[width=\linewidth]{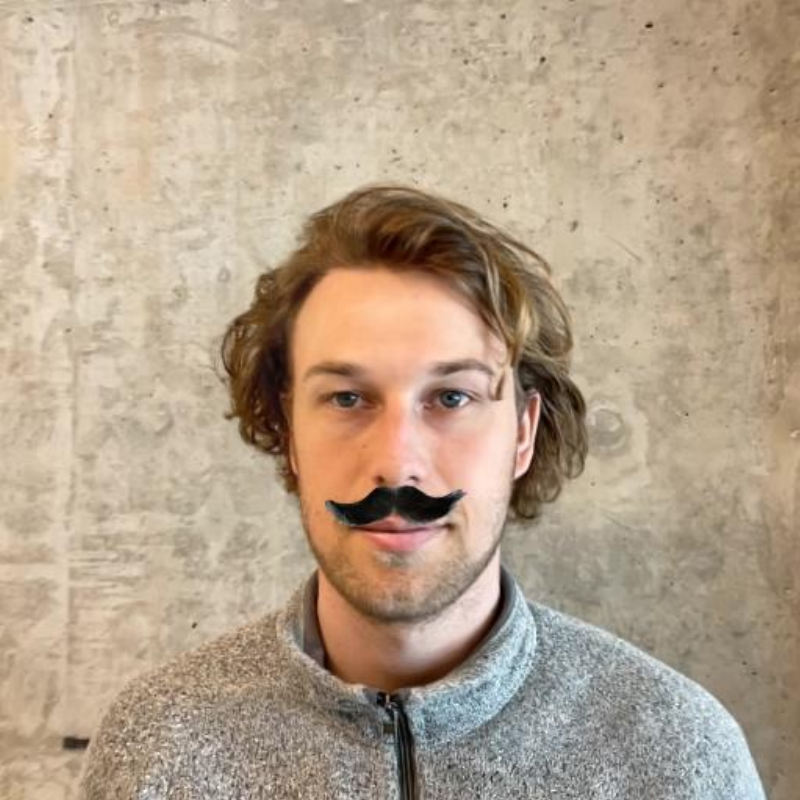}
            \caption{\name}
        \end{subfigure}
    }
    \captionsetup{skip=2pt}
    \caption{Ablation of Global-Local Collaborative Spatial Awareness.}
    \label{fig:Hierarchical Spatial Aware}
    \vspace{-0.2cm}
\end{figure}

\noindent\textbf{Effectiveness of Global-Local Collaborative Spatial Awareness.} To verify the effectiveness of global-local collaborative spatial-aware strategy, we conduct ablation studies comparing the following variants: no ssds, only \( \mathcal{L}_\textit{ssds-global} \), only \( \mathcal{L}_\textit{ssds-local} \), and the combination of both. 
As shown in \Cref{fig:Hierarchical Spatial Aware}, global prompt like “A man with moustache” often lead to ambiguous placements, while local prompt such as “A moustache is under the nose and above the upper lip” provide precise constraints but may ignore global plausibility. 
Our method balances both by reweighting key spatial terms and progressively shifting from local to global focus, resulting in more accurate and semantically coherent placements.

\begin{figure}[htbp]  
    \begin{minipage}{0.45\linewidth}
      \centering \small \textit{``Add a pair of sunglasses \\ on the forehead''}
    \end{minipage}%
    \begin{minipage}{0.45\linewidth}
      \centering \small \textit{``Add an apple on \\ the center of table''}
    \end{minipage}
    \centering
    \begin{subfigure}{0.49\linewidth}
        \centering
        \begin{minipage}{\textwidth}
            \centering
            \begin{subfigure}[b]{0.49\textwidth}
                \includegraphics[width=\textwidth]{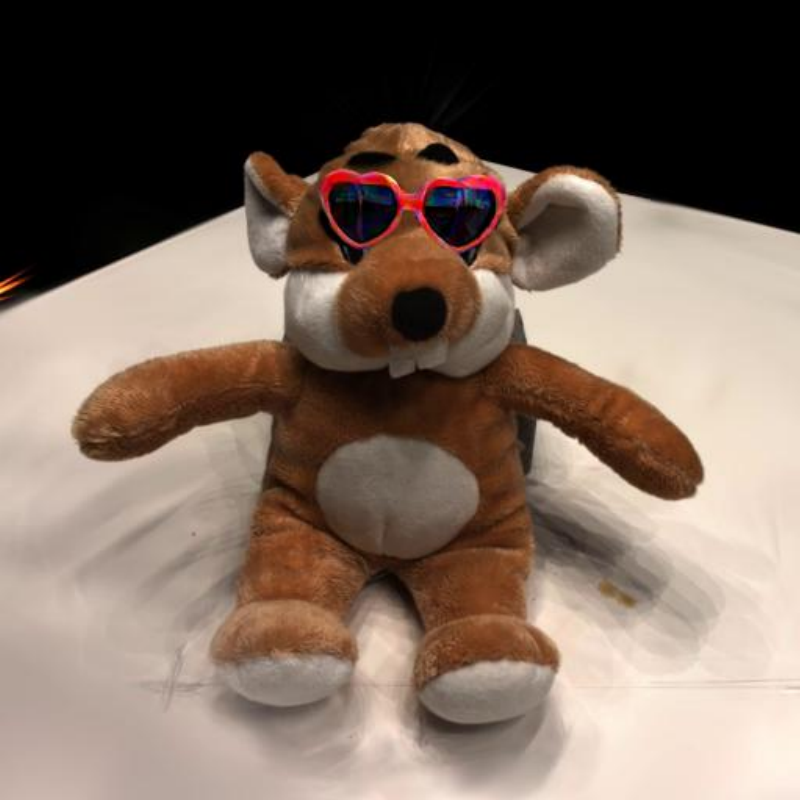}
                \caption{w/o \(\mathcal{L}_{loc}\)}
            \end{subfigure}%
            \hfill
            \begin{subfigure}[b]{0.49\textwidth}
                \includegraphics[width=\textwidth]{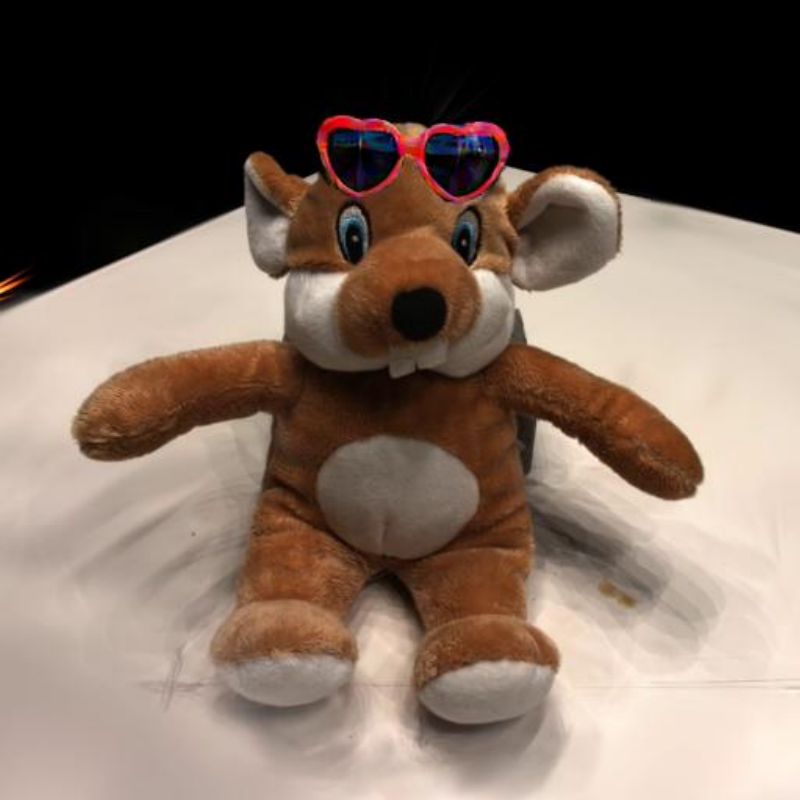}
                \caption{\name}
            \end{subfigure}
        \end{minipage}
    \end{subfigure}
    \hfill
    \begin{subfigure}{0.49\linewidth}
        \centering
        \begin{minipage}{\textwidth}
            \centering
            \begin{subfigure}[b]{0.48\textwidth}
                \includegraphics[width=\textwidth]{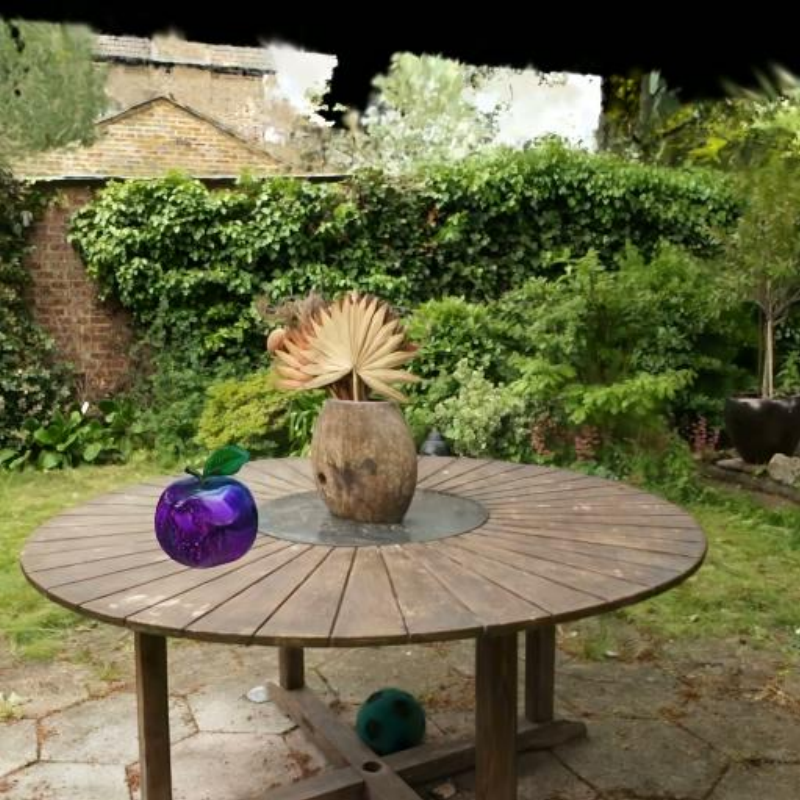}
                \caption{w/o \(\mathcal{L}_{loc}\)}
            \end{subfigure}%
            \hfill
            \begin{subfigure}[b]{0.48\textwidth}
                \includegraphics[width=\textwidth]{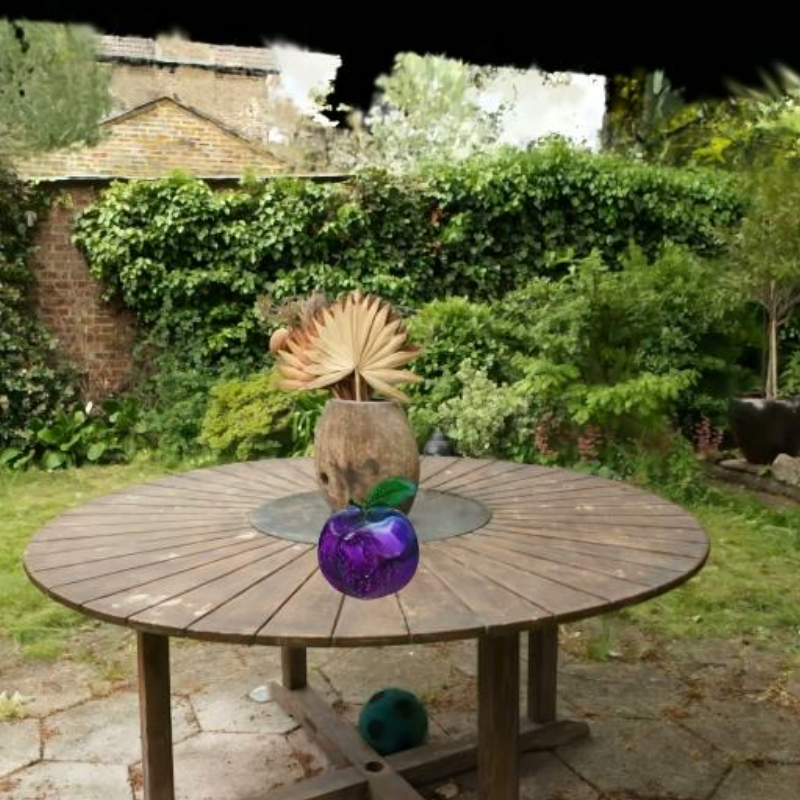}
                \caption{\name}
            \end{subfigure}
        \end{minipage}
    \end{subfigure}
    \captionsetup{skip=2pt}
    \caption{Ablation of Attention-based Localization}
    \label{fig:attention-based}
\end{figure}
\noindent\textbf{Effectiveness of Attention-based localization.} 
We evaluate the contribution of $\mathcal{L}_{loc}$ to enhancing spatial awareness, particularly for rare or ambiguous placements. 
As shown in~\Cref{fig:attention-based}, $\mathcal{L}_{loc}$ encourages object position within the bounding box region inferred by the large model, while allowing flexibility to adjust the DoF. 
Unlike fixed constraints, it softly guides attention toward intended regions, mitigating semantic control failures caused by training data biases.

\begin{figure}[htbp]  
    \centering
    \begin{subfigure}{0.24\linewidth}
        \centering
        \includegraphics[width=\linewidth]{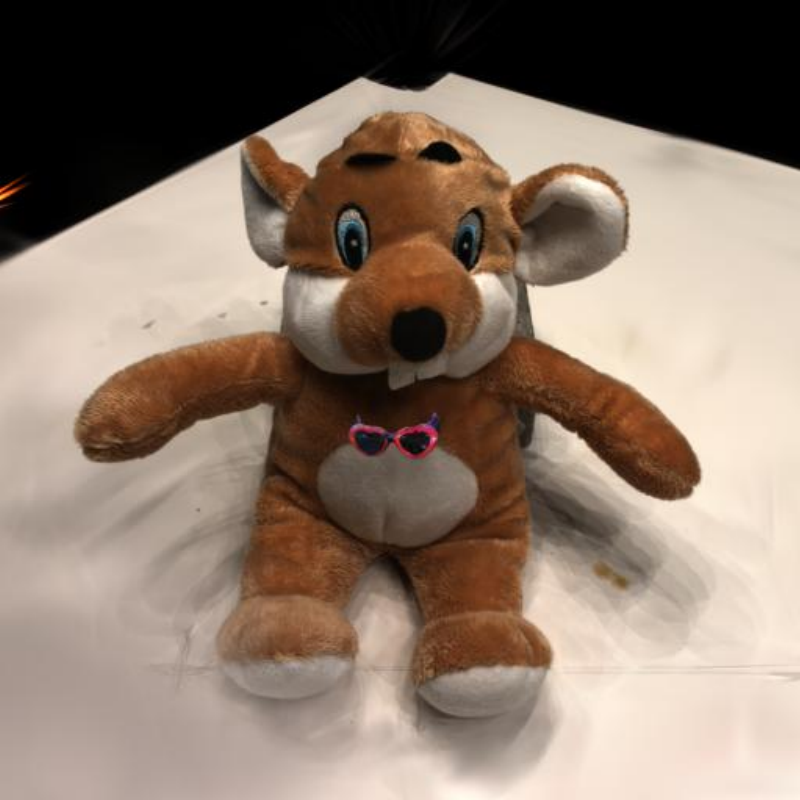}
        \par\textbf{GPT-4V\cite{achiam2023gpt}}
    \end{subfigure}
    \hfill
    \begin{subfigure}{0.24\linewidth}
        \centering
        \includegraphics[width=\linewidth]{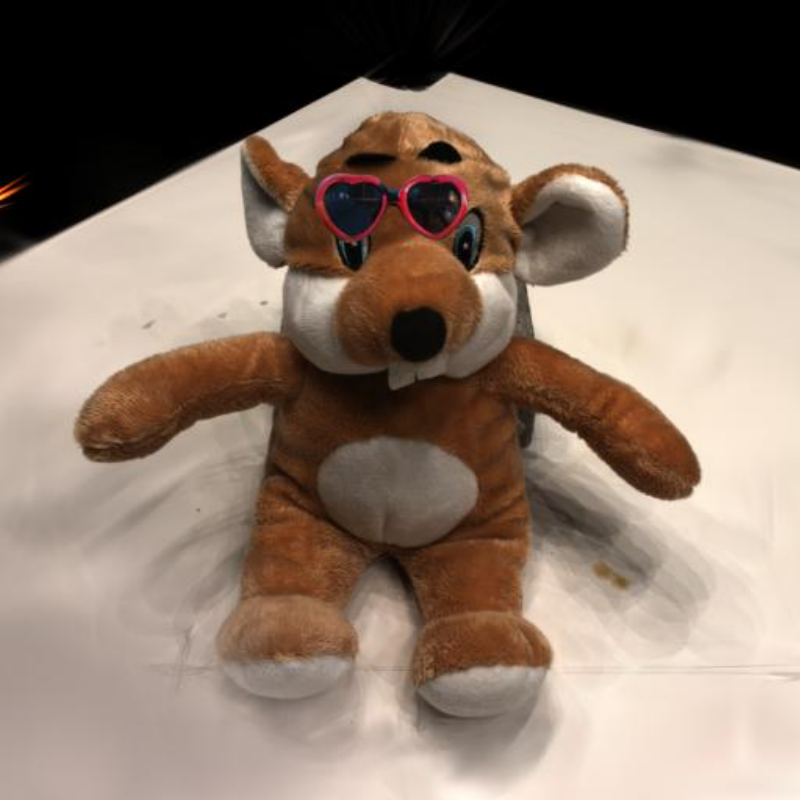}
        \par\textbf{Molmo-7B\cite{deitke2024molmo}}
    \end{subfigure}
    \hfill
    \begin{subfigure}{0.24\linewidth}
        \centering
        \includegraphics[width=\linewidth]{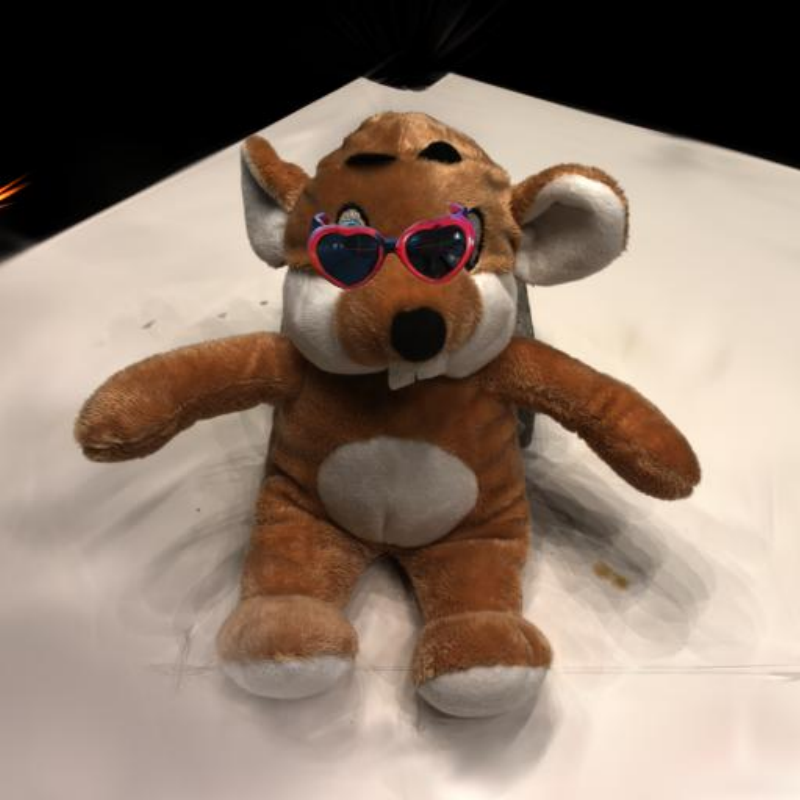}
        \par\textbf{GPT-o1\cite{jaech2024openai}}
    \end{subfigure}
    \hfill
    \begin{subfigure}{0.24\linewidth}
        \centering
        \includegraphics[width=\linewidth]{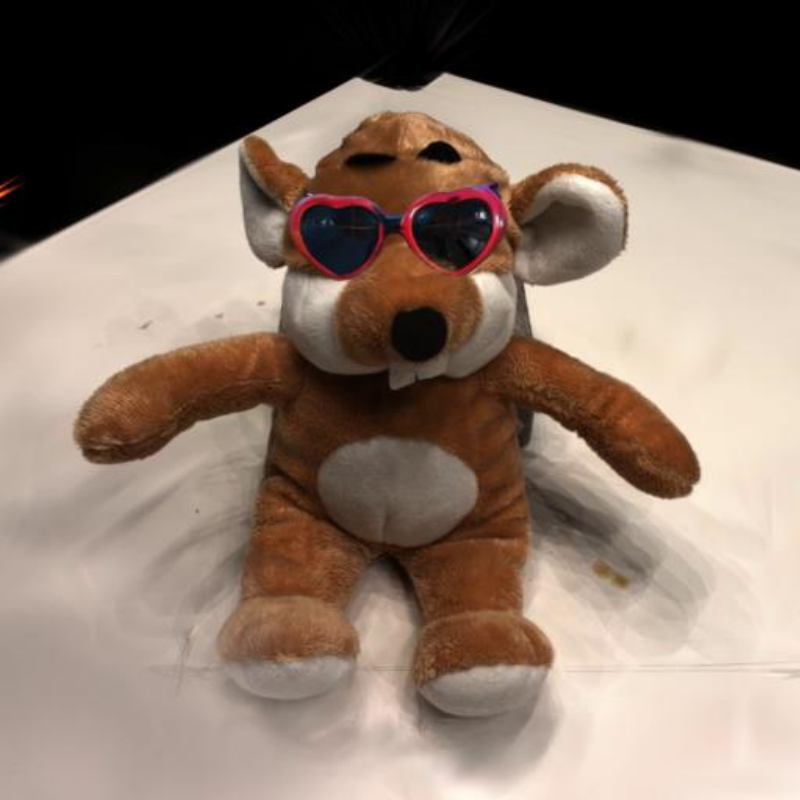}
        \par\textbf{\name}
    \end{subfigure}
    \caption{Qualitative comparison of different MLLMs for DoF learning on the ``Add a pair of sunglasses to the doll''.}
    \label{fig:comparison_dof}
    \vspace{-3mm}
\end{figure}

\noindent\textbf{Comparison of DoF learning directly by Different Multi-Modal Large Language Models vs. \name.}
To evaluate our DoF learning method, we compare it with state-of-the-art multimodal LLMs, including GPT-4V~\cite{achiam2023gpt}, Molmo-7B~\cite{deitke2024molmo}, and GPT-o1~\cite{jaech2024openai}, which directly predict object DoFs. 
Translation and scale are derived from multi-view prompts (e.g., ``\textit{Point the four coordinates of a bounding box to add [object] to/on [target]}'') and lifted to 3D, while rotation follows our initialization. 
As shown in~\Cref{fig:comparison_dof} (``\textit{Add a pair of glasses to the doll}''), our predictions are more plausible. 
Quantitatively, our method achieves the highest alignment with human preferences in projected mIoU across all cases~\Cref{tab:mIoU}.

\begin{table}[h]
    \small
    \centering
    \caption{Quantitative analysis of DoF Optimization in FreeInsert.}
    \vspace{-3mm}
    \setlength{\tabcolsep}{3pt}
    \begin{tabular}{lcccc}
        \toprule
        \textbf{Metric} & \textbf{GPT-4V} & \textbf{Molmo-7B} & \textbf{GPT-o1} & \textbf{\name} \\
        \midrule
        mIoU over 15 cases (\%) & 68.2 & 74.7 & 78.9 & 89.5 \\
        \bottomrule
    \end{tabular}
    \label{tab:mIoU}
    \vspace{-3mm}
\end{table}

\begin{figure}[htbp]  
    \centering
    \begin{subfigure}{0.24\linewidth}
        \centering
        \begin{overpic}[width=\linewidth]{figures/scene/horse/1.3_75_0.pdf}
          \put(65,-5){\includegraphics[width=0.35\linewidth]{figures/object/white_giraffe/white_giraffe_1.pdf}}
        \end{overpic}
        \caption{Original}
    \end{subfigure}
    \hfill
    \begin{subfigure}{0.24\linewidth}
        \centering
        \includegraphics[width=\linewidth]{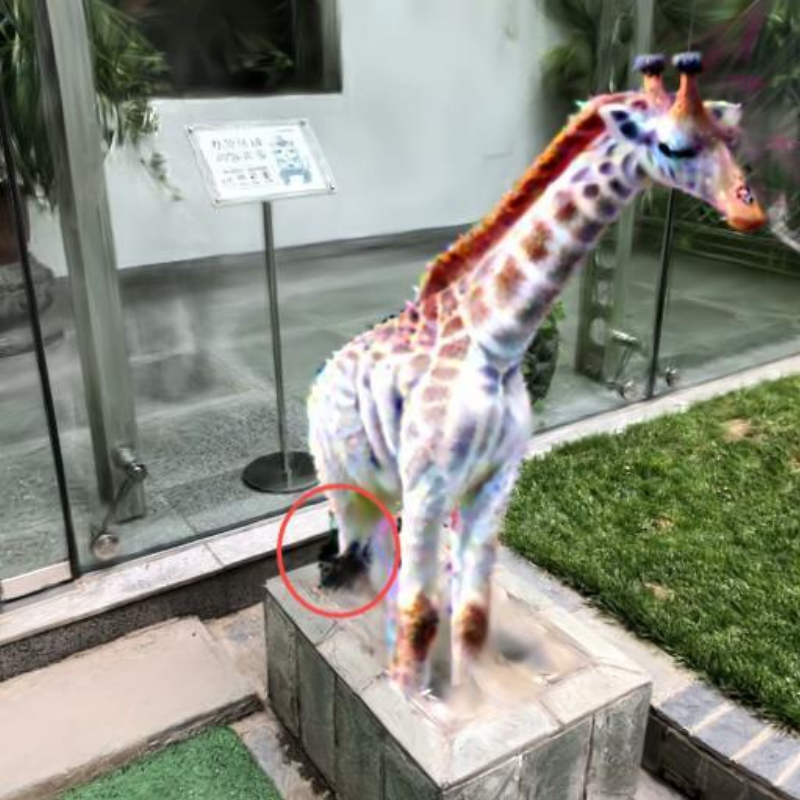}
        \caption{Single Image}
    \end{subfigure}
    \hfill
    \begin{subfigure}{0.24\linewidth}
        \centering
        \includegraphics[width=\linewidth]{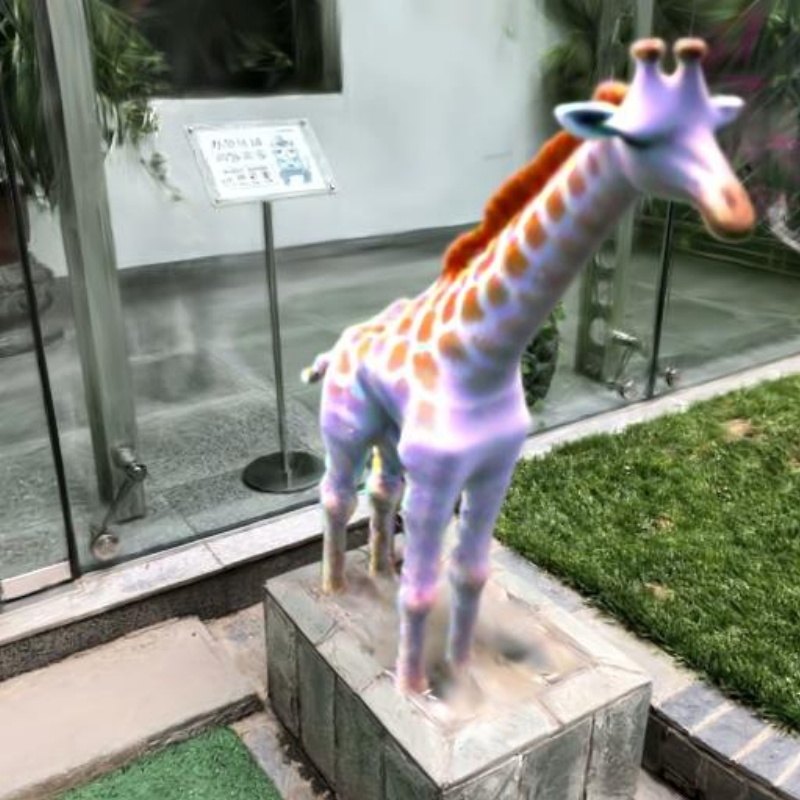}
        \caption{$ \textbf{ \textit{FR}}$=1}
    \end{subfigure}
    \hfill
    \begin{subfigure}{0.24\linewidth}
        \centering
        \includegraphics[width=\linewidth]{figures/result/horse_to_giraffe/1.pdf}
        \caption{$ \textbf{ \textit{FR}}$=3}
    \end{subfigure}
    \captionsetup{skip=2pt}
    \caption{Ablation of Viewpoint Frequency Balancing}
    \label{fig:view frequency}
\end{figure}
\noindent\textbf{Effectiveness of Viewpoint Frequency Balancing.} \Cref{fig:view frequency} compares object appearance optimization when fine-tuning LoRA with a single object image versus combining it with multi-view images at different sampling frequency ratios (${FR}$). Using only inserted-object image leads to shape inconsistencies and artifacts across views (item (b)), while incorporating multi-view data improves consistency but may reduce detail. 
Our experiments show that ${FR=3}$ offers the best trade-off between multi-view consistency and single-view image quality.

\section{Conclusion}
In this work, we presented \name, a novel framework for text-driven object insertion in 3D scenes that eliminates the need for spatial priors such as 2D masks or 3D bounding boxes. 
By disentangling object generation from spatial placement, \name enables unsupervised, flexible, and semantically guided editing through natural language. 
Leveraging the reasoning capabilities of foundation models—including MLLMs, LGM, and diffusion models—our method extracts structured semantics from user instructions to guide object reconstruction and spatial integration. 
Through hierarchical spatial aware refinement and appearance enhancement, \name achieves accurate placement and high visual fidelity. 
Extensive experiments confirm the effectiveness of our approach in enabling coherent, precise, and user-friendly 3D object insertions, paving the way for more scalable and intuitive scene editing in open-world scenarios.
While MLLMs generally demonstrate strong spatial reasoning abilities, they may occasionally misinterpret spatial relations or overlook subtle scene context, which can lead to suboptimal insertion results.
We are optimistic that unsupervised object insertion will spark new insights and serve as a valuable inspiration for future research in the community. As part of future work, we aim to further explore and enhance the spatial reasoning capabilities of large models to support more generalizable and robust object insertion across diverse 3D environments.

\bibliographystyle{ACM-Reference-Format}
\bibliography{sample-base}

\end{document}